\pdfoutput=1 

\documentclass[final]{cvpr}

\usepackage{times}
\usepackage{epsfig}
\usepackage{graphicx}
\usepackage{amsmath}
\usepackage{amssymb}

\usepackage{color}
\usepackage{bbding}
\usepackage{array,multirow,textcomp}
\usepackage{rotating}
\usepackage[skip=0.5\baselineskip]{caption}
\usepackage{makecell}
\usepackage{booktabs}
\usepackage{adjustbox}
\usepackage{array}

\usepackage{titling}

\newcolumntype{C}[1]{>{\centering\let\newline\\\arraybackslash\hspace{0pt}}m{#1}}

\usepackage{colortbl}
\definecolor{mygray}{gray}{.9}
\definecolor{gray5}{gray}{.5}
\definecolor{ggray}{RGB}{127,127,127}
\definecolor{reda}{RGB}{192,0,0}
\definecolor{redb}{RGB}{217,148,143}
\definecolor{myyellow}{RGB}{190,144,0}
\definecolor{mygreen}{RGB}{80,100,40}
\definecolor{myblue}{RGB}{30,90,100}
\definecolor{gold}{RGB}{255,204,51}

\newcolumntype{R}[2]{%
    >{\adjustbox{angle=#1,lap=\width-(#2)}\bgroup}%
    l%
    <{\egroup}%
}

\usepackage[pagebackref=true,breaklinks=true,colorlinks,bookmarks=false]{hyperref}

\def\cvprPaperID{6116} 
\def\confYear{CVPR 2021}

\begin{document}
\title{\Large \bf 
CompositeTasking: Understanding Images by Spatial Composition of Tasks 
}

\author{
Nikola Popovic$^{1}$, Danda Pani Paudel$^{1}$, Thomas Probst$^{1}$, Guolei Sun$^{1}$, Luc Van Gool$^{1,2}$\\
$^{1}$Computer Vision Laboratory, ETH Zurich, Switzerland\\
$^{2}$VISICS, ESAT/PSI, KU Leuven, Belgium\\
{\tt\small \{nipopovic, paudel, probstt, guolei.sun, vangool\}@vision.ee.ethz.ch}
}
\date{}

\maketitle

\begin{abstract}
We define the concept of CompositeTasking
as the fusion of multiple, spatially distributed tasks, for various aspects of image understanding. Learning to perform spatially distributed tasks is motivated by the frequent availability of only sparse labels across tasks, and the desire for a compact multi-tasking network. To facilitate CompositeTasking, we introduce a novel task conditioning model -- a single encoder-decoder network that performs multiple, spatially varying tasks at once. 
The proposed network takes an image and a set of pixel-wise dense task requests as inputs, and performs the requested prediction task for each pixel. Moreover, we also learn the composition of tasks that needs to be performed according to some CompositeTasking rules, which includes the decision of where to apply which task.
It not only offers us a compact network for multi-tasking, but also allows for task-editing. Another strength of the proposed method is demonstrated by only having to supply sparse supervision per task. The obtained results are on par with our baselines that use dense supervision and a multi-headed multi-tasking design. The source code will be made publicly available at ~\url{www.github.com/nikola3794/composite-tasking}.
\end{abstract}
\section{Introduction}
Intuitively, different image understanding tasks offer complementary information for scene understanding and reasoning~\cite{kim2018visual,van2019disentangled,aditya2019integrating,zador2019critique, wu2017visual, ye2020seeing,chen2019touchdown,sistu2019neurall}. Therefore, networks that can perform multiple visual tasks on the same image are of very high interest~\cite{caruana1997multitask,kevis2019attentive,Kanakis2020ReparameterizingCF,Strezoski2019ManyTL,Vandenhende2020MTINetMT}. A key aspect -- effectively serving the ultimate goal of scene understanding and reasoning -- is often not part of their design. This paper is about this utility question: can we determine  
{\it where in the image it is necessary (or even meaningful) to perform a task?}
For example, the task of recognizing human body parts is meaningful only in the presence of humans. 
Similarly, any attempt to estimate the normals of the sky is absurd. 

One may argue that we cannot know beforehand whether some task is necessary to be performed, without recognizing the image content. The content of the image may then reveal the task necessity. This begs the question whether we can know what task needs to be performed where, while bypassing the content-task pairing altogether? When the answer to {\it where} is known -- either by learning or not -- we aim to design an algorithm that executes the given multi-task instructions in an efficient manner.  For example, some applications of Augmented Reality may require human poses and the normals of the interacting surfaces. We show that such flexibility to locally activate some tasks allows us to design more compact multi-tasking networks.

The task specific annotations of images are often sparse, either by definition or due to missing annotations. Take, for example, facial landmarks or image salience. Sometimes, the annotations may be missing simply because of being futile.  Even the well-curated PASCAL-MT~\cite{pascal-context, kevis2019attentive} dataset has the sparsity of $30.4 \%$, $7.5 \%$, $41.9 \%$, $60.0 \%$, for  semantic segmentation, human body parts, surface normals, and salience, respectively. Such label sparsity only tends to get worse, if the image annotations are crowd-sourced. In fact, it is simply impractical to expect pixel-wise dense annotations for large datasets, even at locations where the annotations are well defined. The case of merging datasets, by cross-label
intersection, follows the same behaviour of sparsity. Under such circumstances, it may be unnecessary to waste computational resources during learning, for image pixels without labels. This calls for an efficient learning paradigm for multi-tasking from sparse labels.
In this work, we show that efficient learning from sparse multi-task labels and executing spatially chosen multi-task instructions go hand-in-hand.   

\begin{figure}[t]
  \centering
      \includegraphics[width=0.95\columnwidth]{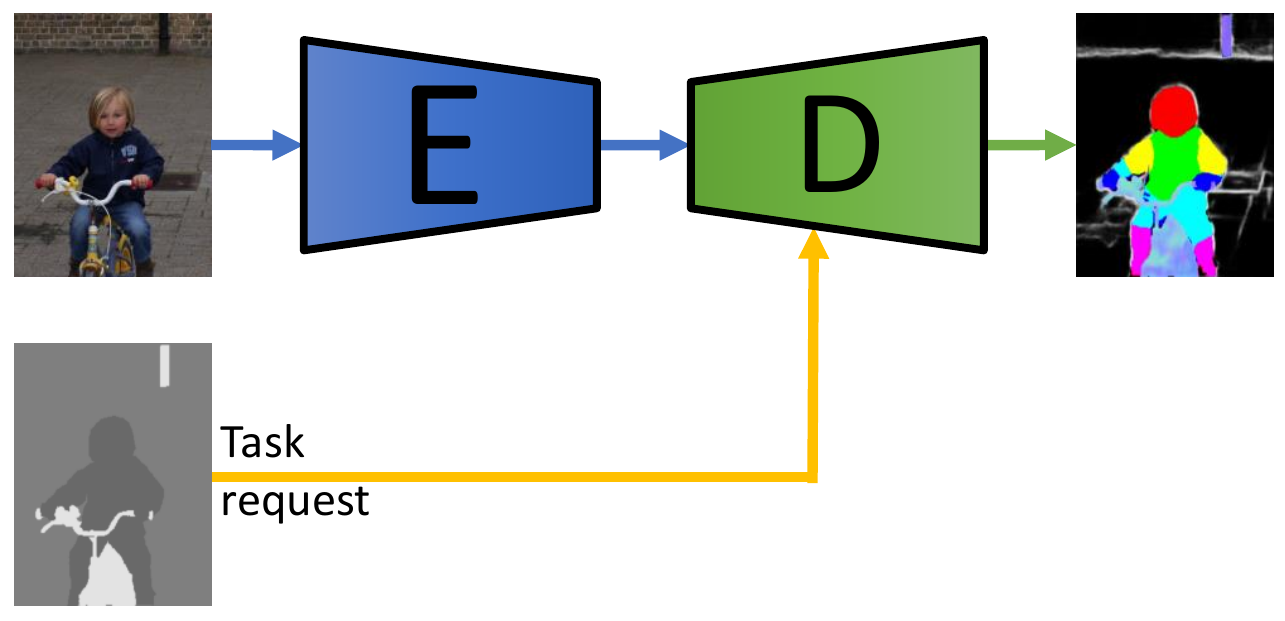}
\captionsetup{font=small}
\caption{\textbf{CompositeTasking.} Given an RGB image and a Task Palette as inputs, our CompositeTasking performs locally request-specific tasks to compute the output.}
\label{fig:overview}
\end{figure}

The key idea of this paper is rather simple. We design a convolutional neural network that performs multiple, pixel-wise tasks.  We feed every image along with a composition of spatially distributed multiple task requests -- which we call Task Palette -- to execute pixel-specific tasks. This process we call CompositeTasking. The proposed network uses a single encoder-decoder architecture to perform all the tasks in one forward pass. The simplicity of such architecture allows us to perform multiple tasks in an efficient and compact manner, thanks to the proposed method. An overview of our network is presented in Fig.~\ref{fig:overview}.

The proposed method for CompositeTasking learns by task-specific batch normalization.
Each task is performed by predicting layer-wise (only on the decoder side) affine batch normalization (BN) parameters, using a small task-conditioned network.
Such design choice dedicates the encoder towards a compact visual representation shared among tasks. On the output side, each task is represented in an image format -- thereby performing the conditional image-to-image translation. The image format chooses some arbitrary embedding for every task. We aim to map images to such embedding, conditioned upon the spatially distributed task requests. The task specific losses are then computed by mapping the predicted embedding to the task-appropriate label representations. For example, the predicted 3-channel values are mapped to class probabilities for segmentation task to compute cross-entropy, whereas, pixel normals are directly regressed by minimizing the angular distance between the prediction and the normal's label. This design choice enforces tasks to share network parameters even on the decoder side.  Surprisingly, such simple design already offers us very competitive results.          

During inference, only a small part of the embedding network performs computation. This allows our CompositeTasking network to use an efficient single encoder - single decoder architecture for all tasks. Furthermore, our training strategy enables users to request any task at any pixel. In fact, we also propose to learn the Task Palette, in case it is missing. The inferred palette follows some hand-crafted rules for task requests. It is then fed back to our network to execute the spatially distributed, rule-based tasks.           

Learning pixel-specific tasking has several benefits, which may be obvious when a parallel to the image segmentation is drawn. In this work, we demonstrate the benefits in regard to a couple of chosen applications, namely, learning the Task Palette, task editing, and rule transfer.  In the following, we summarize key contributions of our work.

\begin{itemize}
    \item We introduce the new problem of CompositeTasking which we demonstrate to be useful for images. 
    \item A novel method for CompositeTasking is also proposed. It is  significantly superior in terms of computational efficiency, and competitive in terms of performance for image understanding tasks.
    \item Applications of the proposed CompositeTasking network, namely on predicting with an estimated Task Palette, task editing, and rule transfer, are also demonstrated in this paper.
\end{itemize}

\section{Related Work}
\label{sec:related_work}

\noindent\textbf{Multi-task learning (MTL).}
MTL is concerned with learning multiple tasks simultaneously, while exerting shared influence on model parameters. The potential benefits are manifold, and include speed-up of training or inference, higher accuracy, better representations, as well as higher parameter or sample efficiency. A comprehensive survey on architectures, optimization and other aspects of MTL can be found in~\cite{Crawshaw2020MultiTaskLW}.
On one hand, many MTL methods in the literature perform multiple tasks by a single forward pass, using shared trunk~\cite{Bragman2019StochasticFG,Lu2017FullyAdaptiveFS,Vandenhende2019BranchedMN,Liu2019EndToEndML,Doersch2017MultitaskSV}, cross talk~\cite{Misra2016CrossStitchNF}, or prediction distillation~\cite{Xu2018PADNetMG,Zhang2018JointTL,Zhang2019PatternAffinitivePA,Vandenhende2020MTINetMT} architectures. 
On the other hand, the following methods perform one task at a time by conditioning a shared encoder, using feature masking~\cite{Strezoski2019ManyTL}, task-specific projections~\cite{zhao2018modulation}, attention mechanisms~\cite{kevis2019attentive} or parametrized convolutions~\cite{Kanakis2020ReparameterizingCF}, while using one decoder head for each task.
With CompositeTasking, we are bridging the gap between the two paradigms, by performing multiple tasks on the input image within one forward pass, by performing one task at a time -- for each pixel. In stark contrast to conditioning a shared encoder, as done in~\cite{kevis2019attentive,Kanakis2020ReparameterizingCF,RA_series,Bilen2017UniversalRT,zhao2018modulation,Strezoski2019ManyTL}, we instead learn an unconditioned encoder, together with pixel-wise conditioning of a single unified decoder to output the task composition.

\noindent\textbf{Conditional normalization (CN).}
Conditional normalization is the workhorse of many methods solving diverse problems in multi-domain learning~\cite{RA_series,Rebuffi2018EfficientPO}, image generation~\cite{Karras2019ASG,Brock2019LargeSG,park2019SPADE}, image editing~\cite{ntavelis2020sesame}, style transfer~\cite{Huang2017ArbitraryST}, and super-resolution~\cite{Wang2018RecoveringRT}.
The operating principle is as simple as applying condition-dependent affine transformations on the normalized batch~\cite{ioffe2015BN}, local response~\cite{Krizhevsky2017ImageNetCW}, instance~\cite{Ulyanov2016InstanceNT}, layer~\cite{Ba2016LayerN}, or feature group~\cite{Wu2018GroupN}, allowing features to occupy different regions in the space, depending on the triggered condition. While many of the aforementioned tasks however only require a conditioning on image level -- encoding for instance domain, class, or style -- we are in need to perform pixel-wise varying conditioning.
Inspired by the success of dense conditioning for semantic image synthesis, we realise our pixel-wise task conditioning via spatially-adaptive normalization~\cite{Wang2018RecoveringRT,park2019SPADE}.
To our knowledge, we are presenting the first method that learns multiple tasks with a single conditional unified head for all tasks, and is even so capable of performing multiple tasks -- for different regions -- in one forward pass.

\noindent\textbf{Learning from partial labels.}
Crowdsourcing platforms such as Amazon Mechanical Turk or reCAPTCHA made image annotation affordable, and have brought valuable contributions for the computer vision community~\cite{Russakovsky2015ImageNetLS,Lin2014MicrosoftCC,Manen2017PathTrackFT,Zhang2018TheUE}. In order to make best use of all annotators, an efficient approach is required for large-scale labelling, which may come at the price of only obtaining partial labels for each image. This trade-off is still favorable, since partially annotating more images typically outperforms dense labelling of fewer images, due to the increased variety of images seen during training~\cite{Durand2019LearningAD}.
Although the partial label problem has been addressed for diverse tasks such as segmentation~\cite{xu2015learning, Alonso2017CoralSegmentationTD}, depth densification~\cite{Qiu2019DeepLiDARDS}, or multi-label classification~\cite{Durand2019LearningAD,Huynh2020InteractiveMC}, it has only been tackled from the perspective of a single task at a time.
In contrast, with our approach one can handle partial annotations of multiple tasks in the same image. Offering the ability to focus the learning of interesting tasks in interesting regions can significantly boost sample efficiency during training. 

\section{CompositeTasking Network}
\label{sec:composite_netowrk}
In this paragraph, we introduce the formal notations. The input image is denoted as $\mathcal{I} \in \mathbb{R}^{3 \times H \times W}$, where $H$ is height and $W$ is the width of an image. The image is represented using $3$ color channels. Next, we introduce the Task Palette, which is denoted as $\mathcal{T} \in [1,...,K]^{H \times W}$. It is of the same spatial dimension as the input image $H \times W$, and it takes one of $K$ discrete values, where $K$ is the number of considered prediction tasks. The Task Palette specifies which task to predict at which pixel location. The model takes the image $\mathcal{I}$ and the Task Palette $\mathcal{T}$ as inputs and produces $\mathcal{O} = \mathsf{M} (\mathcal{I}, \mathcal{T})$, where $\mathcal{O} \in \mathbb{R}^{3 \times H \times W}$. The output has the same spatial dimension as the input image $H \times W$ and also has $3$ output channels. To construct the output $\mathcal{O}$, the model predicts task $t_{yx}$ at output location $o_{yx}$. The output $\mathcal{O}$ is called the Composite Task, since it is a spatial composition of considered tasks $1,..,K$. Pixel-wise, every task is represented as a $3$D vector $o_{yx} \in \mathbb{R}^3$. An overview  of our architecture is presented in Figure~\ref{fig:UNET-likeesign}. A detailed network diagram can be found in 
the supplementary materials.

\subsection{Network Overview}
The proposed model is divided into two parts: the encoder and the decoder network. This is inspired by the U-net \cite{ronneberger2015u} architecture. The encoder only processes the input image $\mathcal{I}$. The decoder takes the features processed by the encoder, along with the Task Palette $\mathcal{T}$, to produce the output $\mathcal{O}$.
The encoder's job is to learn to produce a very rich feature representation that is sufficient to predict all $K$ tasks. This representation is expected to capture enough information to be translated into different spatial compositions of tasks. 
The decoder's job is to take that feature representation, as well as a specific Task Palette $\mathcal{T}$, and translate it into the output $\mathcal{O}$. For the sake of simplicity, we choose $3$ channels for the output $\mathcal{O}$ to treat this problem as image-to-image translation. Choice for higher number channels can also be made, if needed. Such choice is made merely for the convenience, which is also empirically supported.

Having the same number of output channels for each task, as well as the ability to predict different tasks at different spatial locations, allows us to have a truly multi-tasking network. This means that the exact same network can predict any considered task at any considered location with the exact same architecture.
We believe that most of the visual tasks can also be embedded within few channels.
In fact, a similar practice is common in the domain of instance segmentation \cite{DBLP:conf/nips/NewellHD17,8014800,DBLP:journals/corr/abs-1708-02551,DBLP:journals/pami/LiangLWSYY18,DBLP:journals/corr/FathiWRWSGM17,DBLP:journals/corr/abs-1708-02550,DBLP:conf/cvpr/NevenBPG19} and an another related work~\cite{achille2019task2vec}.
Our image format-based output chooses some arbitrary embedding for every task. We aim to map images to such embedding, conditioned upon the spatially distributed task requests. The task specific losses are then computed by mapping the predicted embedding to the task-appropriate label representations.
This allows our network to produce the same output format, thereby making addition of new tasks very simple. More importantly, additional tasks require no additional modules, network heads, etc. The architecture of the CompositeTasking network, as well as its parameter count remain exactly the same.

\begin{figure}[t]
  \centering
  \vspace{+2pt}
      \includegraphics[width=0.95\columnwidth]{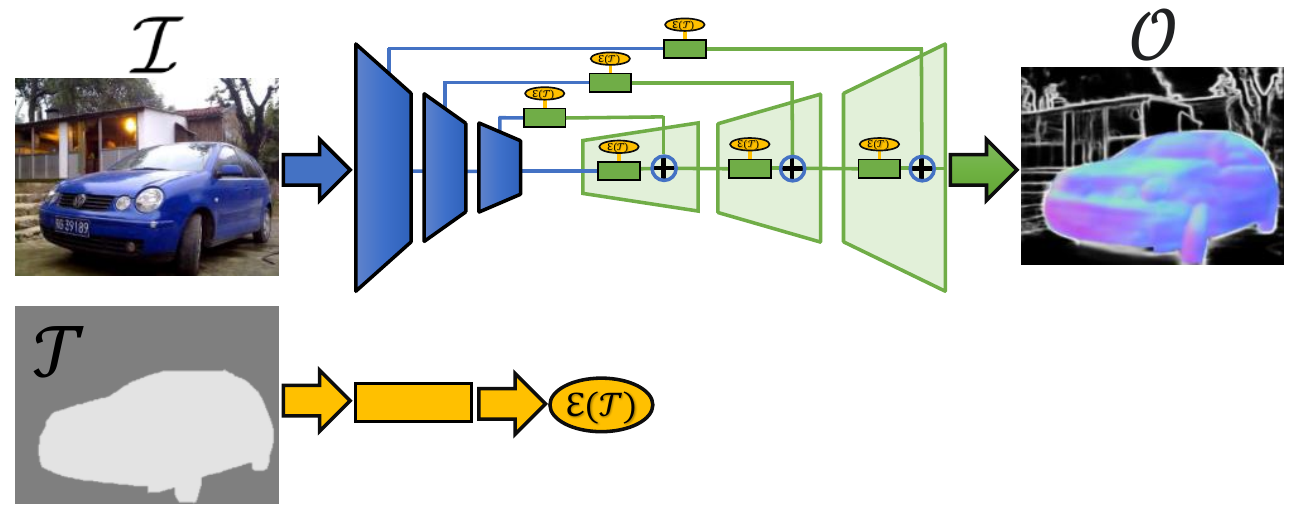}
\captionsetup{font=small}
\caption{\textbf{Overview of the CompositeTasking network architecture.} We follow a U-NET~\cite{ronneberger2015u} like design for Image-to-image translation. \textcolor{blue}{Blue} blocks are modules from the encoder backbone. \textcolor{green}{Green} blocks of the decoder are depicted in Figure \ref{fig:cond_module}. The \textcolor{gold}{yellow} block which produces the Task Palette embedding $\mathcal{E}(\mathcal{T})$ is depicted in Figure \ref{fig:e_embed_net}.\label{fig:UNET-likeesign} }
\label{fig:composit_net}
\end{figure}

\subsection{Conditioning with the Task Palette}
The spatially specific output conditioning is achieved by using a layer described in this section. This layer is inspired by \cite{park2019SPADE}, which was originally used to  generates images with desired semantic structure. We however, perform BatchNorm normalization \cite{ioffe2015BN} only in the decoder, where task specific affine transformation parameters are predicted differently for each pixel conditioned upon the Task Palette $\mathcal{T}$ at that spatial location $t_{yx}$. For the notational convenience, we follow \cite{park2019SPADE}. Let $\mathbf{h}^i$ denote the activation of layer $i$ in the decoder, for a batch of $N$ samples. Let $H^i$ and $W^i$ be the height and width of the activation map and let $C^i$ be the number of channels in layer $i$. Then, the task specific conditioning is achieved by using a layer which computes,
\begin{equation}
    \label{eq:cond}
    h^{i+1}_{ncyx} = \gamma^{i}_{cyx}(t_{yx}) \frac{h^{i}_{ncyx} - \mu^{i}_{c}}{\sigma^{i}_{c}} + \beta^{i}_{cyx}(t_{yx}),
\end{equation}
where $h^{i+1}_{ncyx}$ is the output of our task-specific conditioning layer, and $\mu^{i}_{c}$ and $\sigma^{i}_{c}$ are the mean and standard deviation of the activations in channel $c$:
\begin{align}
    \mu^{i}_{c} = \frac{1}{N H^i W^i} \sum_{n,y,x} h^{i}_{ncyx},\\
    \sigma^{i}_{c} = \sqrt{\frac{1}{N H^i W^i} \sum_{n,y,x} \left((h^{i}_{ncyx})^2 - (\mu^{i}_{c})^2\right)}.
\end{align}

The affine parameters $\gamma^{i}_{cyx}(t_{yx})$ and $\beta^{i}_{cyx}(t_{yx})$ condition the normalized activation $h^{i}_{ncyx}$ based on the requested tasks $t_{yx}$. Unlike \cite{park2019SPADE}, which allows surrounding semantics dependent $\gamma^{i}_{cyx}$ and $\beta^{i}_{cyx}$ by using $3 \times 3$ convolutions on conditioned semantics, we keep operations for each task request independent. In other words, during conditioning, \cite{park2019SPADE} aims to fit pixels meaningfully in the surrounding, whereas we are interested to make each request independent. This choice is motivated by the potential of flexible applications of our method, some of which are demonstrated later in this paper. 
In this process, we first transform the Task Palette to an embedding $\mathcal{E} = f(\mathcal{T})$, where $\mathcal{E} \in \mathbb{R}^{H \times W \times N_w}$.  
The parameters $\gamma^{i}_{cyx}(e_{yx}(t_{yx}))$ and $\beta^{i}_{cyx}(e_{yx}(t_{yx}))$ are then obtained using the embedding $\mathcal{E}$. Here, we present further details of our method using two blocks:  
 (a) task representation; and (b) task composition.

\begin{figure}[t]
  \centering
      \includegraphics[width=0.95\columnwidth]{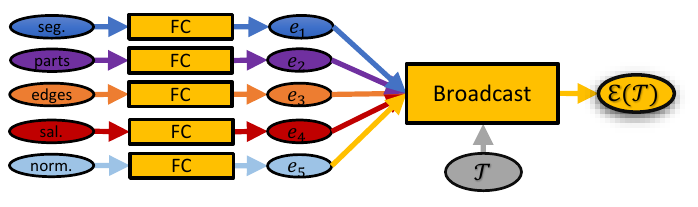}
\captionsetup{font=small}
\caption{\textbf{Task representation block.} Each task is process independently to learn the task-specific embedding. We broadcast these embeddings according to the task request $\mathcal{T}$. The broadcast $\mathcal{E}$ is fed into the composition blocks from Figure~\ref{fig:cond_module}.}
\label{fig:e_embed_net}
\end{figure}

\paragraph{Task representation block.} 
In order to embed the Task Palette $\mathcal{T}$ into $\mathcal{E} = f(\mathcal{T})$ ,  we first learn the task specific embeddings $ \{e_1,e_2,\ldots,e_K\}$ for each task. This is done by embedding all distinct values of the Task Palette $e_k=f(z_k)$, where $z_k$ is the unique task code. The Palette's embedding $\mathcal{E}$ is then obtained by broadcasting  task specific embeddings according to the task requests $\mathcal{T}$. All of the Task Composition Blocks use the same embedding $\mathcal{E}$. In Figure~\ref{fig:e_embed_net}, we can see that each task is processed independently through a fully connected Neural Network, before broadcasting into $\mathcal{E}$.

\begin{figure}[t]
  \centering
      \includegraphics[width=0.95\columnwidth]{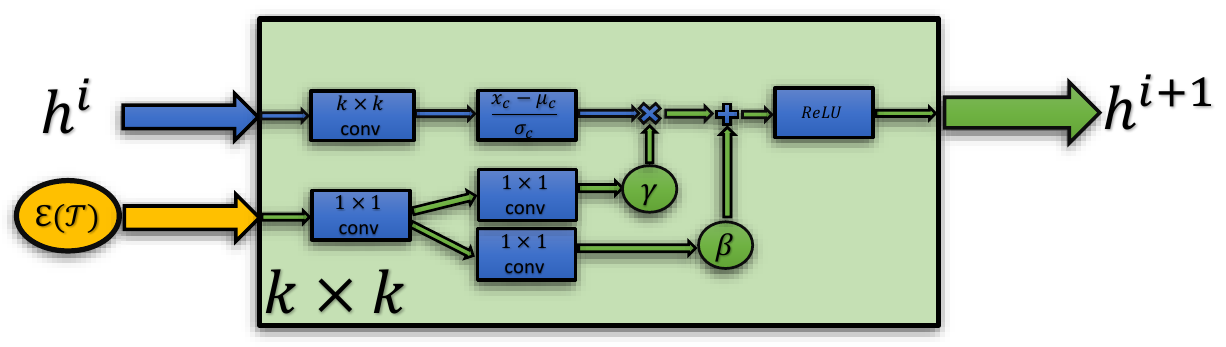}
\captionsetup{font=small}
\caption{{\textbf{Task composition block}. This block receives task embedding  and previous layer's features as inputs, and performs task-conditioned transformation of the  features.}}
\label{fig:cond_module}
\end{figure}

\paragraph{Task composition block.}  
A graphical representation of this block depicted in Figure~\ref{fig:cond_module}.
The task composition block receives the task embedding $\mathcal{E}$ and the features from the previous layer of the network. The conditioning operation of~\eqref{eq:cond} takes place within this block as follows: (i) features are processed using a standard convolution layer; (ii) embedding $\mathcal{E}$ is processed independently for each task using two  layers of $1 \times 1$ convolutions to obtain $\gamma^{i}_{cyx}(e_{yx}(t_{yx}))$ and $\beta^{i}_{cyx}(e_{yx}(t_{yx}))$; (iii)  the operation of~\eqref{eq:cond} is then performed followed by an activation function. Output of this block is the task-conditioned transformed features.  As shown in  Figure~\ref{fig:composit_net}, we use the task composition blocks only in the decoder and skip connections of our network. 
 
\subsection{Computing the Loss and Training}
Every task $k$ has its own loss function of interest $\mathcal{L}_k$ . The total loss is given by:
\begin{equation}
    \label{eq:total_loss}
    \mathcal{L} = \sum_{k} 
    \lambda_{k}
    \mathcal{L}_k (\mathcal{O}, \mathcal{Y}_k, \mathcal{T}),
\end{equation}
where $\mathcal{Y}_k$ are the labels for task $k$. Usually the losses are computed per pixel location as $\mathcal{L}_k(o_{yx}, y_{yx})$, but that doesn't necessarily have to be the case. The hyperparameter $\lambda_k$ helps to define the balance/trade-off of the predictive performance of all $K$ tasks  under consideration. 
The CompositeTasking network uses a standard training procedure. The images from the training set $\mathcal{I}$ are continuously passed to the network as inputs, along with the desired Task Palettes $\mathcal{T}$ to predict the output $\mathcal{O} = \mathsf{M} (\mathcal{I}, \mathcal{T})$. The predicted output $\mathcal{O}$ is given to the loss function \eqref{eq:total_loss}, along with the corresponding labels $\mathcal{Y}_1,...,\mathcal{Y}_K$ and Task Palette $\mathcal{T}$. The final loss is minimized using standard optimization algorithms~\cite{kingma2014adam}.

\section{Tasks and Rules}
\label{sec:tasks_and_rules}
We consider the following dense prediction tasks.

\noindent\textbf{Semantic segmentation.} This is the task of predicting which of the defined semantic classes the pixel belongs to.

\noindent\textbf{Human body parts.} Similar to semantic segmentation, this is the task of predicting which of the defined human body parts the pixel belongs to.

\noindent\textbf{Surface normals.} This is the task of predicting the $3$D orientation of the surface contained in the pixel.

\noindent\textbf{Semantic edges.} This is the task of predicting edges between different objects on the input image.

\noindent\textbf{Saliency.} This is the task of predicting which locations in the image are most conspicuous for human observers.

Since the network output is constrained to 3 channels, we need to embed the tasks accordingly. Surface normals are $3$D vectors by nature, and fit in the output shape. In the case of edges and saliency, the output corresponds to a scalar probability of the positive class. One way to embed them is to predict them at all $3$ channels and calculate the mean at test-time. For the task of semantic segmentation and human parts, the pixel-wise outputs are usually represented as length $C$ vectors representing the probability of each class. To this end, we transform the pixel-wise $3$D output $\mathbf{o}$. First, we define $3$D class anchors $\mathbf{a}_i$ to each class $i$, uniformly spread out in space 
(more details in the supplementary materials).
Then, we compute a score vector $\mathbf{l} \in \mathbb{R}^{C}$ based on the distance to the class centers,
\begin{equation}
    \label{eq:seg_loss}
    l_i = \frac{1}{\| \mathbf{o} - \mathbf{a}_i \|  + \epsilon},
\end{equation}
where $\epsilon$ is a small constant. Hence, $\mathbf{l}$ has the highest value at the index of the closest class center. Applying a softmax operation $\hat{o}_i = \frac{e^{l_i}}{\sum_{j=1}^{C}e^{l_j}}$ transforms $\mathbf{l}$ to a probability measure that can be used with the common loss functions. One can argue that if we look at $\hat{o}$ as the predicted class-probabilities, it is biased by the arbitrary definition of class anchors. If the class $i$ has the highest predicted probability $i=\arg\!\max_j \hat{o}_j$, only one of the classes $\{j: a_j \in \mathcal{N}(\hat{o}_i)\}$ close to $ \hat{o}_j$ can have the second highest probability. Therefore we can not analyze class similarity by looking at these predicted probabilities. However, since our fist priority for the segmentation tasks is predicting the correct class, this offers a simple and effective solution. In case one is interested in making more detailed conclusions from the prediction, alternative approaches can be taken.

The rules for constructing Task Palettes $\mathcal{T}$ for our experiments are as follows.

\noindent\textbf{Single task rule $\mathcal{S}$}: The Task Palette $\mathcal{P}$ has the same value $k$ at every location $\forall x,y: t_{yx}=k$. Task $k$ can be changed every time the Task Palette is requested from this rule.

\noindent\textbf{Random mosaic rule $\mathcal{R}_{1r}$}: The image is spatially divided into four rectangles by intersecting a vertical and horizontal line through a randomly chosen point $\mathbf{c}=(c_x, c_y)$. Each region gets a task assigned to it randomly. The assigned tasks, as well as the point $\mathbf{c}$, can be changed every time the Task Palette is requested from this rule.
    

\noindent\textbf{Semantic rules $\mathcal{R}_2$ and $\mathcal{R}_3$}: These rules assign the tasks with respect to the image semantics.

\noindent\textbf{Random rule $\mathcal{R}_{rnd}$}: This rule assigns a randomly chosen task to each pixel independently.

More details on these rules can be found in the supplementary materials.

Our rationale for choosing these rules is based on our desire to analyse the behaviour of our method. Rule $\mathcal{S}$ is used in the field for solving specific problems. Rule $\mathcal{R}_{1r}$ is one way of seeing what happens if you train and test the network by mixing tasks in the output randomly, without any specific rule or structure behind it. Rules $\mathcal{R}_2$ and $\mathcal{R}_3$ represents rules with some semantic meaning behind it. Finally, rule $\mathcal{R}_{rnd}$ is designed to test the proposed method's limits.

\section{Implementation Details}

\subsection{Data Set Description}
The experiments are conducted on the PASCAL-MT data set from \cite{kevis2019attentive}. While constructing the data set authors distilled labels for some of the tasks, while others were used from PASCAL~\cite{PASCAL} or PASCAL-Context~\cite{pascal-context}. The data set contains $4998$ training and $5105$ validation images. We predict the tasks mentioned in section \ref{sec:tasks_and_rules}.  We evaluate the performance of semantic segmentation and human body parts with the mean intersection over union (mIoU). We evaluate the performance of saliency with computing the maximal mIoU over different thresholds. We compute the prediction of surface normals as the mean angular error (mErr). And finally, we evaluate the performance of edges with the optimal dataset F-measure (odsF) \cite{odsF}, using the implementation from \cite{seism}. These evaluation metrics are in concordance with recent multi-tasking work \cite{kevis2019attentive,Kanakis2020ReparameterizingCF}.

\subsection{Experimental Setup}
\label{sec:experimental_setup}
In our experiments we use the following models:

\noindent\textbf{CompositeTasking network (CTN)}. This is the network we proposed in section \ref{sec:composite_netowrk}. The network uses a ResNet34 \cite{he2016resnet} encoder. The decoder is build using spatially varying conditioning blocks from Figure~\ref{fig:cond_module}, and it is much smaller than the encoder, in terms of network parameters. The conditioning blocks use $1 \times 1$ regular convolutions in the skip connections and $3 \times 3$ elsewhere, which performed well empirically.

\noindent\textbf{Single task networks baseline (STN)}. Here we have different networks for different tasks. Each network has the same architecture as the CompositeTasking network, but instead of the spatially varying conditioning, they use a regular BatchNorm. This way the network has the same capacity. In the case of the single tasking rule $\mathcal{S}$ from section \ref{sec:tasks_and_rules}, each task will be supervised by a different network. With the other CompositeTasking rules $\mathcal{R}_{1r}$ and $\mathcal{R}_{2}$ it is going to do the same, which means that the network for a specific task will only be supervised on the pixels that correspond to that task in the Task Palette $\mathcal{T}$.

\noindent\textbf{Multi-head network baseline (MHN)}. Here we have a network with a shared encoder, and a different decoder for different tasks. This is a standard approach in multi-task learning~\cite{RA_series,Bilen2017UniversalRT,kevis2019attentive,Kanakis2020ReparameterizingCF}. The encoder is the same as in the CompositeTasking network, while the decoders have the same architecture, but use regular BatchNorm instead of the task specific conditioning. Similarly as above, this network is going to be supervised by supervising each decoder only on the pixels from its corresponding task.

Since CompositeTasking is a new concept, we decided to evaluate the performance of our proposed CompositeTasking network (Figure \ref{fig:composit_net}) with standard baselines. The STN is a common pipeline for solving specific tasks in computer vision, while the MHN is a common pipeline for solving multiple tasks simultaneously in a multi-tasking fashion. 

For more implementational details and hyper-parameter values look at the supplementary materials.
\section{Experiments}
\label{sec:experiments}

\subsection{General Behaviour}
\label{sec:general_behaviour}

To compare the performance of a method $m$ with baseline models from Section~\ref{sec:experimental_setup},  we use the average per-task drop with respect to the single-tasking baseline $b$, $\Delta_m=\frac{1}{T}\sum_{i=1}^{T}(-1)^{l_i}\frac{M_{m,i}-M_{b,i}}{M_{b,i}}$, where $l_i = 1$ if a lower value is better for measure $M_i$ of task $i$, and $0$ otherwise~\cite{kevis2019attentive}. 

\begin{figure}[t]
    \centering
    \captionsetup{font=small}
    \vspace*{2pt}
      \includegraphics[width=0.95\columnwidth]{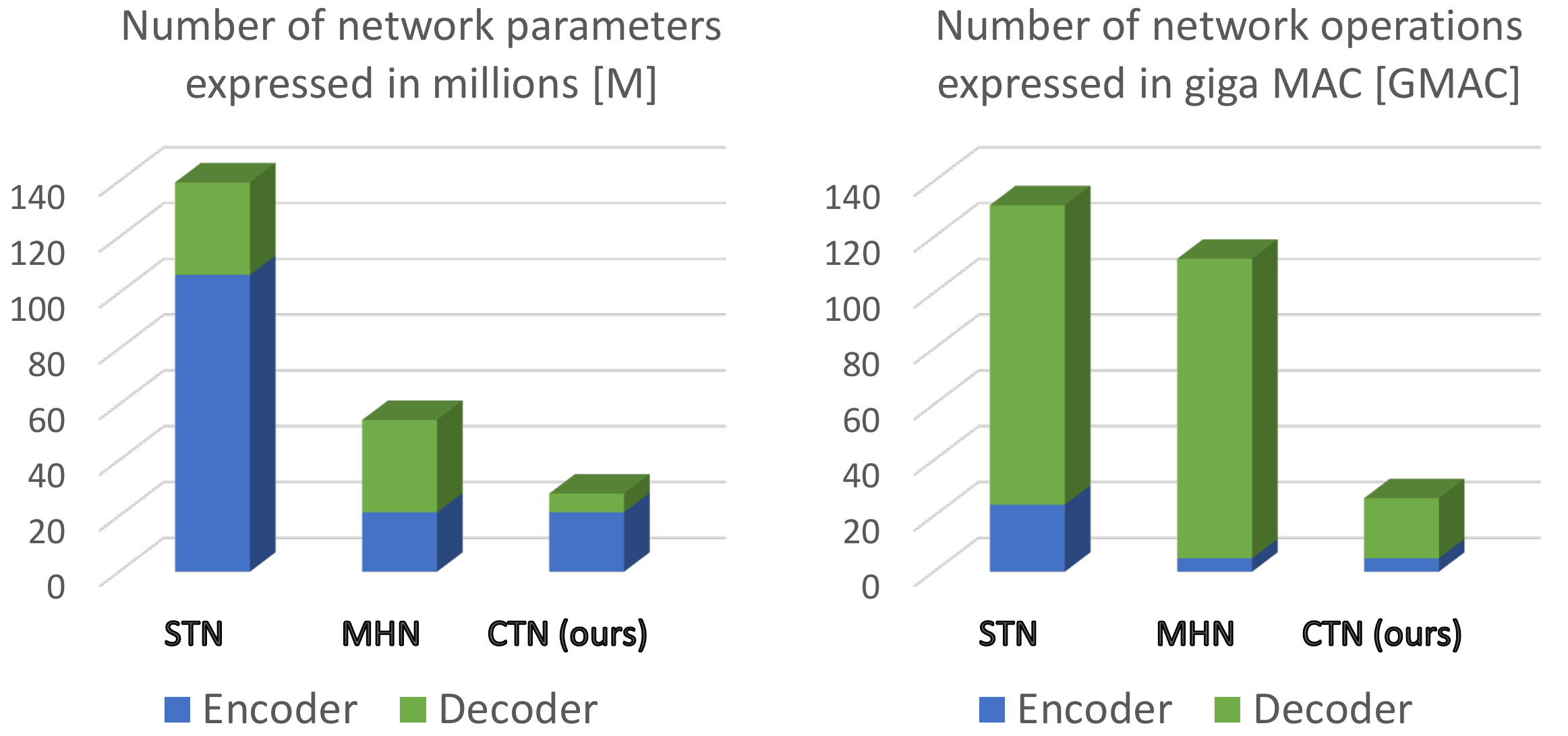}
\caption{\textbf{Model Complexity.} Memory and computational requirements of the compared methods for predicting $5$ tasks. }
\label{fig:net_complexity}
\end{figure}

We first evaluate on the single-task rule $\mathcal{S}$. From Table~\ref{table:random_mosaics}, we can see that the CompositeTasking network is on par with the baselines, even though it is trained with randomly cropped label regions of rule $\mathcal{R}_{1r}$, and tested on the single-task setting $\mathcal{S}$. This is very interesting, since it is substantially more compact in terms of memory and computational complexity, as can be seen in Figure \ref{fig:net_complexity}. Also, this experiment tells us that a lot can be learned even if only arbitrary regions for labels are being presented during training, as is the case with rule $\mathcal{R}_{1r}$. More interestingly, the part of the label that is being presented during training is a random rectangle region without any semantic meaning behind the chosen region, and still the network performs competitively to the strongest multi-head baseline trained on complete labels. A few examples of the networks predictions are presented in Figures~\ref{fig:mosaic_prediction} and~\ref{fig:st_predictions}. More examples are presented 
in the supplementary materials.
We can see that it poses no problem for the network to sharply switch from predicting one task to another with negligible boundary artifacts, while using the exact same architecture to predict different tasks. For reference, the results from Table~\ref{table:random_mosaics} can be compared to SotA baselines in~\cite{kevis2019attentive} (Tables 2 and 3) and~\cite{Kanakis2020ReparameterizingCF} (Table 3).

In Table \ref{table:r2_rule}, we see the performance evaluated using the semantic rule $\mathcal{R}_{2}$ where the tasks are being requested only at sparse, but meaningful and compact regions for each task. When we supervise by using rule $\mathcal{R}_{2}$, again we see that the CompositeTasking network performs on par with the baselines that have a lot more parameters. This shows that it is not necessary to waste so much resources when dealing with very sparse labels. The performance of our model is almost the same as the much more demanding baseline, that has a separate network for each task.

\begin{table}[t]
	\centering\small
    \captionsetup{font=small}
	\caption{{\textbf{Testing the models on the single task rule}. Our method achieves the same performance as the baselines with much more capacity, when trained on randomly cropped label regions $\mathcal{R}_{1r}$.}}
	\label{table:random_mosaics}
	\resizebox{\columnwidth}{!}{
	\setlength\tabcolsep{6pt}
	\renewcommand\arraystretch{1}
	\begin{tabular}{|c|c|c|c|c|c|c||c|c|c}
		\hline
\rowcolor{mygray}
Training rule & Method & Edge$\uparrow$& SemSeg$\uparrow$ & Parts$\uparrow$ & Normals$\downarrow$ & Sal$\uparrow$ & $\Delta_m\%$$\uparrow$ \\  \hline \hline
\multirow{2}{*}{$\mathcal{S}$} 
& STN & 69.50 & 63.69 & 58.76 & 15.58 & 69.38 & 0.0\%  \\  \cline{2-8} 
& MHN & 68.10 & 60.77 & 54.21 & 16.44 & 67.21 &  -4.60 \%  \\  \hline \hline
\multirow{2}{*}{$\mathcal{R}_{1r}$}
& STN & 68.30 & 59.82 & 49.88 & 16.07 & 69.94 & -5.05\% \\  \cline{2-8} 
& MHN & 67.70 & 61.64 & 52.84 & 16.40 & 67.70 & -4.71\%  \\  \cline{2-8} 
& CTN(Ours) & 68.60 & 62.45 & 52.59 & 16.93 & 67.81 & -4.93\% \\  \hline
\end{tabular}
}
\end{table}

\begin{table}[t]
	\centering\small
\captionsetup{font=small}
	\caption{{\textbf{Testing the models on the semantic rule}. Our method achieves the same performance as the baselines with much more capacity, when trained on the semantic rule $\mathcal{R}_{2}$.}}
	\label{table:r2_rule}
	\resizebox{\columnwidth}{!}{
	\setlength\tabcolsep{6pt}
	\renewcommand\arraystretch{1}
	\begin{tabular}{|c|c|c|c|c|c|c||c|c|c}
		\hline
\rowcolor{mygray}
Training rule & Model & Edge$\uparrow$& SemSeg$\uparrow$ & Parts$\uparrow$ & Normals$\downarrow$ & Sal$\uparrow$ & $\Delta_m\%$$\uparrow$ \\  \hline \hline
\multirow{2}{*}{$\mathcal{S}$}  
& STN & 65.80 & 79.32 & 56.21 & 14.75 & 73.23 &  0.0\% \\ \cline{2-8} 
& MHN & 64.60 & 73.94 & 50.65 & 15.68 & 71.01 & -5.57\%\\  \hline \hline
\multirow{3}{*}{$\mathcal{R}_{1r}$} 
& STN & 64.80 & 74.12 & 44.81 & 15.42 & 73.22 & -6.58\% \\ \cline{2-8}  
& MHN & 65.20 & 75.97 & 49.07 & 15.62 & 71.74 & -5.15\% \\ \cline{2-8} 
& CTN(Ours) & 62.30 & 76.73 & 48.87 & 16.45 & 71.40 &  -7.13\%\\  \hline \hline
\multirow{3}{*}{$\mathcal{R}_{2}$}
& STN & 63.90 & 83.91 & 59.63 & 17.13 & 70.04 & -2.30\% \\ \cline{2-8}
& MHN & 64.40 & 83.71 & 58.44 & 17.44 & 67.02 &  -3.87 \% \\  \cline{2-8}
& CTN(Ours) & 69.20 & 84.70 & 59.74 & 18.12 & 67.95 &  -2.37\% \\  \hline
\end{tabular}
}
\end{table}

\begin{figure}[t]
  \centering
  \captionsetup{font=small}
  \scriptsize
  \setlength{\tabcolsep}{1pt}
  \vspace*{2pt}
  \newcommand{\sz}{0.189}           
  \newcommand{\cgs}{\hspace{3pt}}   
  \begin{tabular}{ccccc}
    \multirow{1}{*}[15pt]{\rotatebox{90}{\textbf{Image}}}
    \includegraphics[width=\sz\columnwidth]{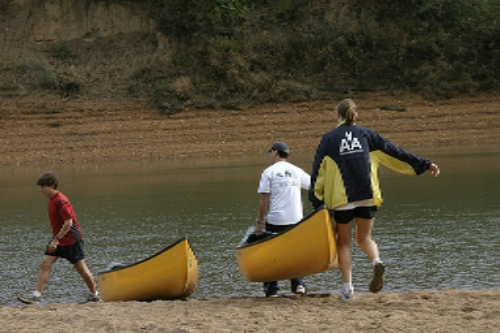} &
    \includegraphics[width=\sz\columnwidth]{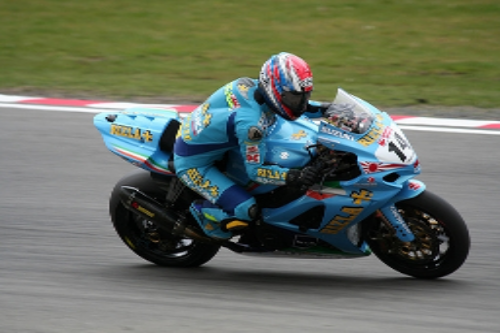} &
    \includegraphics[width=\sz\columnwidth]{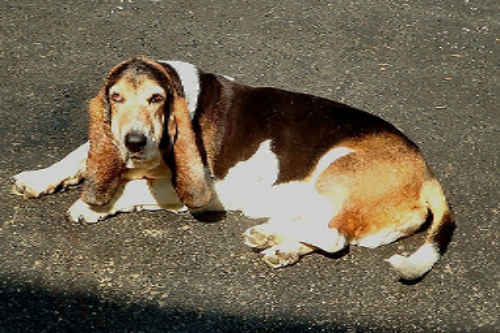} &
    \includegraphics[width=\sz\columnwidth]{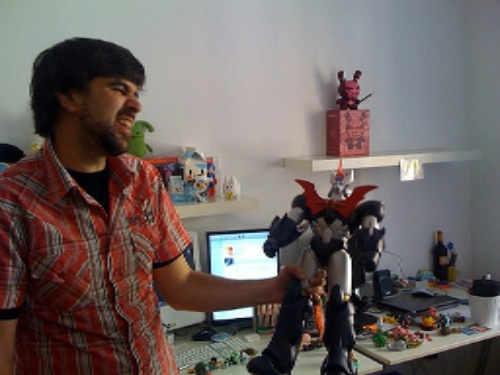} &
    \includegraphics[width=\sz\columnwidth]{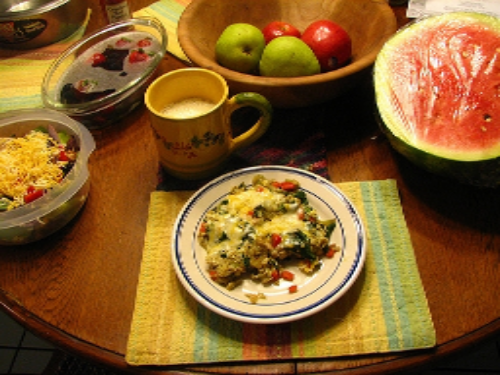} \\
    
    \multirow{1}{*}[15pt]{\rotatebox{90}{\textbf{Pred}}}
    \includegraphics[width=\sz\columnwidth]{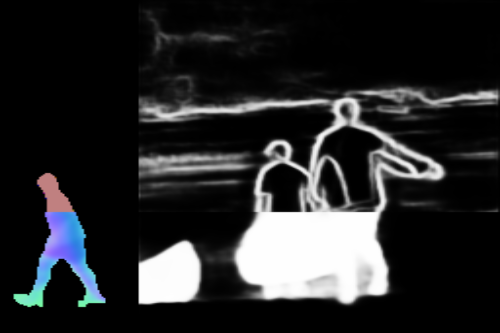} &
    \includegraphics[width=\sz\columnwidth]{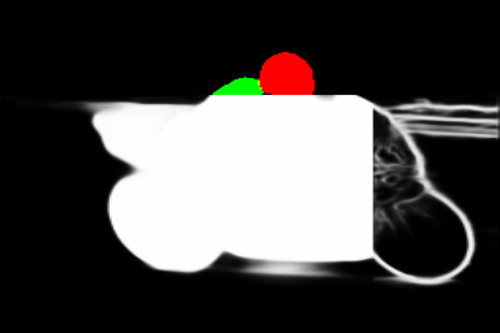} &
    \includegraphics[width=\sz\columnwidth]{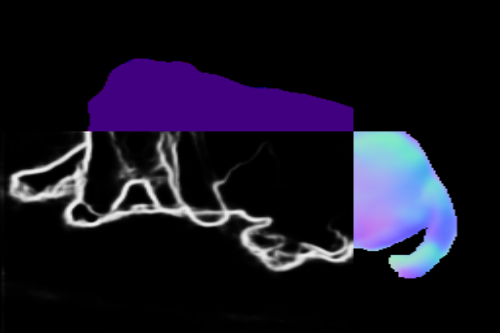} &
    \includegraphics[width=\sz\columnwidth]{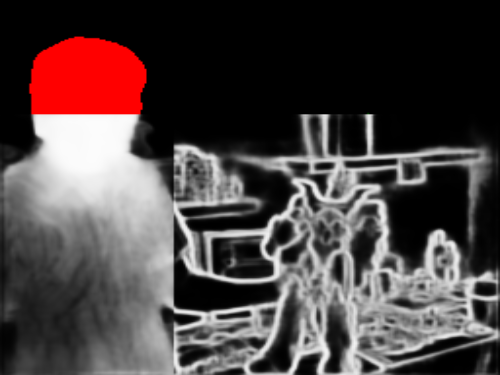} &
    \includegraphics[width=\sz\columnwidth]{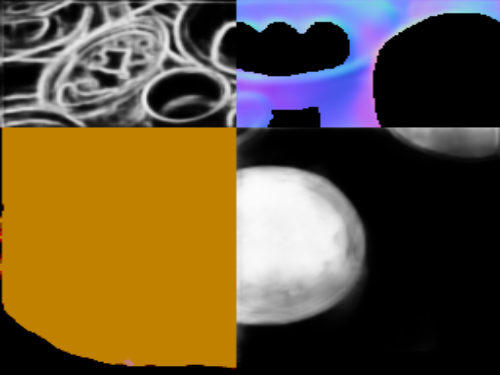}

  \end{tabular}
 \caption{\textbf{Random mosaic compositions.} CompositeTasking network predictions on requests with the $\mathcal{R}_{1r}$ rule. The network shows the ability to sharply switch to a different tasks at neighbour pixels.}
    \label{fig:mosaic_prediction}
\end{figure}

\begin{figure}[t]
  \centering
  \captionsetup{font=small}
  \scriptsize
  \setlength{\tabcolsep}{1pt}
  \vspace*{2pt}
  \newcommand{\sz}{0.1575}           
  \newcommand{\cgs}{\hspace{3pt}}   
  \begin{tabular}{cccccc}
    \includegraphics[width=\sz\columnwidth]{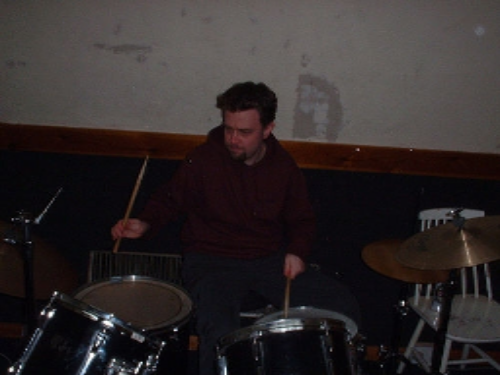} &
    \includegraphics[width=\sz\columnwidth]{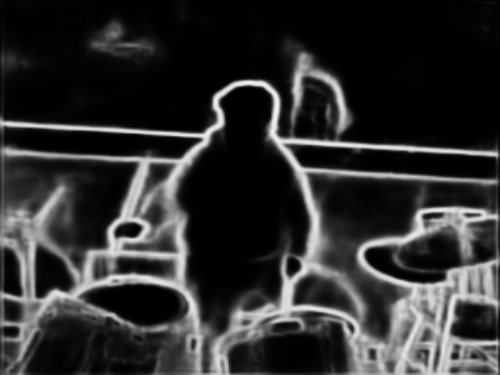} &
    \includegraphics[width=\sz\columnwidth]{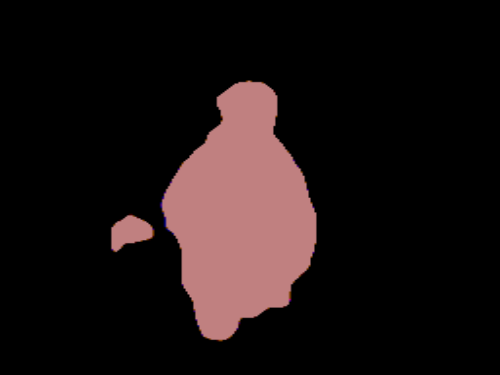} &
    \includegraphics[width=\sz\columnwidth]{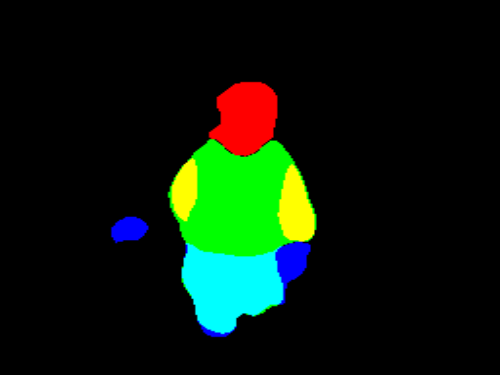} &
    \includegraphics[width=\sz\columnwidth]{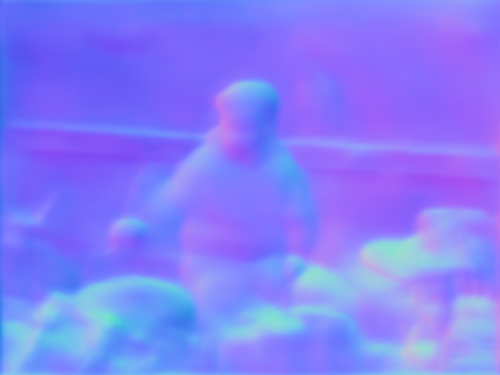} &
    \includegraphics[width=\sz\columnwidth]{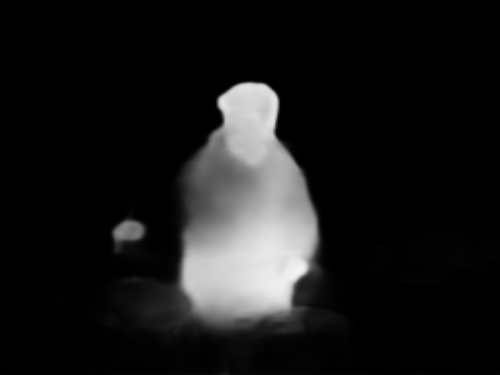} \\
    
    \includegraphics[width=\sz\columnwidth]{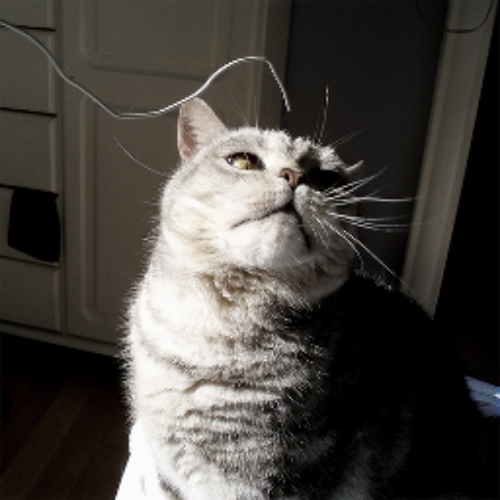} &
    \includegraphics[width=\sz\columnwidth]{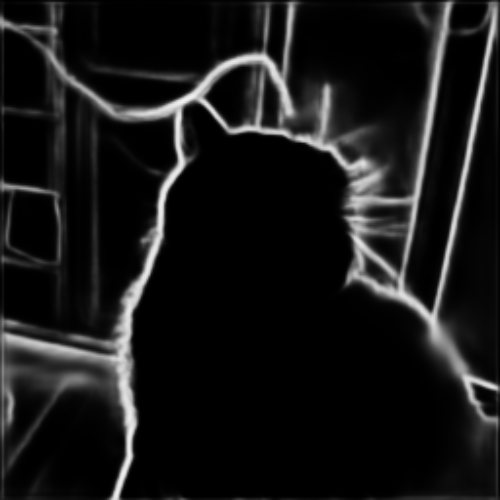} &
    \includegraphics[width=\sz\columnwidth]{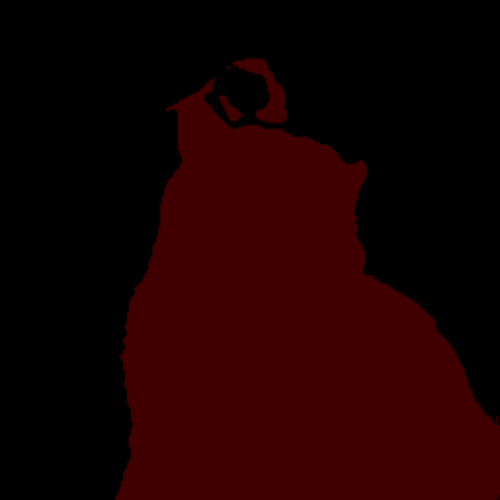} &
    \includegraphics[width=\sz\columnwidth]{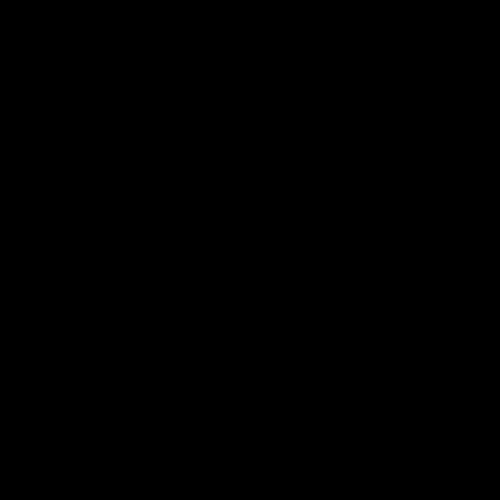} &
    \includegraphics[width=\sz\columnwidth]{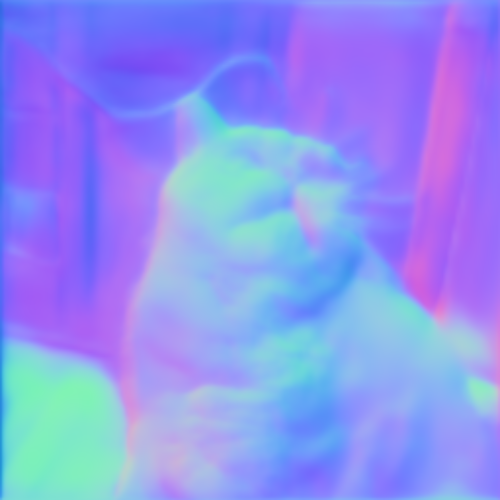} &
    \includegraphics[width=\sz\columnwidth]{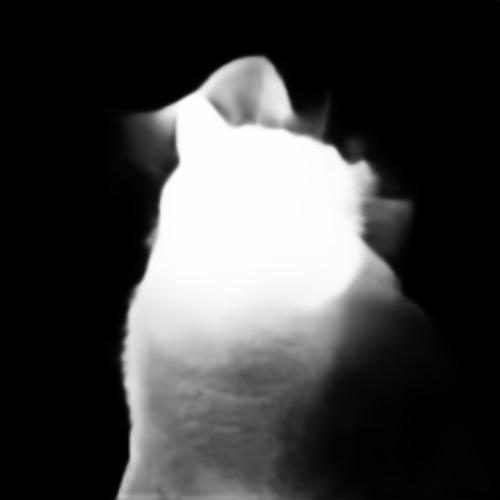} 

  \end{tabular}
 \caption{\textbf{Single task predictions.} Even though our model is made for CompositeTasking, it can also make predictions on requests with the $\mathcal{S}$ rule. }
    \label{fig:st_predictions}
\end{figure}

\subsection{Learning What to Do Where}
\label{sec:learning_where}

Up until now we have considered cases when it is known what tasks to predict where (the Task Palette $\mathcal{T}$ is known). This is definitely interesting in some use-cases, like for example Augmented Reality applications where a user can specifically request what he wishes the algorithm to do. It is even more interesting for the Task Palette to be predicted by the network itself, given the input image.  One such example is when we have a semantic rule like $\mathcal{R}_{2}$, where for every image we can supervise what needs to be predicted where. We trained a network to predict the Task Palette from the input image ($75.04 \%$ mIoU), during supervision with $\mathcal{R}_{2}$. This can be used for  automatic data labelling when we are interested in obtaining labels for multiple different tasks, but only in sparse regions of interest.

\begin{figure}[t]
  \centering
  \captionsetup{font=small}
  \scriptsize
  \setlength{\tabcolsep}{1pt}
  \vspace*{2pt}
  \newcommand{\sz}{0.189}           
  \newcommand{\cgs}{\hspace{3pt}}   
  \begin{tabular}{ccccc}
    Image & Task palette & Prediction & Predicted Palette & Prediction \\
    \includegraphics[width=\sz\columnwidth]{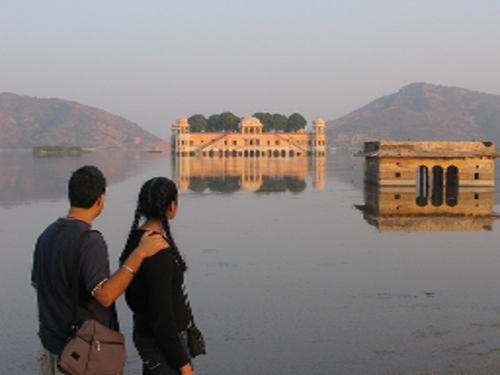} &
    \includegraphics[width=\sz\columnwidth]{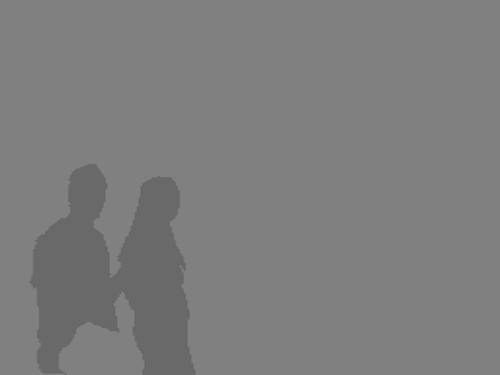} &
    \includegraphics[width=\sz\columnwidth]{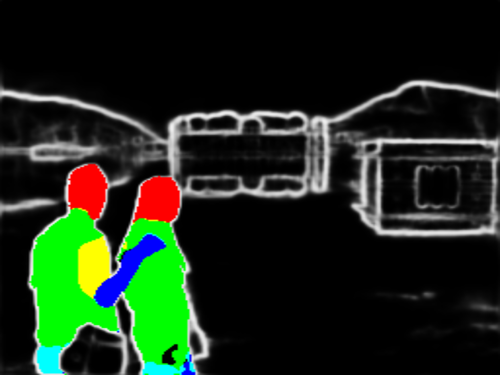} &
    \includegraphics[width=\sz\columnwidth]{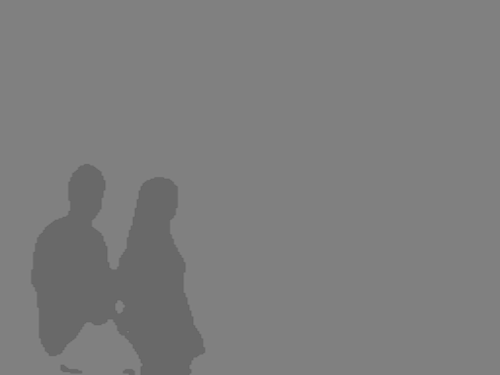} &
    \includegraphics[width=\sz\columnwidth]{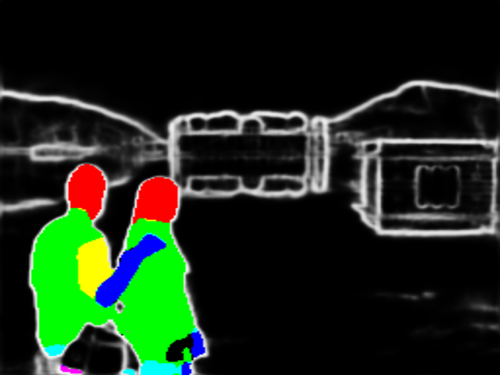}\\
    
    \includegraphics[width=\sz\columnwidth]{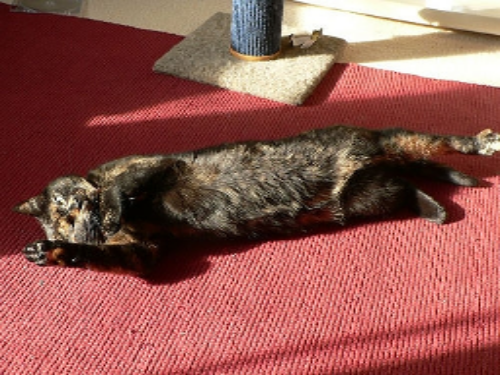} &
    \includegraphics[width=\sz\columnwidth]{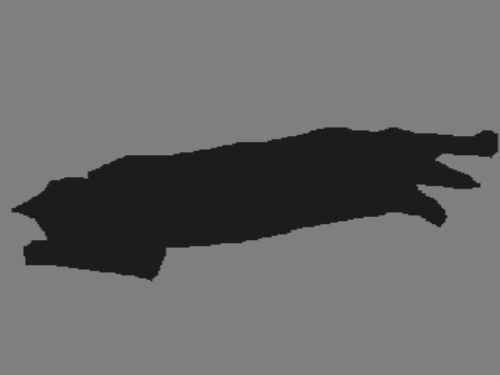} &
    \includegraphics[width=\sz\columnwidth]{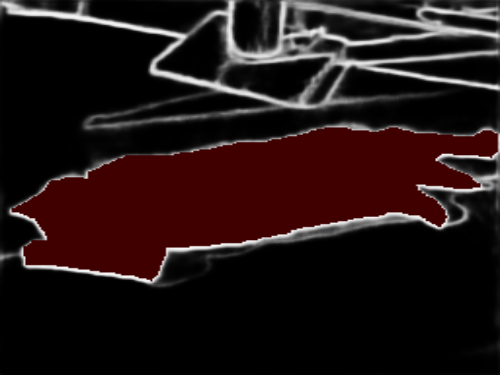} &
    \includegraphics[width=\sz\columnwidth]{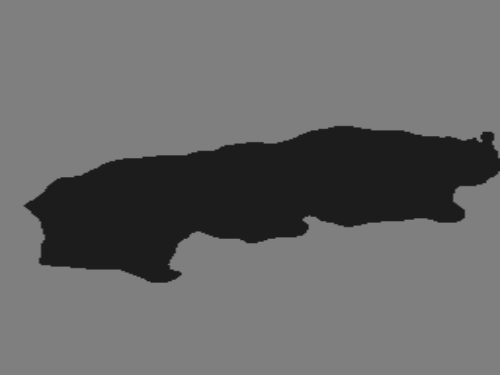} &
    \includegraphics[width=\sz\columnwidth]{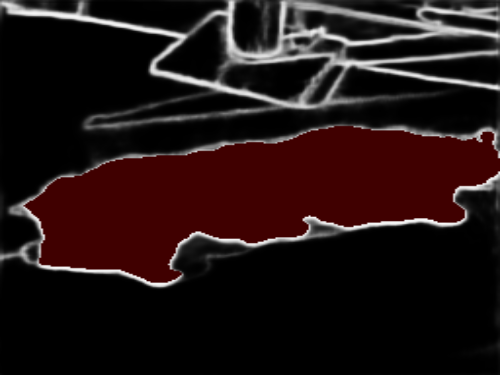}\\
    
    \includegraphics[width=\sz\columnwidth]{images/predicting_task_map/3_img.png} &
    \includegraphics[width=\sz\columnwidth]{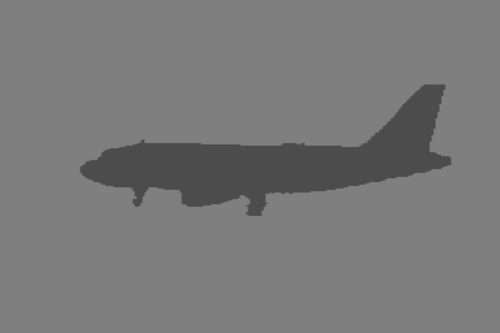} &
    \includegraphics[width=\sz\columnwidth]{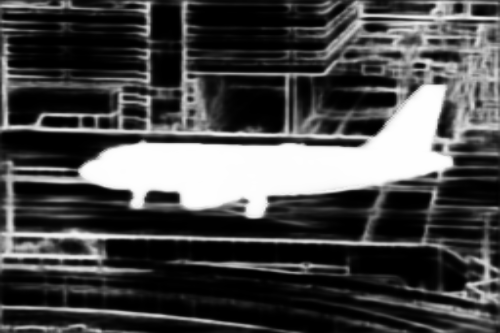} &
    \includegraphics[width=\sz\columnwidth]{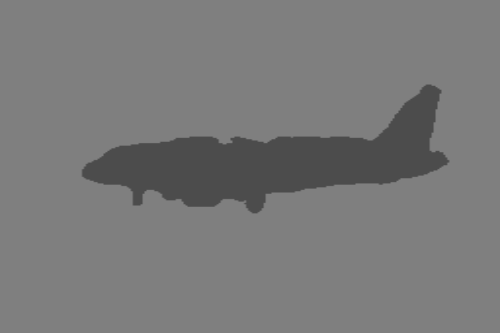} &
    \includegraphics[width=\sz\columnwidth]{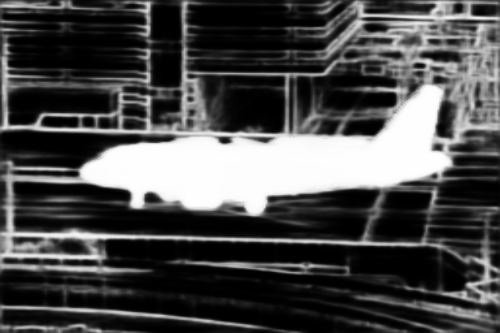}

  \end{tabular}
 \caption{\textbf{Learning what to do where.} CompositeTasking network predictions with the learned Task Palette. A separate network is learned to predict the Task Palette with $75.04 \%$ mIoU.
 More examples can be found in the supplementary materials.
 }
    \label{fig:task_map_prediction}
\end{figure}

\subsection{Task Palette Editing}

The world is striving towards automated processes, but things are not perfect just yet. Data labelling is a very unpleasant and time-consuming job if someone has to make dense annotations. Very often we are interested in labels of different tasks at different spatial locations. For example, we want to predict surface normals of cars so that we can perform realistic re-rendering, at the same time  semantic segmentation of the surrounding objects and edges everywhere else. That is a very clear rule that can be learned by the setup proposed in \ref{sec:learning_where}. This will not be perfect however, and from time to time mistakes will be made. In such scenarios there may be a human in the loop, making sure that all mistakes along the execution pipeline are corrected. Our framework offers to take the predicted Task Palette, where most of it is correct, and edit the mistakes in certain regions. In that setup, most of the work is done automatically, and the human in the loop puts her focus mostly on regions of high interest to the use-case. One such visual example is presented in Figure~\ref{fig:rule_editing}, 
 while more can be found in the supplementary materials.
This highlights further the flexibility of our proposed method.

\begin{figure}[t]
  \centering
  \captionsetup{font=small}
  \scriptsize
  \setlength{\tabcolsep}{1pt}
  \vspace*{2pt}
  \newcommand{\sz}{0.189}           
  \newcommand{\cgs}{\hspace{3pt}}   
  \begin{tabular}{ccccc}
    Image & Task Palette & Prediction & Edited Palette & Prediction \\
    \includegraphics[width=\sz\columnwidth]{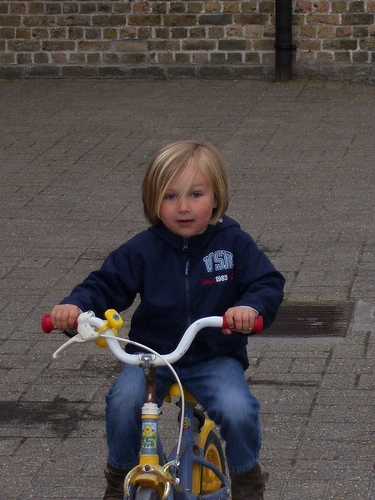} &
    \includegraphics[width=\sz\columnwidth]{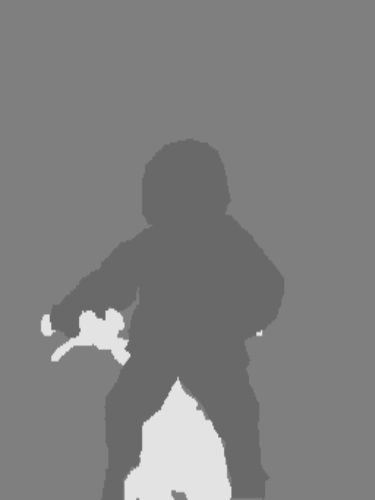} &
    \includegraphics[width=\sz\columnwidth]{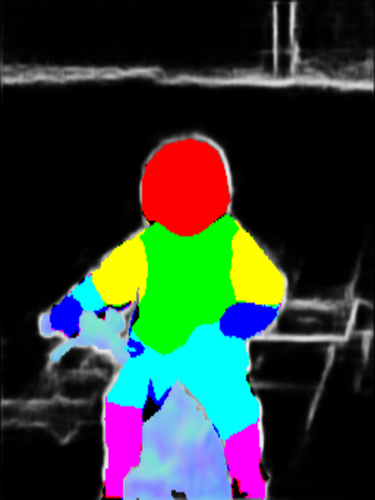} &
    \includegraphics[width=\sz\columnwidth]{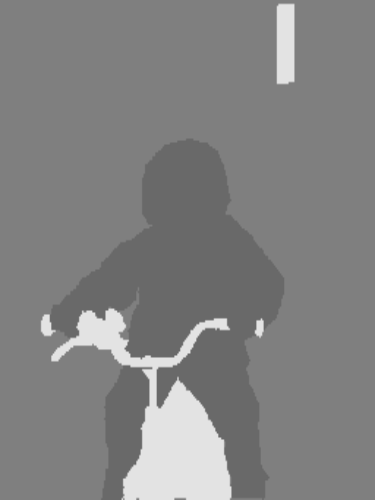} &
    \includegraphics[width=\sz\columnwidth]{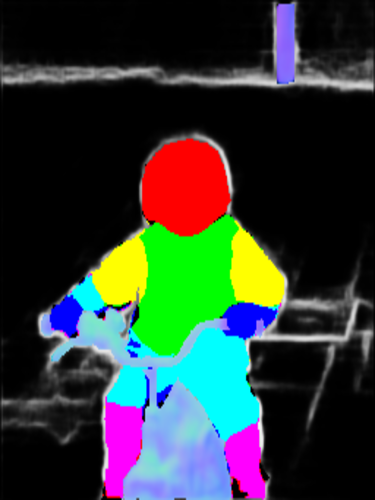}

  \end{tabular}
 \caption{\textbf{Task Palette editing.} The original Task Palette, extracted from the label, did not look satisfactory. A correction was made manually. A prediction of the CompositeTasking network is shown before and after correction. }
    \label{fig:rule_editing}
\end{figure}

\subsection{Breaking the Rule}
``The only constant in life is change'', as Heraclitus once wisely said.
In fact, also prediction tasks are constantly evolving and are subject to change according to use case, context or available resources. One could think of an existing task rule, say $\mathcal{R}_{2}$, needs to be adapted to cater for a changing use case that demands a new rule $\mathcal{R}_{3}$. It can be similar in one way, but different in another compared to $\mathcal{R}_{2}$. For instance, the same tasks are used from $\mathcal{R}_{2}$, but now $\mathcal{R}_{3}$ requires different tasks to be performed in different regions.

One practical example of this is predicting surface normals. Often, the accurate normal labels are obtained by having accurate 3D models. The 3D models however may cover only a part of the scene, therefore of the image. This builds a rule of having normals only for the objects with 3D models. In fact, similar datasets exist. For example, datasets with 3D models of household objects like chairs and tables~\cite{lpt2013ikea}, and of the human body~\cite{Bogo2014FAUST}. Using the model trained on such datasets, one may be interested to predict normals beyond the reason of 3D models. Here we show that our CompositeTasking network can indeed be trained by breaking the rule. 

We break the old rule by simply requesting to execute the task of the new rule. This is then followed by fine-tuning our network on the new rule, if necessary.
As shown in Table \ref{table:breaking_the_rule}, our model trained on $\mathcal{R}_{2}$ is already doing good on a newly introduced rule $\mathcal{R}_{3}$, without any fine-tuning. The rule $\mathcal{R}_{3}$ is somewhat similar to $\mathcal{R}_{2}$, and the model shows the ability to extrapolate on their differences. After fine-tuning it on $\mathcal{R}_{3}$, the performance improves even more, as expected. One example of this is presented in Figure \ref{fig:breaking_the_rule}, 
while more can be found in the supplementary materials.
Interestingly, in Table \ref{table:breaking_the_rule} we observe that the old rule is even improved when training on the new similar rule.

\begin{table}[t]
	\centering\small
    \captionsetup{font=small}
	\caption{{\textbf{Breaking the rule}. Our method can make successful predictions when changing from $\mathcal{R}_{2}$ to a somewhat similar rule $\mathcal{R}_{3}$ (described in the supplementary materials). With fine-tuning on the new rule, the predictions get even better.}}
	\label{table:breaking_the_rule}
	\resizebox{\columnwidth}{!}{
	\setlength\tabcolsep{6pt}
	\renewcommand\arraystretch{1}
	\begin{tabular}{|c|c|c|c|c|c||c|c|c}
		\hline
\rowcolor{mygray}
Testing rule & Training rule & Edge$\uparrow$& Parts$\uparrow$ & Normals$\downarrow$ & Sal$\uparrow$ & $\Delta_m\%$$\downarrow$ \\  \hline 
\multirow{3}{*}{$\mathcal{R}_{3}$} &
$\mathcal{R}_{3}$ & 70.20 & 61.19 & 18.34 & 75.35 & 0.0\% \\  \cline{2-7}
& $\mathcal{R}_{2}$ & 69.70 & 59.41 & 20.11 & 65.21 & -6.68\% \\  \cline{2-7}
& Fine-tuned $\mathcal{R}_{2} \rightarrow \mathcal{R}_{3}$ & 69.70 & 60.91 & 18.68 & 75.00 & -0.87\% \\  \hline  \hline
\multirow{2}{*}{$\mathcal{R}_{2}$}
& $\mathcal{R}_{2}$ & 69.20 & 59.74 & 18.12 & 67.95 & 0.0\%  \\  \cline{2-7} 
& Fine-tuned $\mathcal{R}_{2} \rightarrow \mathcal{R}_{3}$
& 69.40 & 60.84 & 17.95 & 68.31 & +0.90\% \\ \hline 
\end{tabular}
}
\end{table}

\begin{figure}[t]
  \centering
  \captionsetup{font=small}
  \scriptsize
  \setlength{\tabcolsep}{1pt}
  \vspace*{2pt}
  \newcommand{\sz}{0.189}           
  \newcommand{\cgs}{\hspace{3pt}}   
  \begin{tabular}{ccccc}
    Image & Pred $R_{2}$ & Pred $R_{3}$ & Pred $R_{2}$ & Pred $R_{3}$ \\
     & (Trained $R_{2}$) & (Trained $R_{2}$) & (Tuned $R_{2} \rightarrow R_{3}$) & (Tuned $R_{2} \rightarrow R_{3}$) \\
    \includegraphics[width=\sz\columnwidth]{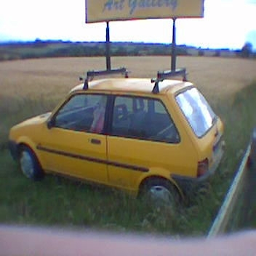} &
    \includegraphics[width=\sz\columnwidth]{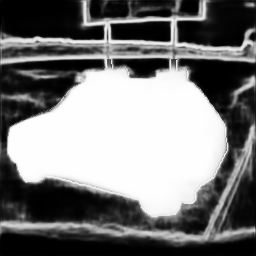} &
    \includegraphics[width=\sz\columnwidth]{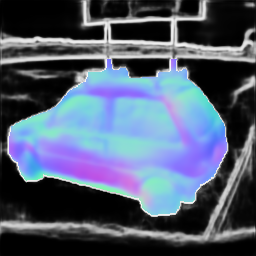} &
    \includegraphics[width=\sz\columnwidth]{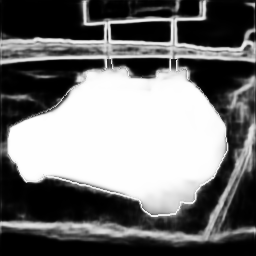} &
    \includegraphics[width=\sz\columnwidth]{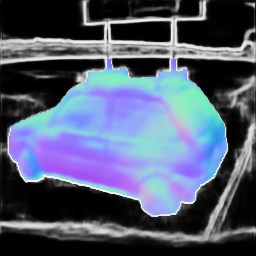}

  \end{tabular}
 \caption{\textbf{Breaking the rule.} Predictions of the CompositeTasking network are show before and after fine-tuning from the old to the new semantic rule. Because the rules are similar in a way, the model can  extrapolate even before fine-tuning. }
    \label{fig:breaking_the_rule}
\end{figure}

\subsection{Random Compositions}

Finally, we are interested to see what happens if our model is evaluated on tasks chosen independently for each pixel at random, denoted as  $\mathcal{R}_{rnd}$. 
Table~\ref{table:rnd_task_map} indicates that our model trained on the mosaic rule $\mathcal{R}_{1r}$ does not perform very well on the random rule $\mathcal{R}_{rnd}$. 
We conjecture this is because the rule $\mathcal{R}_{1r}$ assigns tasks only to large connected regions during training, and no incentive is given to learn the ability to switch tasks with high spatial frequency. Training on the random rule $\mathcal{R}_{rnd}$, consequently improves the performance on $\mathcal{R}_{rnd}$ significantly (Table~\ref{table:rnd_task_map}). A visualization of this results is presented in Figure \ref{fig:rnd_rule_pred}. Interestingly, although only using a single output, we can clearly observe a meaningful execution of different tasks all over the image.  

\begin{table}[t]
	\centering
    \captionsetup{font=small}
	\caption{{\textbf{Evaluating on randomly chosen tasks at each pixel location independently}. Notice the performance difference between training on $\mathcal{R}_{1r}$ vs. $\mathcal{R}_{rnd}$, and testing on $\mathcal{R}_{rnd}$. The edge evaluation is omitted due to the unsuitable evaluation protocol.}}
	\label{table:rnd_task_map}
	\resizebox{\columnwidth}{!}{
	\setlength\tabcolsep{6pt}
	\renewcommand\arraystretch{1}
	\begin{tabular}{|c|c|c|c|c|c|}
		\hline
\rowcolor{mygray}
Trained on rule & Evaluated on rule &  SemSeg $\uparrow$ & Parts$\uparrow$ & Normals$\downarrow$ & Sal$\uparrow$ \\  \hline \hline
$\mathcal{R}_{1r}$ & $\mathcal{S}$ & 62.45 & 52.59 & 16.93 & 67.81  \\  \hline
$\mathcal{R}_{1r}$ & $\mathcal{R}_{rnd}$ & 35.89 & 21.11 & 64.36 & 66.71  \\  \hline 
$\mathcal{R}_{rnd}$ & $\mathcal{R}_{rnd}$ & 59.58 & 52.28 & 17.16 & 67.60   \\  \hline 
$\mathcal{R}_{rnd}$ & $\mathcal{S}$ & 52.26 & 51.88 & 22.65 & 65.34   \\  \hline 
\end{tabular}
}
\end{table}

\begin{figure}[t]
  \centering
  \scriptsize
  \setlength{\tabcolsep}{1pt}
  \vspace*{2pt}
  \newcommand{\sz}{0.3}           
  \newcommand{\cgs}{\hspace{3pt}}   
  \begin{tabular}{ccc}
    \textbf{Image} & \textbf{Task Palette} & \textbf{Prediction}\\
    \includegraphics[width=\sz\columnwidth]{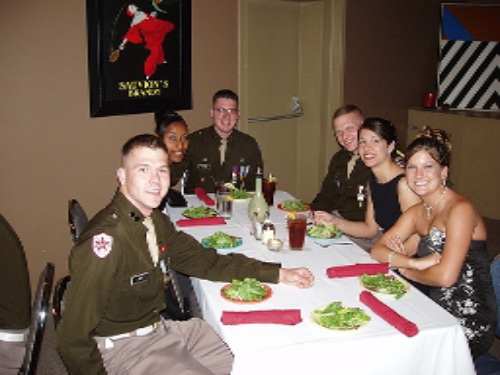} &
    \includegraphics[width=\sz\columnwidth]{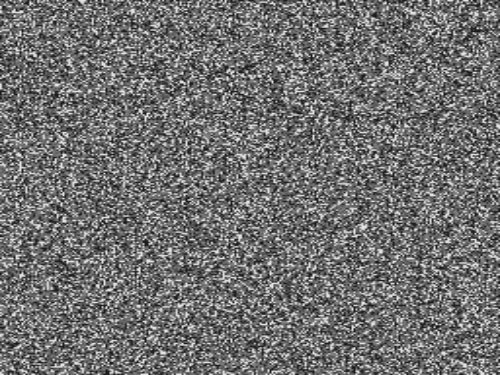} &
    \includegraphics[width=\sz\columnwidth]{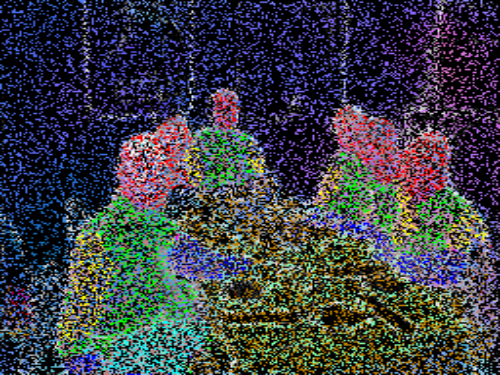}

  \end{tabular}
 \caption{\textbf{Random tasks chosen at each pixel independently.} Predictions includes all five tasks of rule $\mathcal{R}_{rnd}$.}
    \label{fig:rnd_rule_pred}
\end{figure}

\section{Discussion}

We feel that this is only the tip of the iceberg. In \cite{mittal2020emoticon}, for example,  it is crucial to fuse different tasks like face detection, pose estimation, scene understanding and depth estimation to obtain state-of-the-art performance for the high-level task of emotion recognition. Many state-of-the-art pipelines for high-level tasks share that approach \footnote{Andrej Karpathy, senior director of AI at Tesla, \href{https://www.youtube.com/watch?v=oBklltKXtDE&ab_channel=PyTorch}{recently said} that they use a multi-tasking system with $48$ shared backbones and $1000$ different output task heads in their self-driving Autopilot high-level task} and more often than not, different predicted tasks feel important only at different spatial regions of the input image. A great potential is seen for CompositeTasking here. This compactness in terms of memory and computation efficiency can sometimes determine whether some solution to the problem can be practically implemented or not. Also, wasting resources when there is no need for it is never welcome. 

While supervising the high-level predictions, one can also attempt to learn the rule of what is beneficial to predict where, even if such a rule is not known a priori. This can bring a new level of understanding how the very complex Deep Learning models make decisions on high-level tasks, by observing the requests that the network is making during inference.
\section{Conclusion}
In this work, we introduced the concept of CompositeTasking as the fusion of multiple, spatially distributed tasks, motivated by the frequent availability of only sparse labels across tasks, and the desire for a compact multi-tasking network. To this end, we studied a novel task conditioning model -- a single encoder-decoder network that performs multiple, spatially varying tasks at once. We showed that CompositeTasking offers efficient multi-task learning from only sparse supervision, with performance competitive to dense supervision and a multi-headed multi-tasking design. Moreover, we demonstrated the unique flexibility by our approach with regards to interactive task editing, and rules transformations.

\clearpage

\appendix
\title{ \Large \bf
CompositeTasking: Understanding Images by Spatial Composition of Tasks 
Supplementary material
}
\maketitle

\section{Supplementary Structure}

In Section \ref{sec:architecture} of this supplementary material, we discuss more details about our proposed network's architecture and we provide a diagram with all it's details. In Section \ref{sec:rules} we provide more details about some Task Palette rules. In Section \ref{sec:visual_results} we provide additional visual results of our method. In Section \ref{sec:implementation_details_supp} we provide more details on implementational details and hyperparameters used to conduct experiments. In Section \ref{sec:more_exp} we provide further analysis, mostly in depth details about the hyperparameter choices. Finally, in Section \ref{sec:limitations} we discuss some limitations of our proposed method.
\section{Architecture}
\label{sec:architecture}
 
\begin{figure*}[t]
  \centering
  \captionsetup{font=small}
      \includegraphics[width=0.95\linewidth]{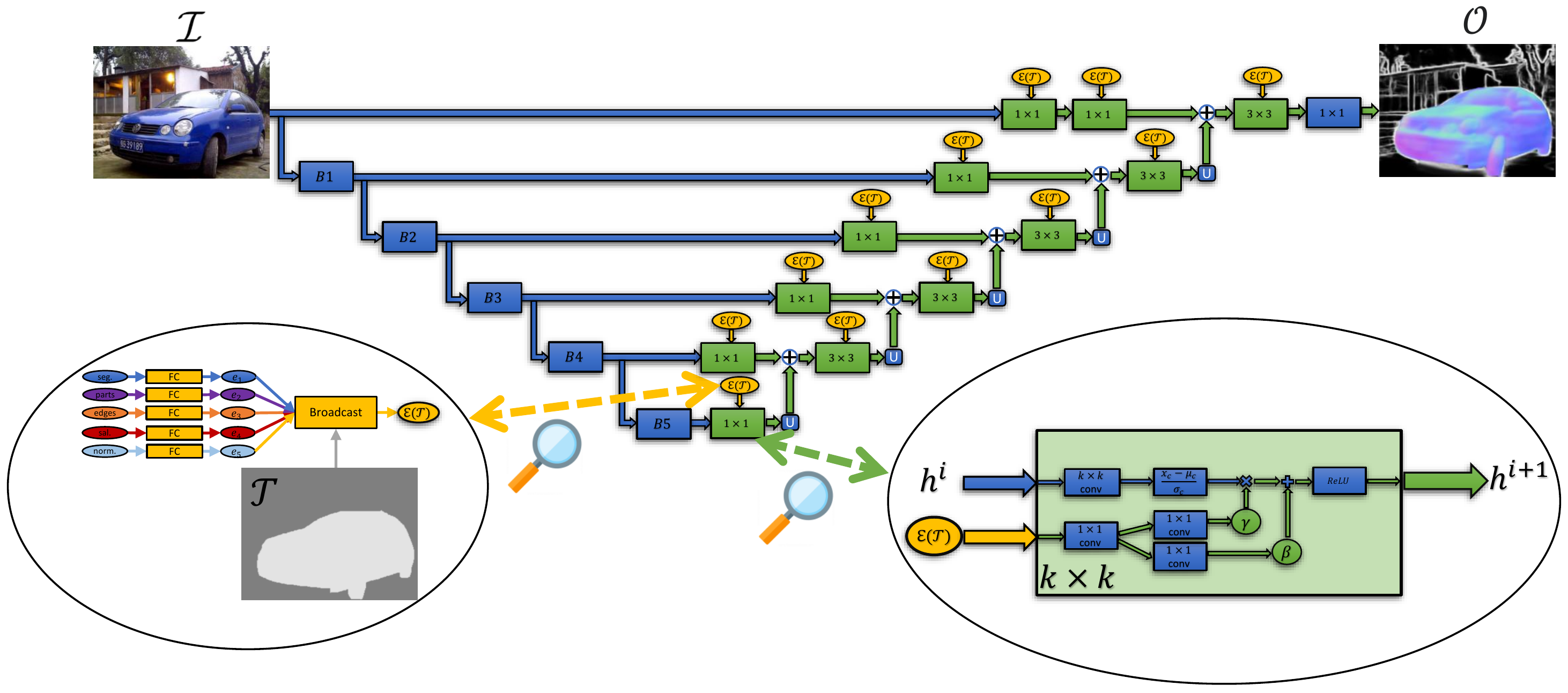}
\caption{{\textbf{CompositeTasking network architecture}. \textcolor{blue}{Blue} blocks process features unconditionally, while the \textcolor{green}{green} blocks process features in a spatial task-conditioning manner. The network is composed out of an \textcolor{blue}{encoder} and \textcolor{green}{decoder}. The \textcolor{green}{decoder} is composed only out of task composition blocks and upsampling operations. The task composition block is depicted in the lower-right corner of this diagram, while the upsampling operations are denoted with the blue rectangle containing the letter "U". The task composition block takes in a Task Palette embedding $\mathcal{E}(\mathcal{T})$, which is obtained using the \textcolor{gold}{task representation block}, depicted in the lower-left part of this diagram.}}
\label{fig:full_composite_network}
\end{figure*}

The full CompositeTasking network diagram is presented in Figure~\ref{fig:full_composite_network}. The encoder is the well known ResNet$34$ network. 
The choice for ResNet$34$ trades-off between performance and memory/computational demands.
In the diagram, the encoder is divided into $5$ blocks: B$1$,...,B$5$. Each block B$i$ is a part of the encoder that takes features of spatial size $(\frac{H}{2^{i-1}}, \frac{W}{2^{i-1}})$, and reduces the spatial size to $(\frac{H}{2^i}, \frac{W}{2^i})$. In the case of ResNet$34$, the spatial size is reduced with a strided convolution.

The Task Palette Embedding $\mathcal{E}(\mathcal{T})$ is calculated once for each forward pass, with the task representation block from Figure~\ref{fig:full_composite_network}. Since all the task composition blocks use the same embedding $\mathcal{E}(\mathcal{T})$, a spatial pyramid of $\mathcal{E}(\mathcal{T})$ is created to be available in all spatial sizes $(\frac{H}{2^i}, \frac{W}{2^i})$, where $i=0,1,...,5$.
\section{Task Palette Rules}
\label{sec:rules}

\noindent\textbf{Random mosaic rule $\mathcal{R}_{1r}$}: The image is spatially divided into four rectangles by intersecting a vertical and horizontal line through a point $\mathbf{c}=(c_x, c_y)$. The point is chosen randomly as $c_x \sim U[\frac{W}{4}, 3\frac{W}{4}]$, $c_y \sim U[\frac{H}{4}, 3\frac{W}{4}]$. Each region $r \in \{a,b,c,d\}$ receives its task $k_r$. In other words:
    \begin{equation}
        t_{yx} = 
        \begin{cases}
            k_a,\,  \text{for} \, x \leq c_x, y \leq c_y \\
            k_b,\,  \text{for} \, x > c_x, y \leq c_y \\
            k_c,\,  \text{for} \, x \leq, y > c_y \\
            k_d,\,  \text{for} \, x > c_x, y > c_y 
        \end{cases}
    \end{equation}
The specific tasks $k_a, ... k_d$, as well as $\mathbf{c}$, can be changed every time the Task Palette is requested from this rule.

\noindent\textbf{The rule $\mathcal{R}_2$ requests}:
\begin{itemize}
    \item \textbf{Surface normals} on pixels which belong to the semantic classes: bottle, chair, dining table, potted plant, sofa and tv monitor. In other word, this rule requests surface normals on common \emph{household objects} from the dataset.
    \item \textbf{Human body parts} on pixels which belong to \emph{humans}.
    \item \textbf{Segmentation} on pixels which belong to birds, horses, cows, cats, dogs and sheep. In other words, this rule requests segmentation on \emph{animals} from the dataset.
    \item \textbf{Saliency} on aeroplanes, bicycles, boats, buses, cars, motorbikes and trains. In other words, this rule requests saliency on \emph{vehicles} from the dataset.
    \item \textbf{Edges} everywhere else.
\end{itemize}

\noindent\textbf{The rule $\mathcal{R}_3$ requests}:
\begin{itemize}
    \item \textbf{Surface normals} on pixels which belong to chairs, dining tables, sofas, bicycles, buses, cars, motorbikes and trains. In other words, it requests surface normals on \emph{some vehicles} and \emph{some common household objects} from the dataset.
    \item \textbf{Human body parts} on pixels which belong to \emph{humans}.
    \item \textbf{Saliency} on aeroplanes, boats, birds, horses, cows, cats, dogs, sheep, bottles, potted plants and tv monitors. In other words, it requests saliency on \emph{other vehicles}, \emph{other household objects} and \emph{animals} from the dataset.
    \item \textbf{Edges} everywhere else.
\end{itemize}

The rule $\mathcal{R}_3$ is designed to be similar to rule $\mathcal{R}_2$. It uses all the same tasks as rule $\mathcal{R}_2$, except for semantic segmentation, which has been left out. The tasks of predicting edges and human body parts are requested at the exact same locations. The tasks of surface normals, and saliency are requested on a few classes where they are already requested in rule $\mathcal{R}_2$, but mostly on different classes.
\section{Additional Qualitative Results}
\label{sec:visual_results}

Here we provide more visual examples of the CompositeTasking network's capabilities

\begin{figure*}[t]
  \centering
  \captionsetup{font=small}
  \scriptsize
  \setlength{\tabcolsep}{1pt}
  \vspace*{-2pt}
  \newcommand{\sz}{0.189}           
  \newcommand{\cgs}{\hspace{3pt}}   
  \begin{tabular}{ccccc}
    \multirow{1}{*}[15pt]{\rotatebox{90}{\textbf{Image}}}
    \includegraphics[width=\sz\linewidth]{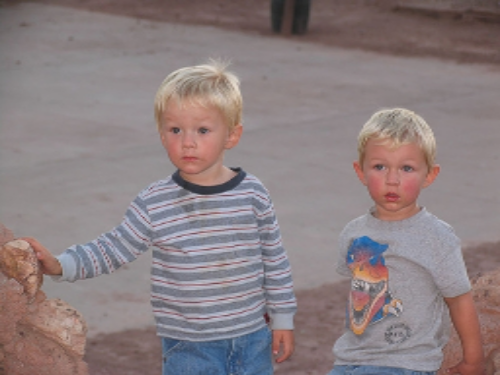} &
    \includegraphics[width=\sz\linewidth]{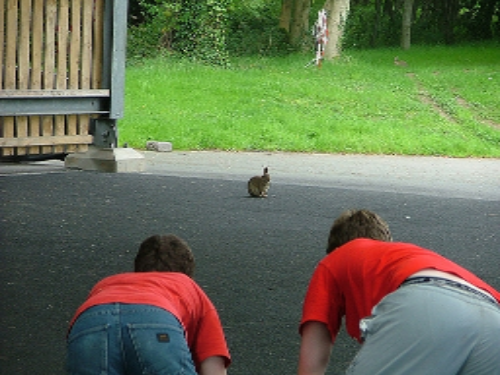} &
    \includegraphics[width=\sz\linewidth]{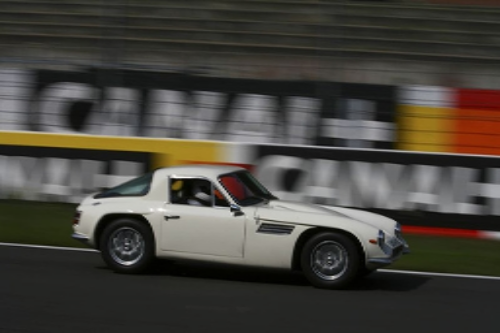} &
    \includegraphics[width=\sz\linewidth]{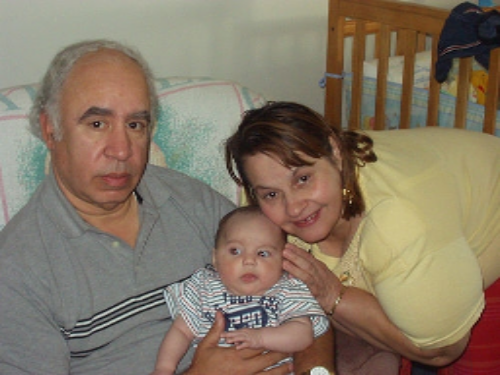} &
    \includegraphics[width=\sz\linewidth]{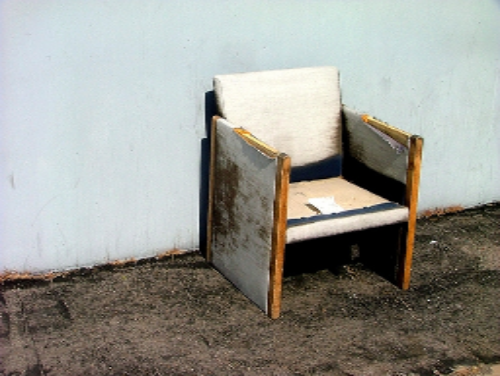} \\
    
    \multirow{1}{*}[15pt]{\rotatebox{90}{\textbf{Pred}}}
    \includegraphics[width=\sz\linewidth]{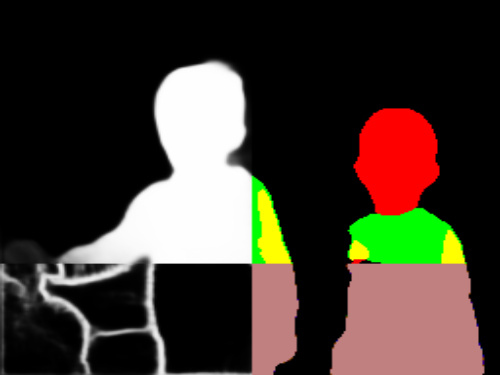} &
    \includegraphics[width=\sz\linewidth]{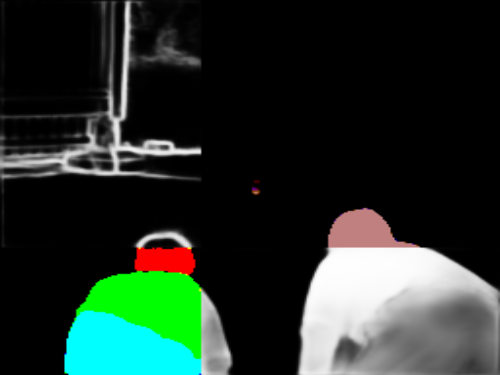} &
    \includegraphics[width=\sz\linewidth]{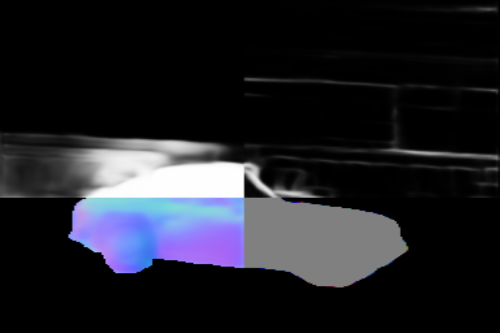} &
    \includegraphics[width=\sz\linewidth]{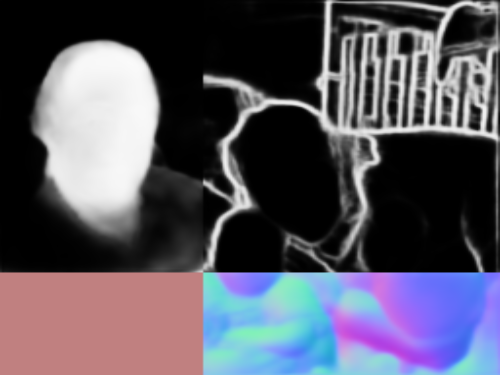} &
    \includegraphics[width=\sz\linewidth]{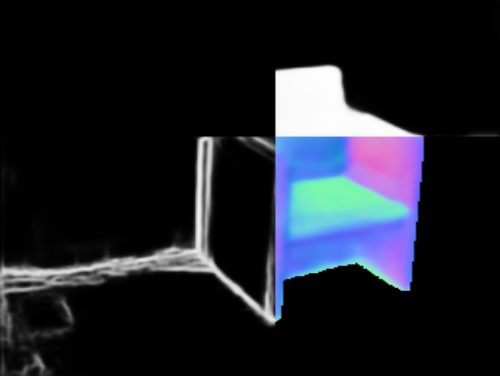}\\
    
    \multirow{1}{*}[15pt]{\rotatebox{90}{\textbf{Image}}}
    \includegraphics[width=\sz\linewidth]{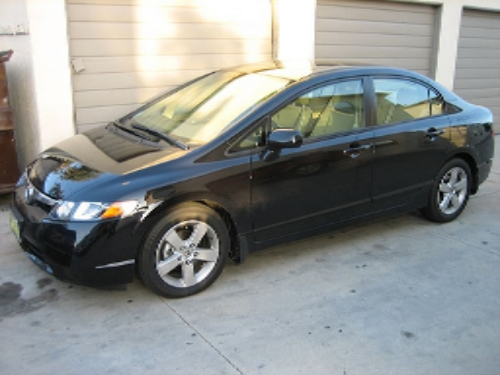} &
    \includegraphics[width=\sz\linewidth]{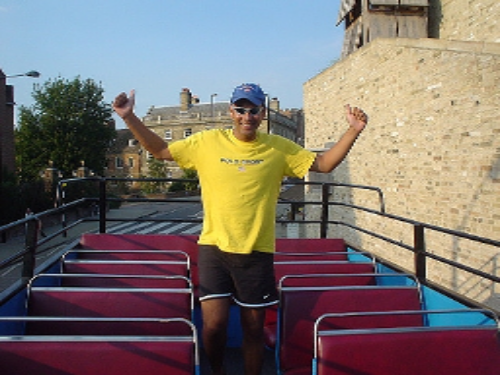} &
    \includegraphics[width=\sz\linewidth]{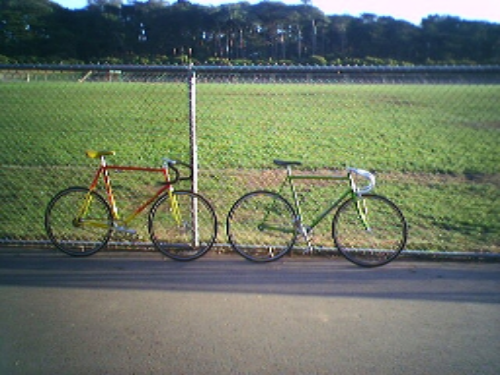} &
    \includegraphics[width=\sz\linewidth]{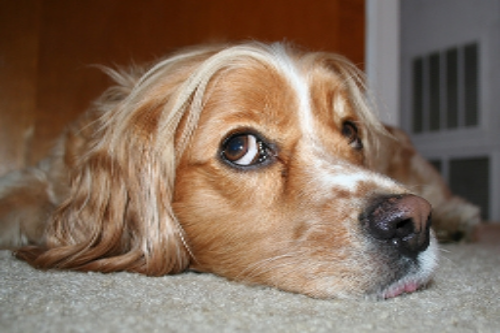} &
    \includegraphics[width=\sz\linewidth]{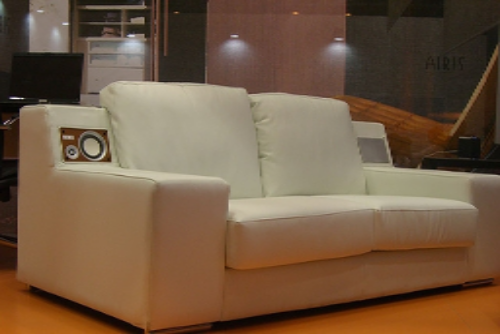} \\
    
    \multirow{1}{*}[15pt]{\rotatebox{90}{\textbf{Pred}}}
    \includegraphics[width=\sz\linewidth]{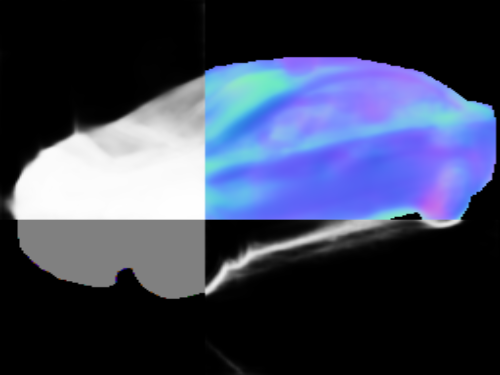} &
    \includegraphics[width=\sz\linewidth]{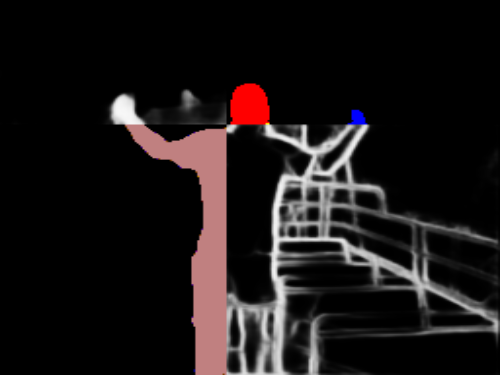} &
    \includegraphics[width=\sz\linewidth]{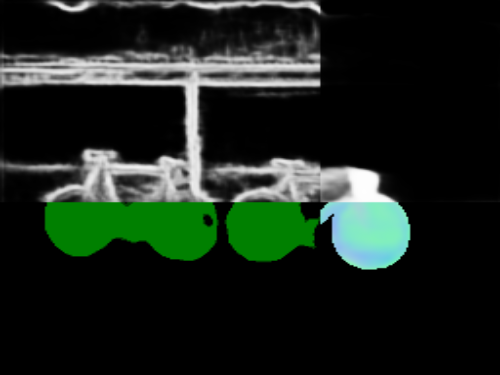} &
    \includegraphics[width=\sz\linewidth]{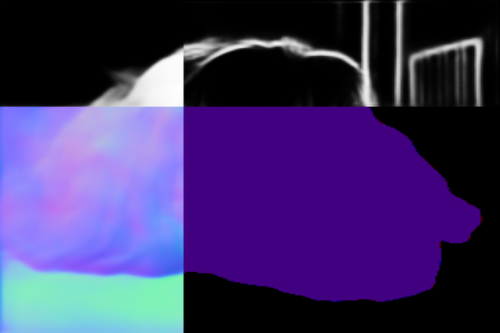} &
    \includegraphics[width=\sz\linewidth]{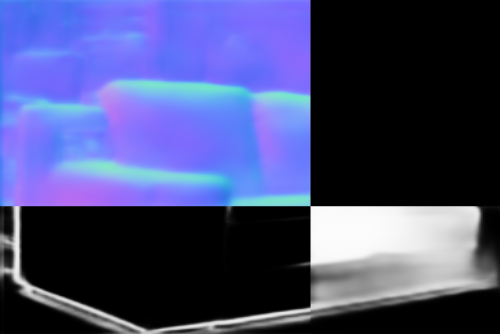} 

  \end{tabular}
 \caption{\textbf{Random mosaic compositions.} CompositeTasking network predictions on requests with the $\mathcal{R}_{1r}$ rule.}
    \label{fig:mosaic_prediction_supp}
\end{figure*}

\noindent\textbf{Random mosaics.} Additional visual examples from the experiment presented in Figure~\ref{fig:mosaic_prediction} are presented in Figure~\ref{fig:mosaic_prediction_supp}. We can see that the network successfully predicts different tasks simultaneously at different pixel locations, all during the same forward pass. It shows the ability to sharply change the task which is being predicted in different spatial locations, with respect to the Task Palette $\mathcal{T}$.

\begin{figure*}[t]
  \centering
  \captionsetup{font=small}
  \scriptsize
  \setlength{\tabcolsep}{1pt}
  \vspace*{-2pt}
  \newcommand{\sz}{0.1575}           
  \newcommand{\cgs}{\hspace{3pt}}   
  \begin{tabular}{cccccc}
    \textbf{Image} & \textbf{Edges} & \textbf{Segmentation} & \textbf{Human parts} & \textbf{Surface normals} & \textbf{Saliency} \\
    \includegraphics[width=\sz\linewidth]{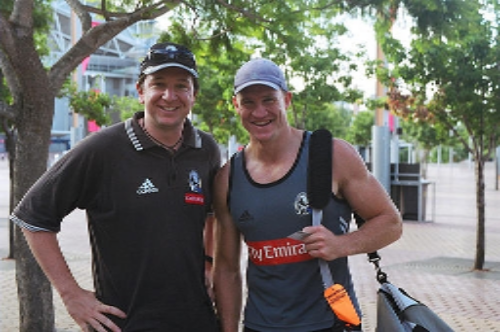} &
    \includegraphics[width=\sz\linewidth]{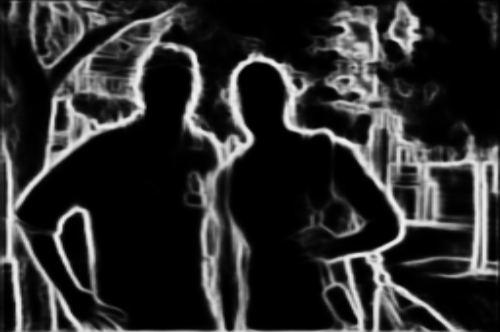} &
    \includegraphics[width=\sz\linewidth]{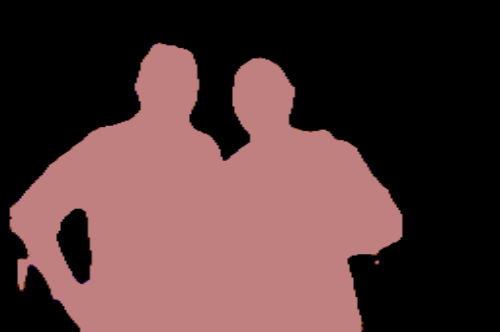} &
    \includegraphics[width=\sz\linewidth]{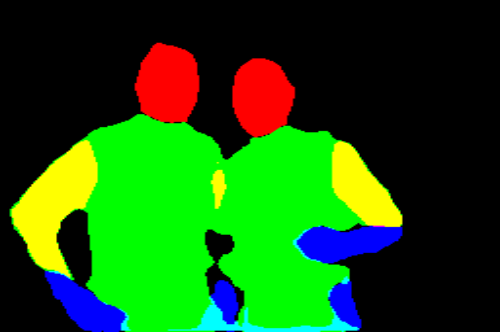} &
    \includegraphics[width=\sz\linewidth]{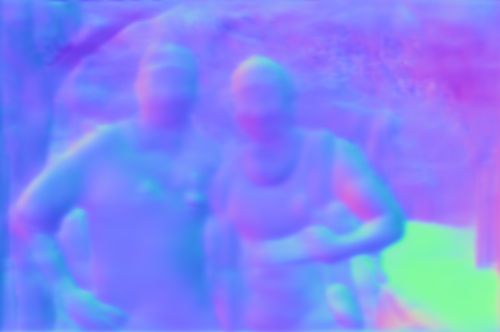} &
    \includegraphics[width=\sz\linewidth]{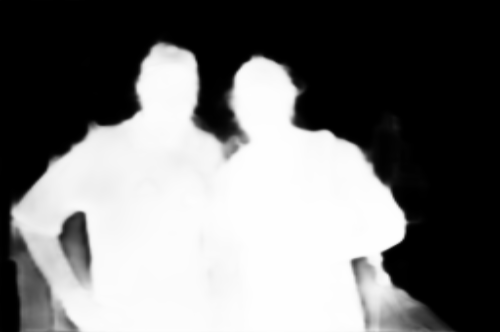} \\ 
        
    \includegraphics[width=\sz\linewidth]{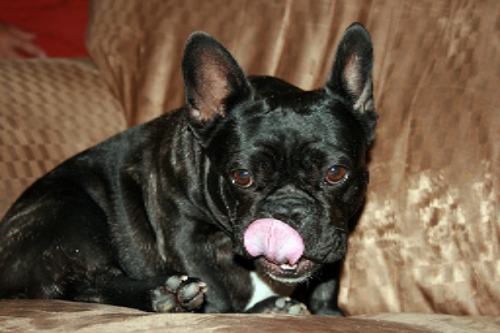} &
    \includegraphics[width=\sz\linewidth]{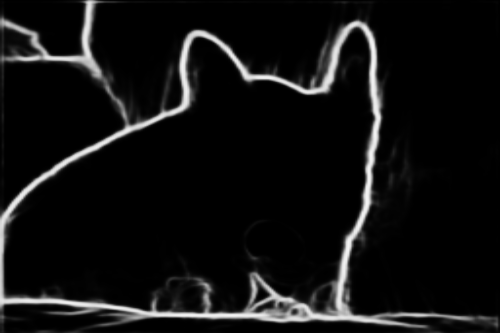} &
    \includegraphics[width=\sz\linewidth]{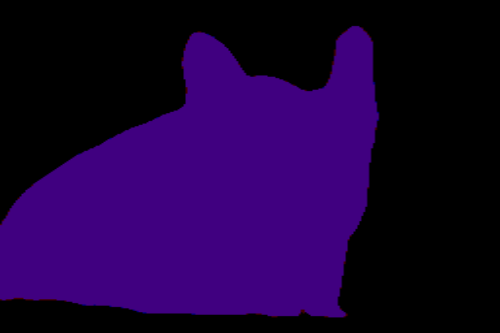} &
    \includegraphics[width=\sz\linewidth]{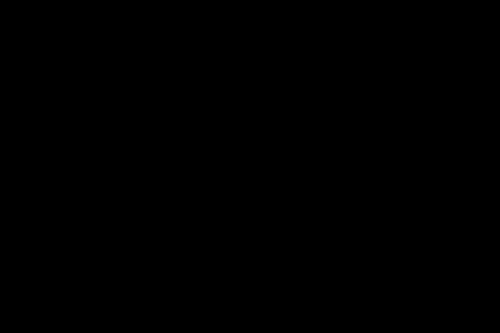} &
    \includegraphics[width=\sz\linewidth]{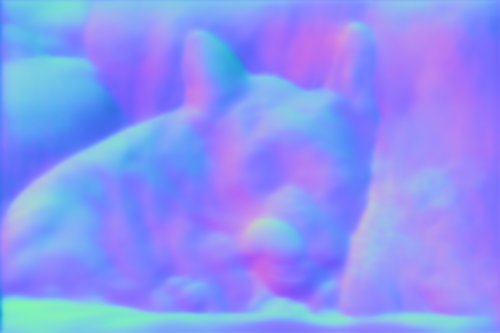} &
    \includegraphics[width=\sz\linewidth]{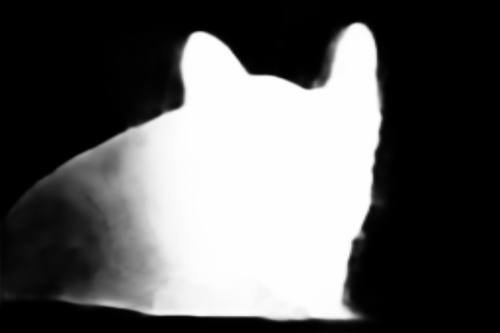} \\
        
    \includegraphics[width=\sz\linewidth]{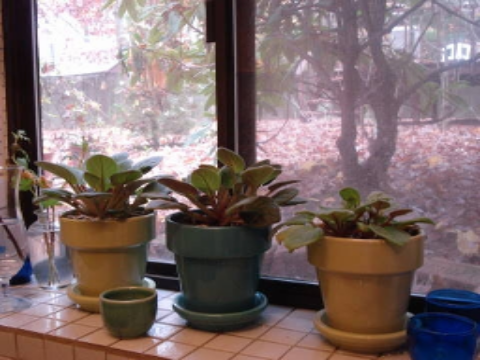} &
    \includegraphics[width=\sz\linewidth]{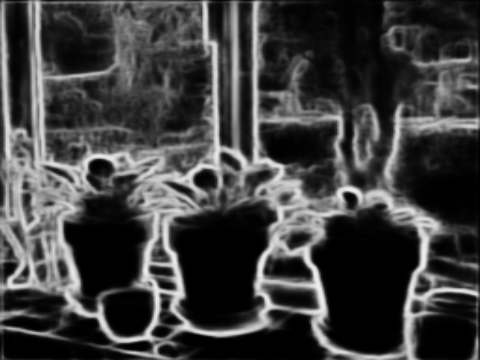} &
    \includegraphics[width=\sz\linewidth]{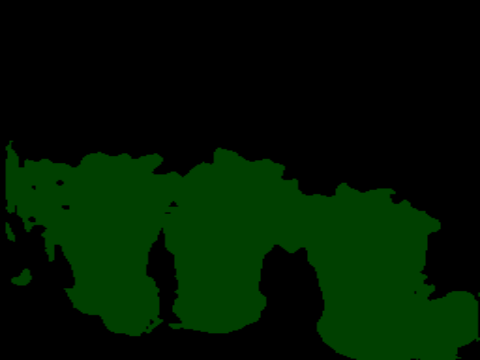} &
    \includegraphics[width=\sz\linewidth]{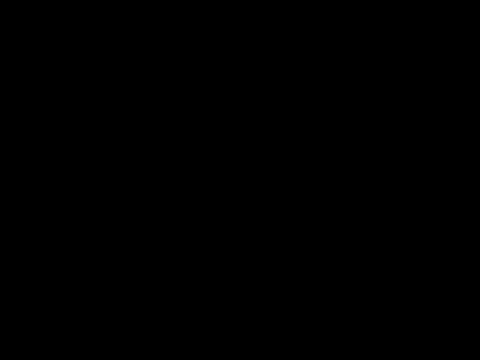} &
    \includegraphics[width=\sz\linewidth]{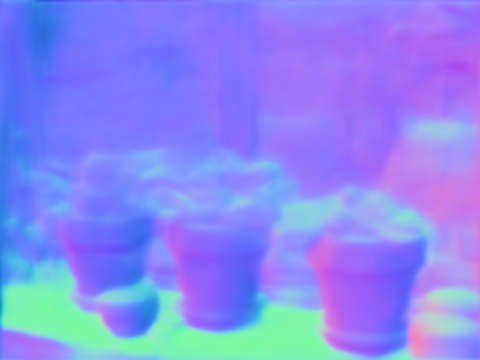} &
    \includegraphics[width=\sz\linewidth]{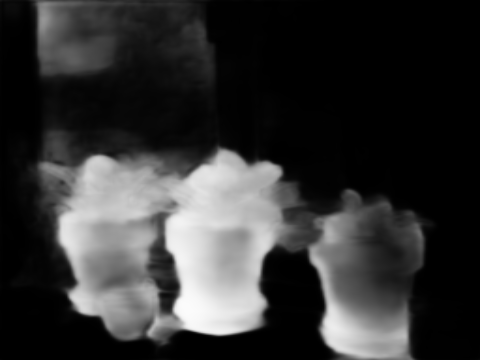} \\     
    
    \includegraphics[width=\sz\linewidth]{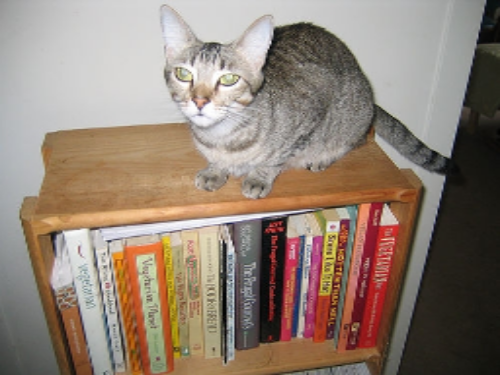} &
    \includegraphics[width=\sz\linewidth]{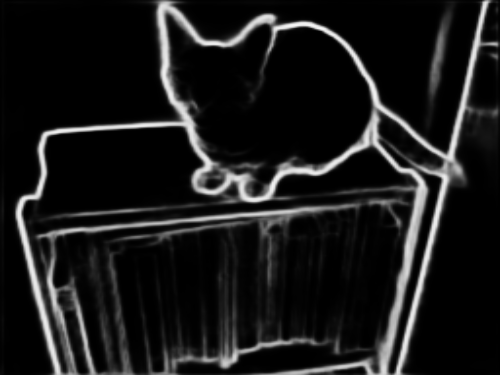} &
    \includegraphics[width=\sz\linewidth]{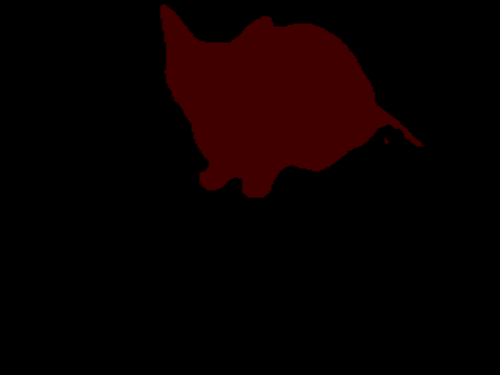} &
    \includegraphics[width=\sz\linewidth]{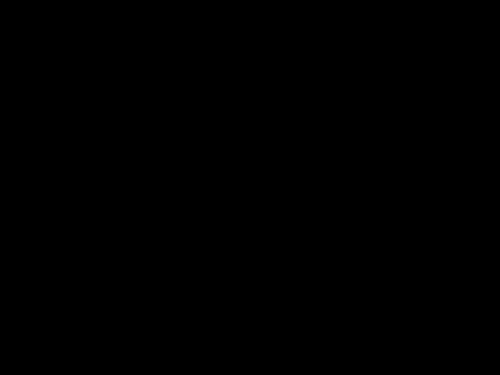} &
    \includegraphics[width=\sz\linewidth]{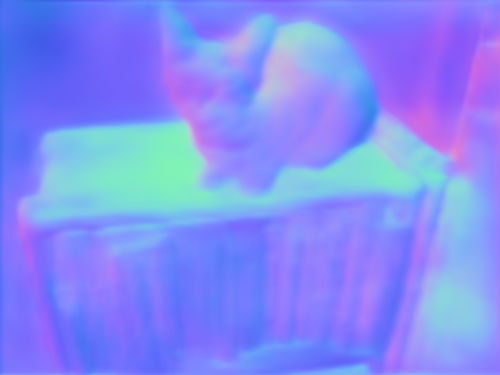} &
    \includegraphics[width=\sz\linewidth]{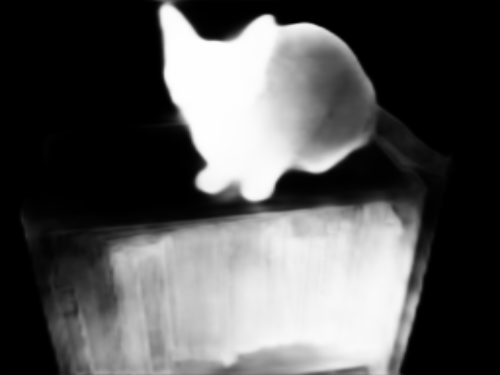} \\
    
    \includegraphics[width=\sz\linewidth]{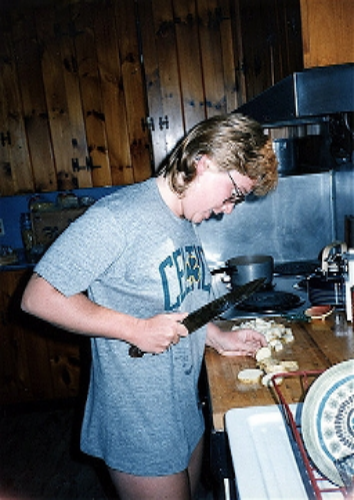} &
    \includegraphics[width=\sz\linewidth]{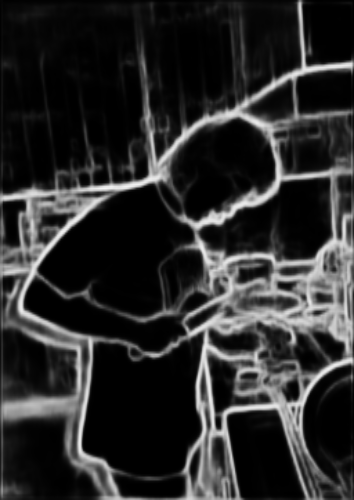} &
    \includegraphics[width=\sz\linewidth]{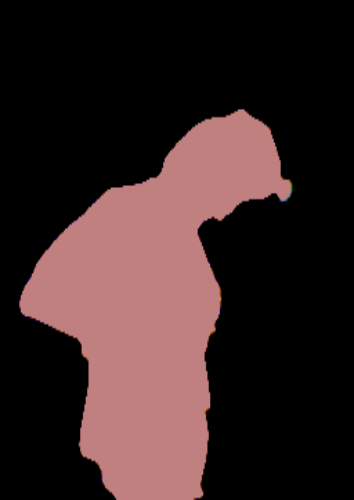} &
    \includegraphics[width=\sz\linewidth]{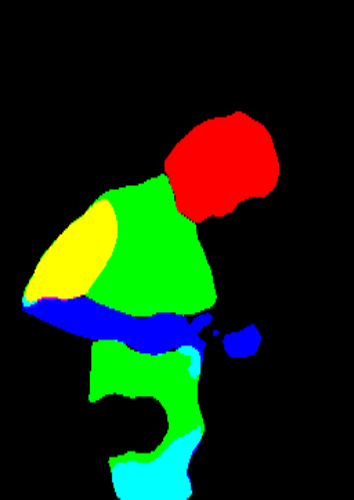} &
    \includegraphics[width=\sz\linewidth]{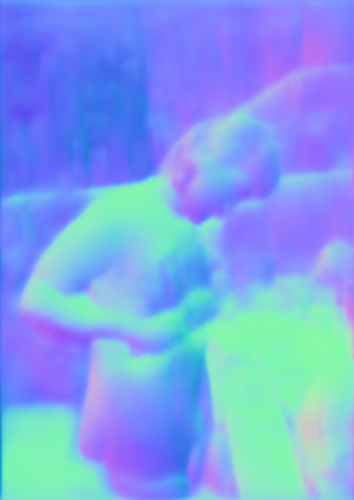} &
    \includegraphics[width=\sz\linewidth]{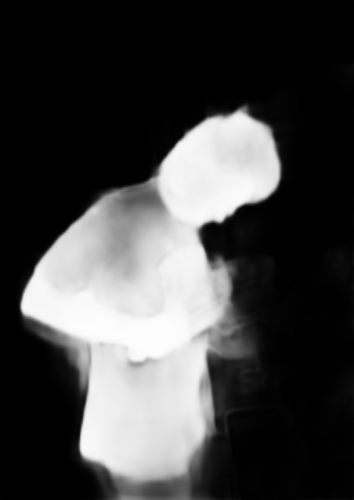} \\
        
    \includegraphics[width=\sz\linewidth]{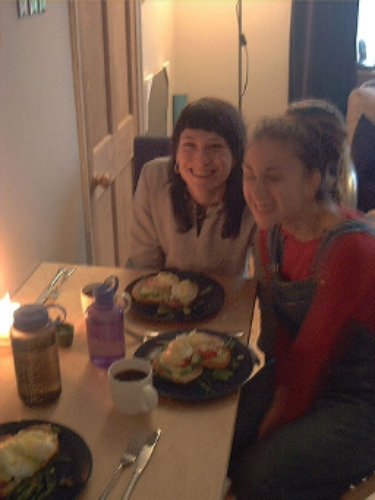} &
    \includegraphics[width=\sz\linewidth]{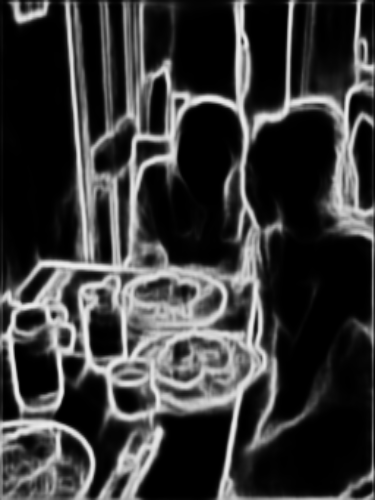} &
    \includegraphics[width=\sz\linewidth]{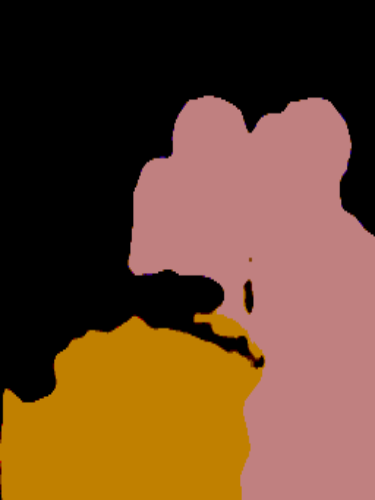} &
    \includegraphics[width=\sz\linewidth]{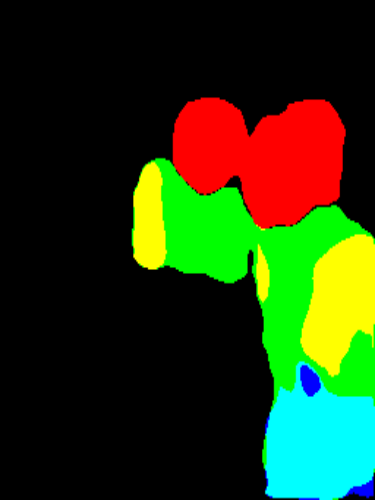} &
    \includegraphics[width=\sz\linewidth]{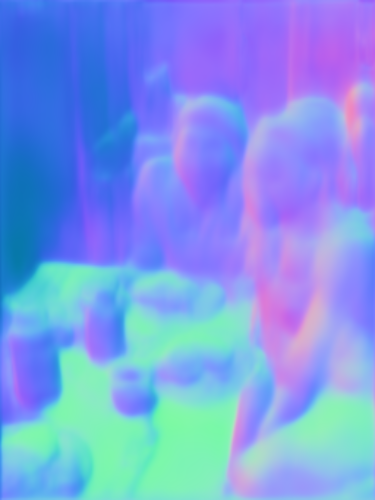} &
    \includegraphics[width=\sz\linewidth]{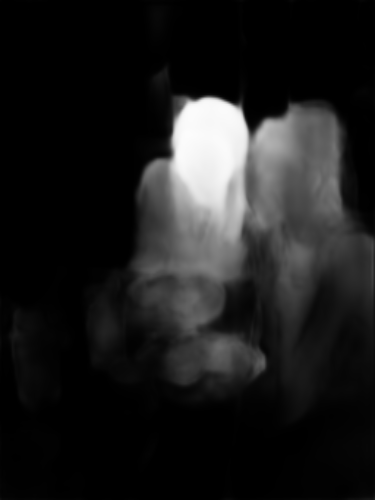}  \\
        
    \includegraphics[width=\sz\linewidth]{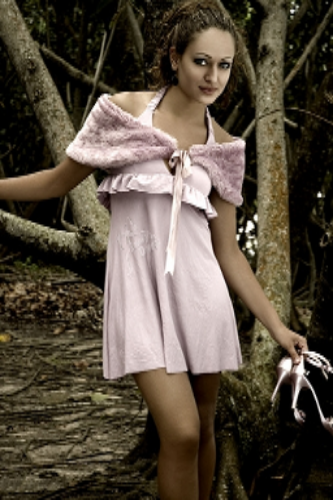} &
    \includegraphics[width=\sz\linewidth]{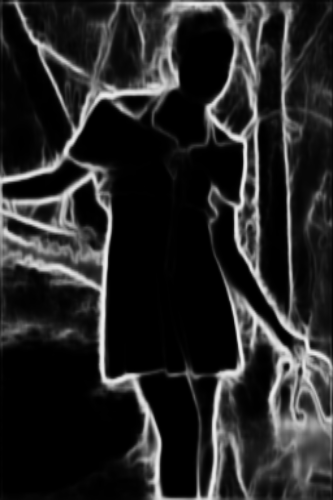} &
    \includegraphics[width=\sz\linewidth]{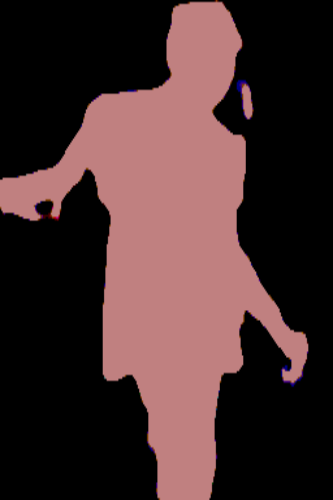} &
    \includegraphics[width=\sz\linewidth]{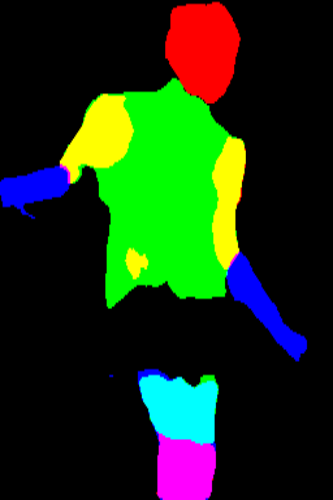} &
    \includegraphics[width=\sz\linewidth]{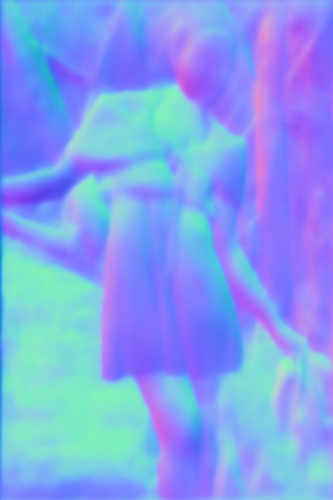} &
    \includegraphics[width=\sz\linewidth]{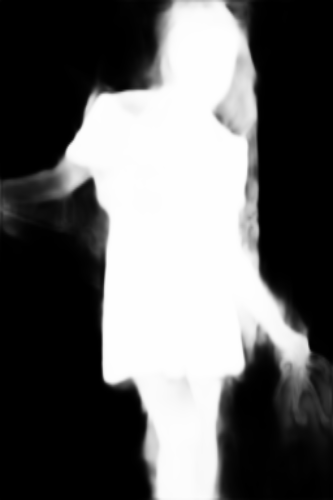}

  \end{tabular}
 \caption{\textbf{Single task predictions.} Even though our model is made for CompositeTasking, it can also make predictions on requests with the $\mathcal{S}$ rule. }
    \label{fig:st_predictions_supp}
\end{figure*}

\noindent\textbf{Single task predictions.} Additional visual examples from the experiments presented in Figure~\ref{fig:st_predictions} are presented in Figure~\ref{fig:st_predictions_supp}. 
Although the network is designed to be used in a CompositeTasking fashion, here we see that our network can make all task predictions successfully.
In this case, images are encoded only once followed by multiple task-specific decodings.

\begin{figure*}[t]
  \centering
  \captionsetup{font=small}
  \scriptsize
  \setlength{\tabcolsep}{1pt}
  \vspace*{-2pt}
  \newcommand{\sz}{0.189}           
  \newcommand{\cgs}{\hspace{3pt}}   
  \begin{tabular}{ccccc}
    \textbf{Image} & \textbf{Task palette} & \textbf{Prediction} & \textbf{Predicted Palette} & \textbf{Prediction} \\    
    \includegraphics[width=\sz\linewidth]{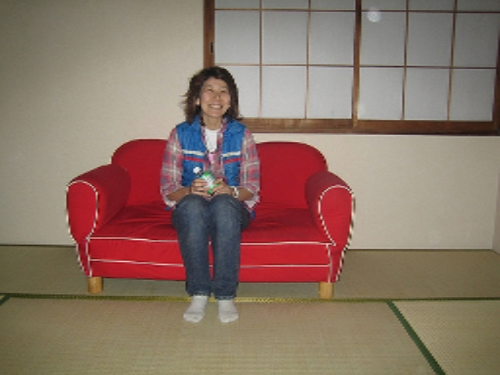} &
    \includegraphics[width=\sz\linewidth]{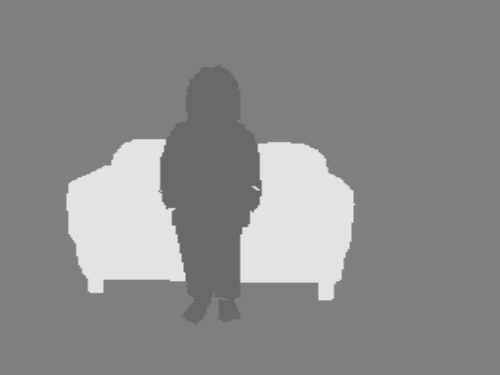} &
    \includegraphics[width=\sz\linewidth]{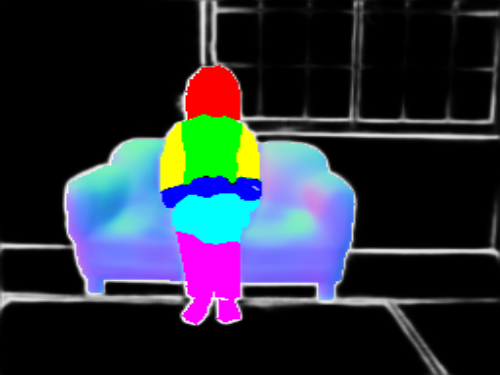} &
    \includegraphics[width=\sz\linewidth]{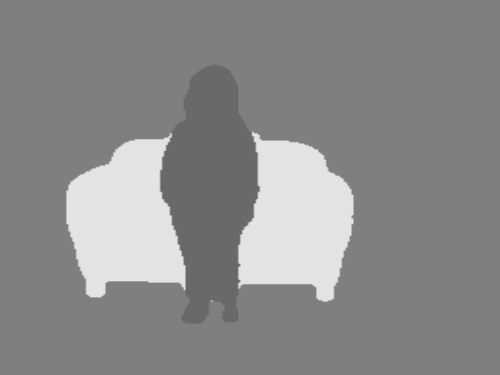} &
    \includegraphics[width=\sz\linewidth]{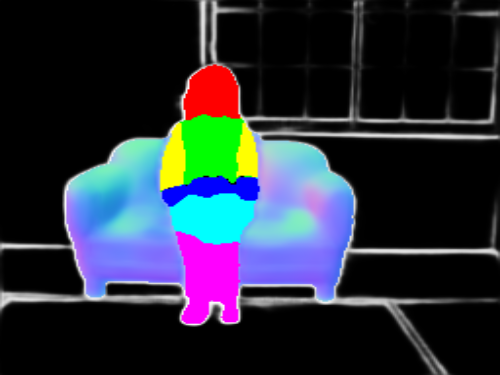}\\
    
    \includegraphics[width=\sz\linewidth]{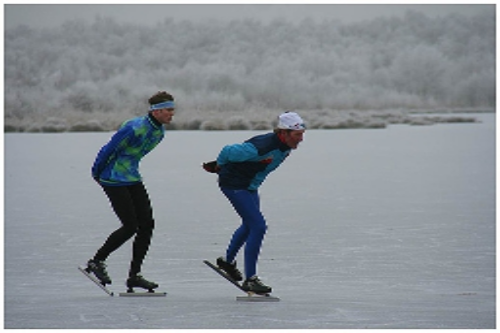} &
    \includegraphics[width=\sz\linewidth]{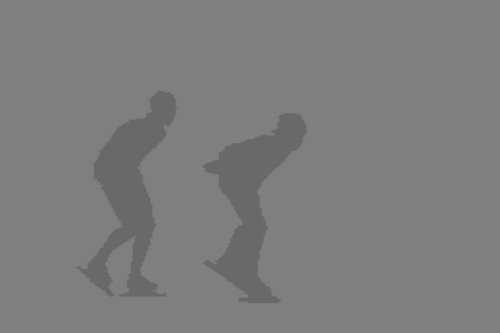} &
    \includegraphics[width=\sz\linewidth]{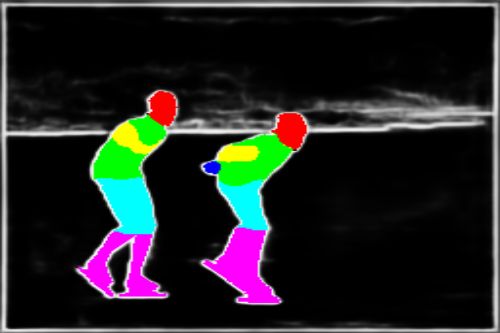} &
    \includegraphics[width=\sz\linewidth]{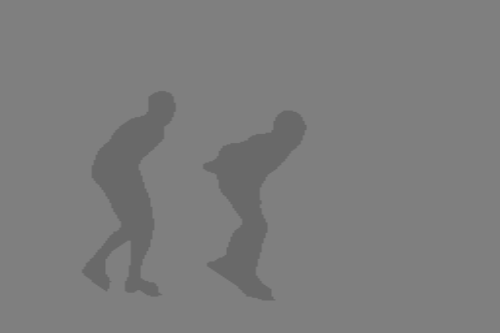} &
    \includegraphics[width=\sz\linewidth]{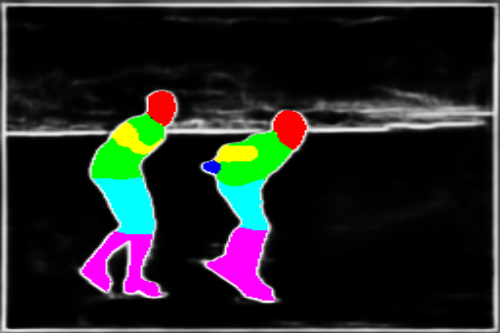}\\

    \includegraphics[width=\sz\linewidth]{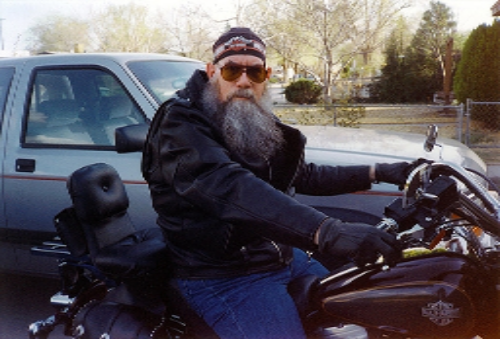} &
    \includegraphics[width=\sz\linewidth]{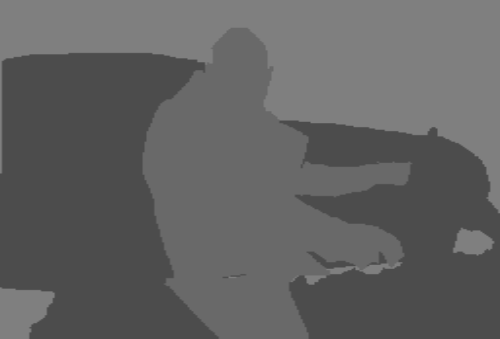} &
    \includegraphics[width=\sz\linewidth]{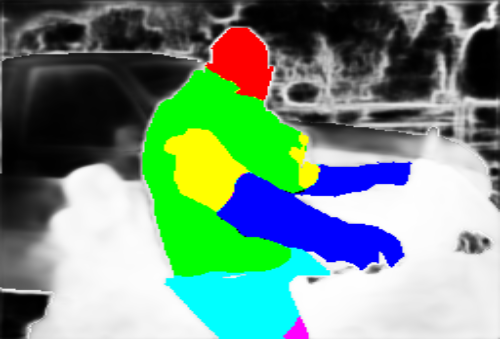} &
    \includegraphics[width=\sz\linewidth]{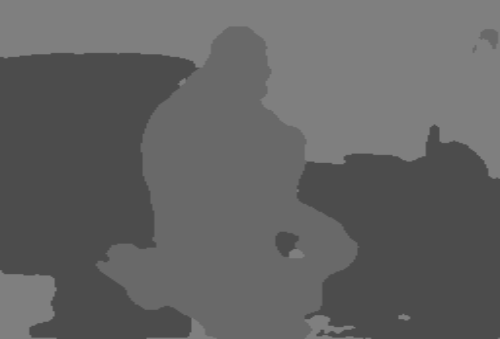} &
    \includegraphics[width=\sz\linewidth]{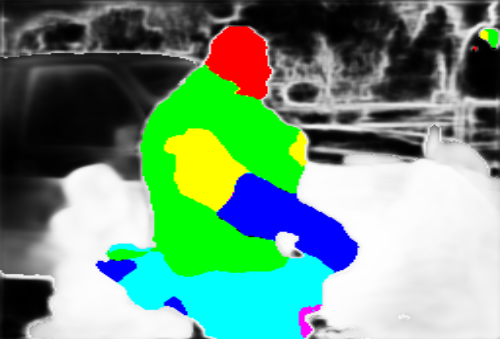}\\

    \includegraphics[width=\sz\linewidth]{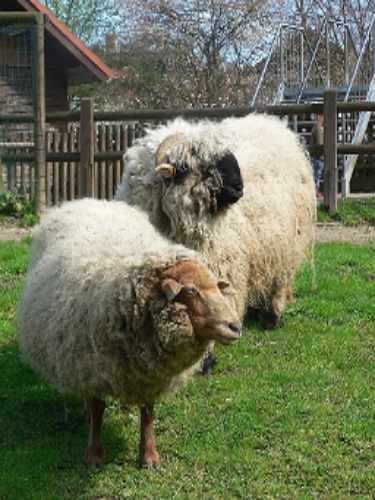} &
    \includegraphics[width=\sz\linewidth]{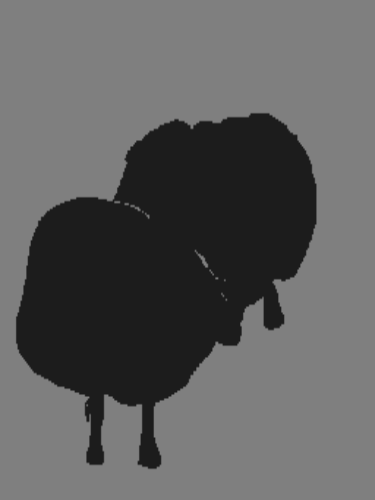} &
    \includegraphics[width=\sz\linewidth]{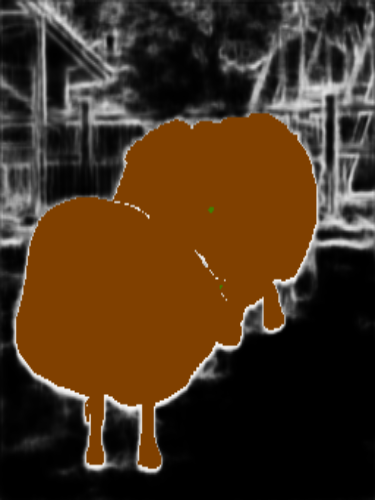} &
    \includegraphics[width=\sz\linewidth]{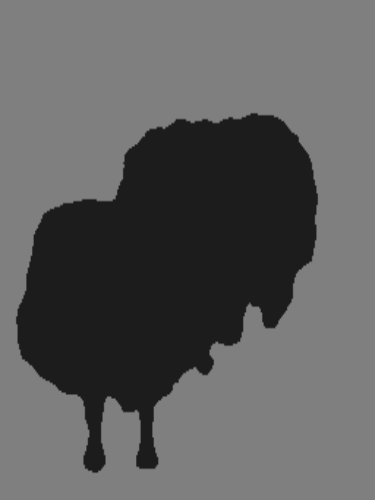} &
    \includegraphics[width=\sz\linewidth]{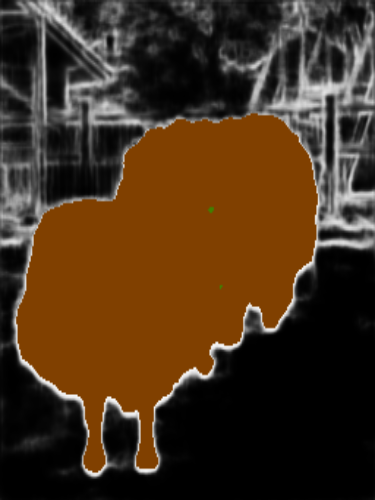}

  \end{tabular}
 \caption{\textbf{Learning what to do where.} CompositeTasking network predictions with the learned Task Palette. A separate network is learned to predict the Task Palette with $75.04 \%$ mIoU. }
    \label{fig:task_map_prediction_supp}
\end{figure*}

\noindent\textbf{Learning what to do where.} Additional visual examples from the experiments presented in Figure~\ref{fig:task_map_prediction} are presented in Figure \ref{fig:task_map_prediction_supp}. We can see the networks ability to predict the Task Palette. Using the predicted palette, the model successfully makes CompositeTasking predictions.

\begin{figure*}[t]
  \centering
  \captionsetup{font=small}
  \scriptsize
  \setlength{\tabcolsep}{1pt}
  \vspace*{-2pt}
  \newcommand{\sz}{0.189}           
  \newcommand{\cgs}{\hspace{3pt}}   
  \begin{tabular}{ccccc}
    \textbf{Image} & \textbf{Task Palette} & \textbf{Prediction} & \textbf{Edited Palette} & \textbf{Prediction} \\
    \includegraphics[width=\sz\linewidth]{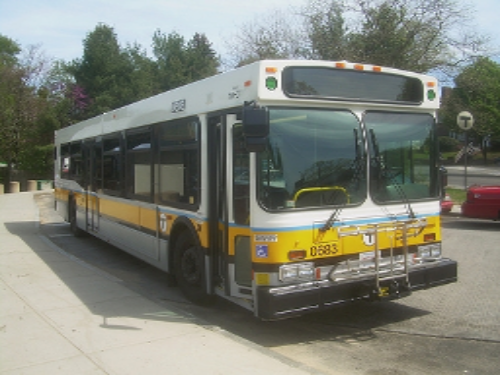} &
    \includegraphics[width=\sz\linewidth]{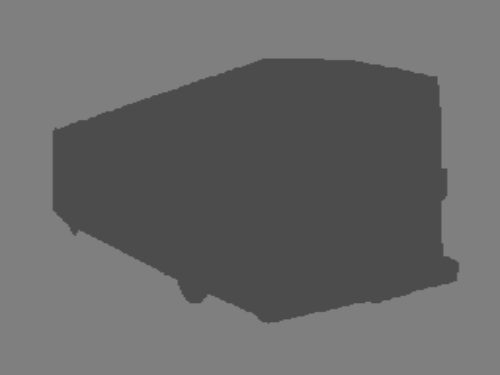} &
    \includegraphics[width=\sz\linewidth]{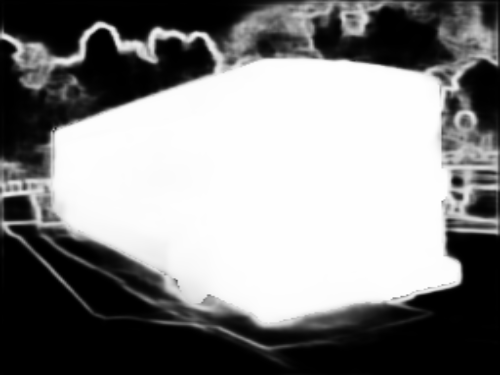} &
    \includegraphics[width=\sz\linewidth]{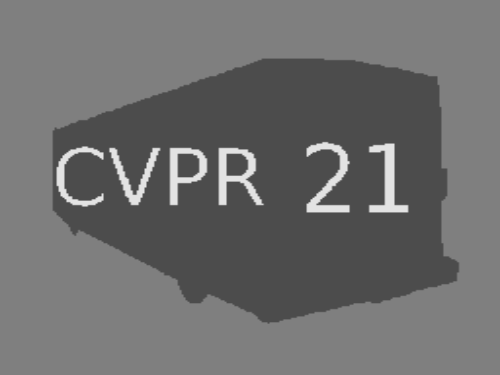} &
    \includegraphics[width=\sz\linewidth]{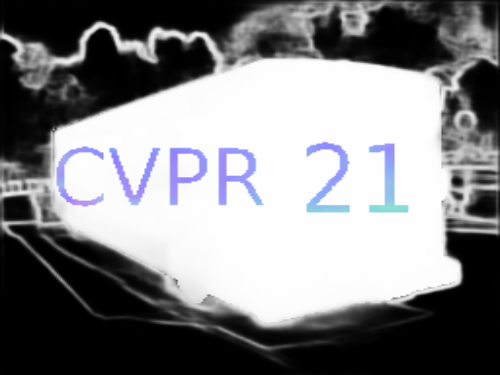} \\
        
    \includegraphics[width=\sz\linewidth]{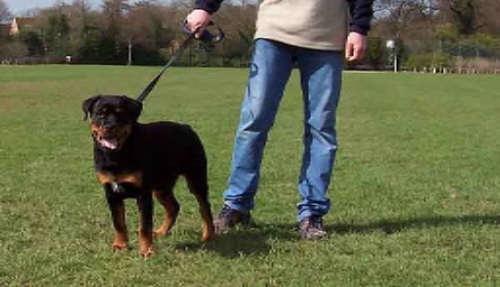} &
    \includegraphics[width=\sz\linewidth]{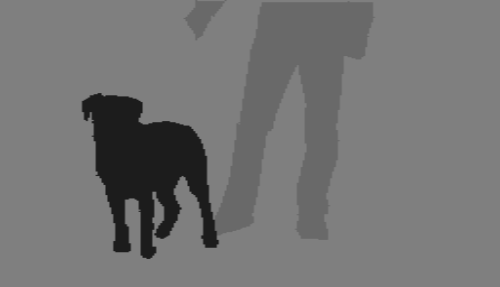} &
    \includegraphics[width=\sz\linewidth]{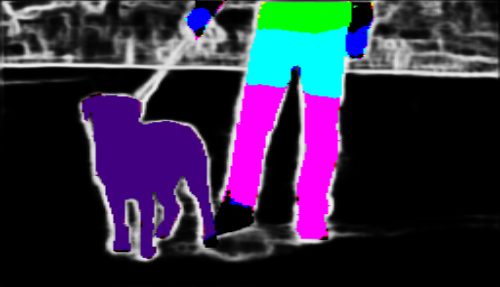} &
    \includegraphics[width=\sz\linewidth]{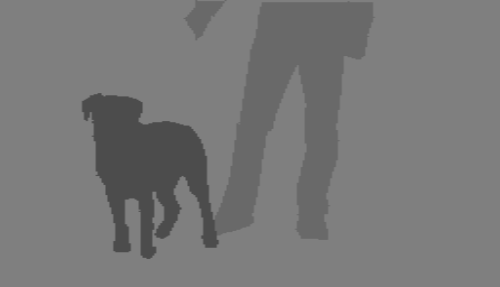} &
    \includegraphics[width=\sz\linewidth]{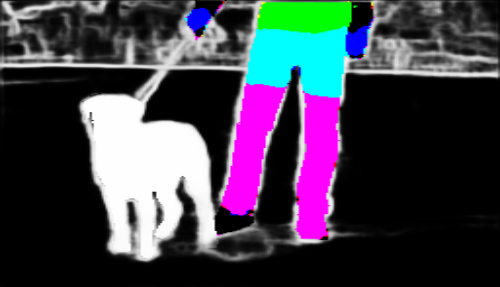} \\

    \includegraphics[width=\sz\linewidth]{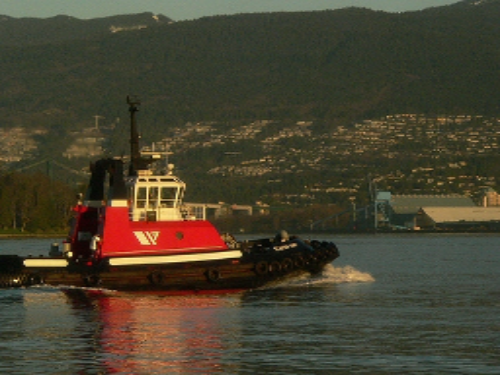} &
    \includegraphics[width=\sz\linewidth]{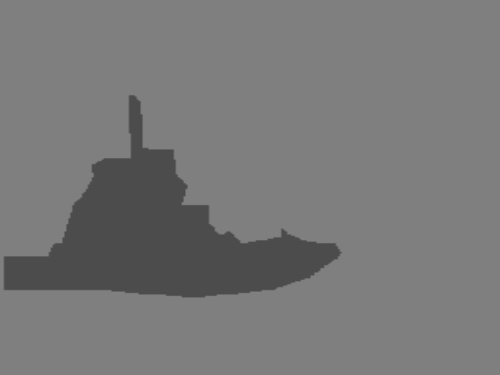} &
    \includegraphics[width=\sz\linewidth]{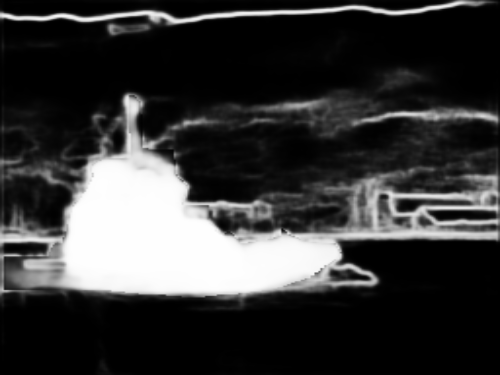} &
    \includegraphics[width=\sz\linewidth]{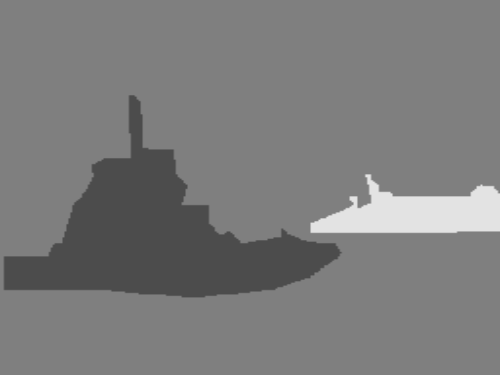} &
    \includegraphics[width=\sz\linewidth]{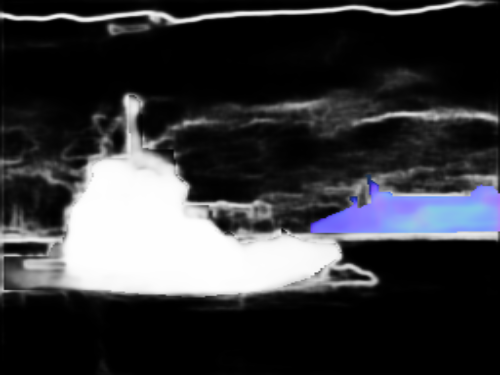} 

  \end{tabular}
 \caption{\textbf{Task Palette editing.} The Task Palette has been modified manually. A prediction of the CompositeTasking network is shown before and after modification. }
    \label{fig:rule_editing_supp}
\end{figure*}

\noindent\textbf{Task Palette editing.} Additional visual examples from the experiments presented in Figure~\ref{fig:rule_editing} are presented in Figure~\ref{fig:rule_editing_supp}. We can see that a user can make any modification to a given Task Palette, and obtain the respective CompositeTasking prediction.

\begin{figure*}[t]
  \centering
  \captionsetup{font=small}
  \scriptsize
  \setlength{\tabcolsep}{1pt}
  \vspace*{-2pt}
  \newcommand{\sz}{0.189}           
  \newcommand{\cgs}{\hspace{3pt}}   
  \begin{tabular}{ccccc}
    \textbf{Image} & \textbf{Pred $R_{2}$} & \textbf{Pred $R_{3}$} & \textbf{Pred $R_{2}$} & \textbf{Pred $R_{3}$} \\
     & \textbf{(Trained $R_{2}$)} & \textbf{(Trained $R_{2}$)} & \textbf{(Tuned $R_{2} \rightarrow R_{3}$)} & \textbf{(Tuned $R_{2} \rightarrow R_{3}$)} \\
    \includegraphics[width=\sz\linewidth]{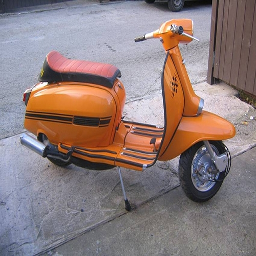} &
    \includegraphics[width=\sz\linewidth]{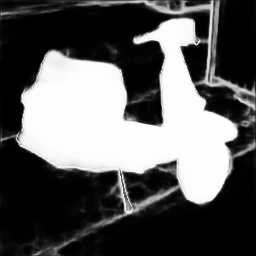} &
    \includegraphics[width=\sz\linewidth]{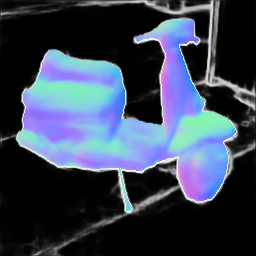} &
    \includegraphics[width=\sz\linewidth]{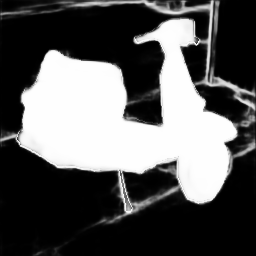} &
    \includegraphics[width=\sz\linewidth]{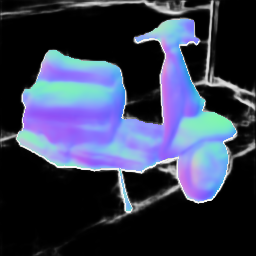}\\

    \includegraphics[width=\sz\linewidth]{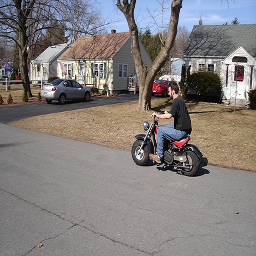} &
    \includegraphics[width=\sz\linewidth]{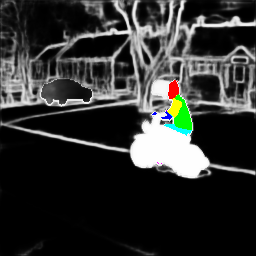} &
    \includegraphics[width=\sz\linewidth]{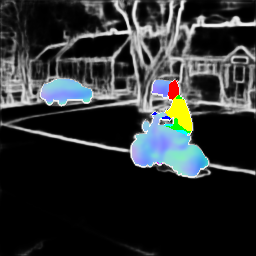} &
    \includegraphics[width=\sz\linewidth]{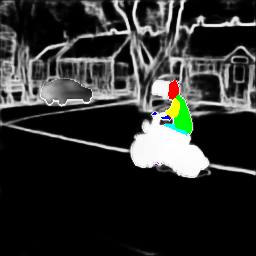} &
    \includegraphics[width=\sz\linewidth]{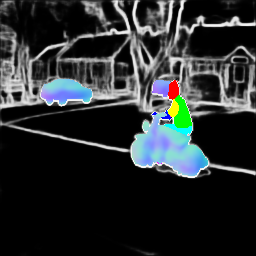}\\
        
    \includegraphics[width=\sz\linewidth]{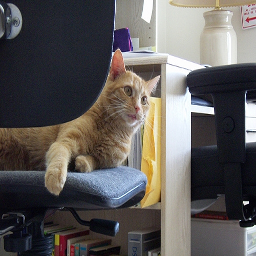} &
    \includegraphics[width=\sz\linewidth]{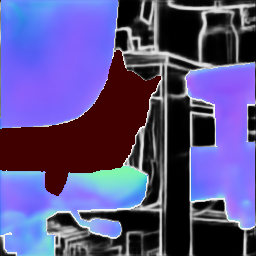} &
    \includegraphics[width=\sz\linewidth]{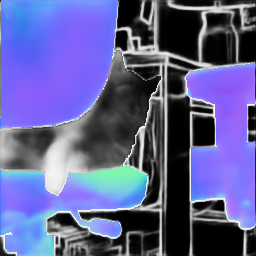} &
    \includegraphics[width=\sz\linewidth]{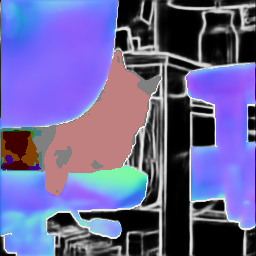} &
    \includegraphics[width=\sz\linewidth]{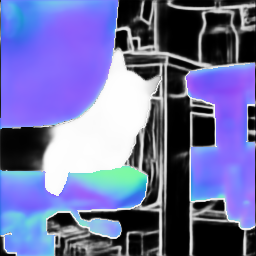}\\
    
    \includegraphics[width=\sz\linewidth]{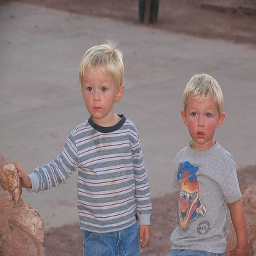} &
    \includegraphics[width=\sz\linewidth]{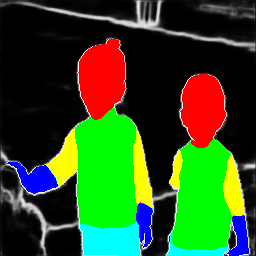} &
    \includegraphics[width=\sz\linewidth]{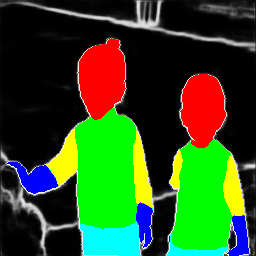} &
    \includegraphics[width=\sz\linewidth]{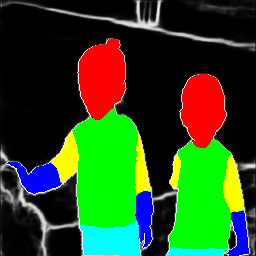} &
    \includegraphics[width=\sz\linewidth]{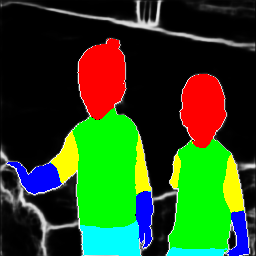}

 \end{tabular}
 \caption{\textbf{Breaking the rule.} Predictions of the CompositeTasking network are show before and after fine-tuning from the old to the new semantic rule. Because the rules are similar in a way, the model can already extrapolate even before fine-tuning. }
    \label{fig:breaking_the_rule_supp}
\end{figure*}

\noindent\textbf{Breaking the rule.} Additional visual examples from the experiments presented in Figure~\ref{fig:breaking_the_rule} are presented in Figure~\ref{fig:breaking_the_rule_supp}. We see the model's ability to transfer the knowledge of the rule it trained on ($\mathcal{R}_2$), to a new rule ($\mathcal{R}_3$). The performance gets even better when fine-tuning is done on the new rule. In the $1$st row we can see that the model was able to predict relatively good surface normals for the motorbike, even though it was never supervised to predict normals on motorbikes. After some fine-tuning on the new rule, it's predictions get even better. Notice how some almost flat surfaces on the vehicle (which have the same surface orientation) get more consistent normal predictions after fine-tuning. Since the new rule $\mathcal{R}_3$ does not contain the task of semantic segmentation anymore, that task is forgotten as can be seen in the $3$rd row. In the $4$th row, human body parts result, obtained using the rule $\mathcal{R}_3$, is shown.

\section{Implementation Details}
\label{sec:implementation_details_supp}

The networks are trained on the train partition of the data set and evaluated on the validation partition of the data set, in absence of a test partition. The images are resized to $256\times 256$ both for training and validation. When using the rule $\mathcal{R}_{1r}$ for training, a new center $\mathbf{c}$ and task region-task assignments $k_r$ are sampled for every batch. Training data was augmented using random horizontal flipping, shifting, scaling, cropping, Gaussian noise, adaptive histogram equalization, brightness change, sharpening, blurring, contrast change, hue change and saturation change. Flipping, scaling and shifting were not used when predicting surface normals. 

Each distinct task $k$ of the Task Palette $\mathcal{T}$  is embedded into a latent vector $\mathbf{w}$ using the embedding network depicted in figure~\ref{fig:full_composite_network}. The embedding network is a fully connected neural network with $6$ layers and leaky ReLU activations. To obtain an embedding for a specific task, its code $z_k$ is given to the embedding network. The codes are vectors of length $20K$, where $K$ is the number of tasks. The code of task $1$ is constructed by putting ones in the first $20$ elements of the vectors, and zeroes everywhere else. For task $2$, we have ones in the vectors positions $21...40$, etc. This way, the code vectors for different tasks will be orthogonal to each other.
The code vectors $z_k$ are also $L_2$ normalized to have unit length.
Since the embedding $\mathcal{E}$ from Figure \ref{fig:e_embed_net} is needed in the decoder in different spatial sizes, a pyramid of downsized versions is prepared, along with the original $\mathcal{E}$. We use downsampling with bilinear interpolation. 

The losses for each task $k$ are computed independently for every pixel belonging to that task in $\mathcal{T}$, and averaged to produce $\mathcal{L}_k (\mathcal{O}, \mathcal{Y}_k, \mathcal{T})$ from (\ref{eq:total_loss}). For semantic segmentation and human body parts we use the focal loss \cite{lin2017focalloss} with $\gamma=2$, which was shown to be more effective in our setup than standard cross-entropy loss. We use weighted cross-entropy loss for edges. The weight for the positive class (edges) is $0.95$ while the weight or the negative class is $0.05$, because the edges are usually present in a substanially smaller portion of the image pixels. We use weighted cross-entropy loss for saliency. The weights are calculated in each batch as the inverse frequency of elements that belong to the positive and negative class, normalized to sum up to $1.0$. For surface normals we use $1 - CS(o,y)$, where $CS$ is the cosine similarity. The chosen loss weights $\lambda_k$ are $3$ for semantic segmentation, $4$ for human parts, $50$ for edges, $8$ for saliency and $4$ for surface normals. This provided the best trade-off in experiments.

The exact $3$D class anchors $\mathbf{a}_i$ (see eq. \ref{eq:seg_loss}) used for the task of semantic segmentation and human body parts can be found in Tables~\ref{table:seg_class_anchors}~and~\ref{table:parts_class_anchors}, respectively.

\begin{table}[t]
	\centering
	\tiny
  \captionsetup{font=small}
	\caption{{\textbf{Class anchors $\mathbf{a}_i$ for the task of semantic segmentation}. These anchors define which part of the $3$D output space corresponds to which class, based on the nearest neighbor principle. They also define the RGB class colors used for all the presented results.}}
	\label{table:seg_class_anchors}
	\vspace*{-2pt}
	\resizebox{\columnwidth}{!}{
	\setlength\tabcolsep{6pt}
	\renewcommand\arraystretch{1}
	\begin{tabular}{|c|c|c|c|c|}
		\hline
\rowcolor{mygray}
Class id & Class name & $x_1$ & $x_2$ & $x_3$ \\  \hline \hline
0 & Background & $0$ & $0$ & $0$ \\  \hline
1 & Cat & $0$ & $0$ & $64$ \\  \hline
2 & Aeroplane & $0$ & $0$ & $128$ \\  \hline
3 & Chair & $0$ & $0$ & $192$ \\  \hline
4 & Potted plant & $0$ & $64$ & $0$ \\  \hline
5 & Sheep & $0$ & $64$ & $128$ \\  \hline
6 & Bicycle & $0$ & $128$ & $0$ \\  \hline
7 & Cow & $0$ & $128$ & $64$ \\  \hline
8 & Bird & $0$ & $128$ & $128$ \\  \hline
9 & Dining table & $0$ & $128$ & $192$ \\  \hline
10 & Sofa & $0$ & $192$ & $0$ \\  \hline
11 & Train & $0$ & $192$ & $128$ \\  \hline
12 & Boat & $128$ & $0$ & $0$ \\  \hline
13 & Dog & $128$ & $0$ & $64$ \\  \hline
14 & Bottle & $128$ & $0$ & $128$ \\  \hline
15 & Horse & $128$ & $0$ & $192$ \\  \hline
16 & TV monitor & $128$ & $64$ & $0$ \\  \hline
17 & Bus & $128$ & $128$ & $0$ \\  \hline
18 & Motorbike & $128$ & $128$ & $64$ \\  \hline
19 & Car & $128$ & $128$ & $128$ \\  \hline
20 & Person & $128$ & $128$ & $192$ \\  \hline
\end{tabular}
}
\end{table}

\begin{table}[t]
	\centering
	\tiny
    \captionsetup{font=small}
	\caption{{\textbf{Class anchors $\mathbf{a}_i$ for the task of human body parts}. These anchors define which part of the $3$D output space corresponds to which class, based on the nearest neighbor principle. They also define the RGB class colors used for all the presented results.}}
	\label{table:parts_class_anchors}
	\vspace*{-2pt}
	\resizebox{\columnwidth}{!}{
	\setlength\tabcolsep{6pt}
	\renewcommand\arraystretch{1}
	\begin{tabular}{|c|c|c|c|c|}
		\hline
\rowcolor{mygray}
Class id & Class name & $x_1$ & $x_2$ & $x_3$ \\  \hline \hline
0 & Background & $0$ & $0$ & $0$ \\  \hline
1 & Head & $0$ & $0$ & $255$ \\  \hline
2 & Neck \& torso & $0$ & $255$ & $0$ \\  \hline
3 & Upper arms & $255$ & $0$ & $0$ \\  \hline
4 & Lower arms & $0$ & $255$ & $255$ \\  \hline
5 & Upper legs & $255$ & $0$ & $255$ \\  \hline
6 & Lower legs & $255$ & $255$ & $0$ \\  \hline
\end{tabular}
}
\end{table}

The model was optimized using the Adam optimization algorithm, using weight decay of magnitude $0.00001$. The encoder is pre-trained on ImageNet , while the weights of the decoder are randomly initialized. Because of that the encoder is optimized with a smaller learning rate of $0.00001$, while the decoder is optimized with the learning rate of $0.001$. If the loss does not improve over the course of $12$ epochs, the learning rates are multiplied by a factor of $0.3$. We use leaky ReLU as the activation function. Since the PASCAL data set only provides the training and validation partitions, the networks are trained for $100$ epochs which was shown to be enough for them to converge. They were trained on Nvidia GPUs with either 12Gb or 16Gb of RAM memory. A batch size of $10$ was used.

\begin{figure*}[hbt!]
  \centering
  \scriptsize
  \setlength{\tabcolsep}{1pt}
  \vspace*{-2pt}
  \newcommand{\sz}{0.1575}           
  \newcommand{\cgs}{\hspace{3pt}}   
  \begin{tabular}{cccccc}
    \textbf{Image} & \textbf{Edges} & \textbf{Segmentation} & \textbf{Human parts} & \textbf{Surface normals} & \textbf{Saliency} \\
    \includegraphics[width=\sz\linewidth]{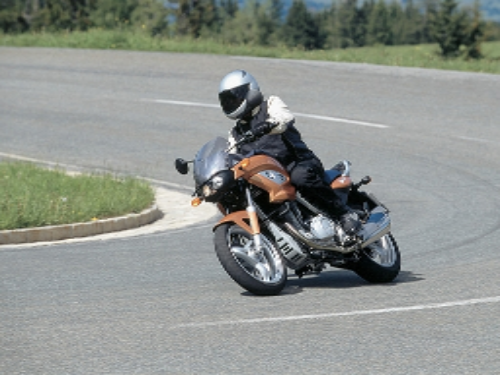} &
    \includegraphics[width=\sz\linewidth]{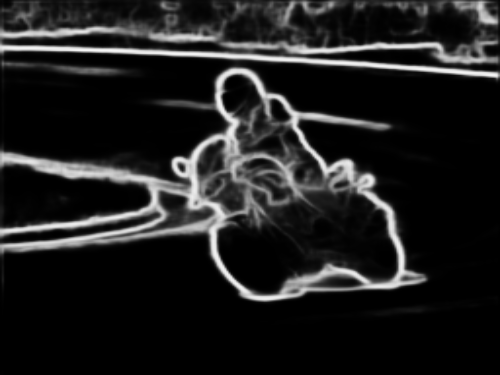} &
    \includegraphics[width=\sz\linewidth]{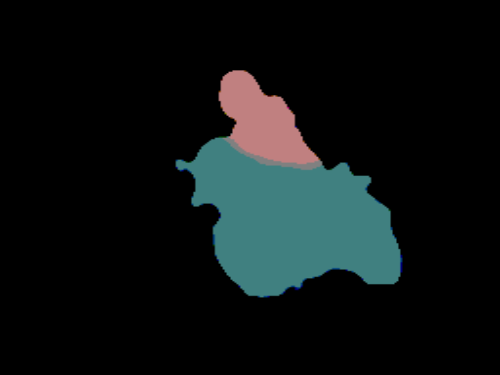} &
    \includegraphics[width=\sz\linewidth]{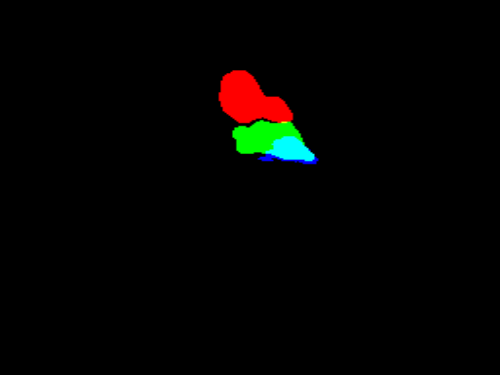} &
    \includegraphics[width=\sz\linewidth]{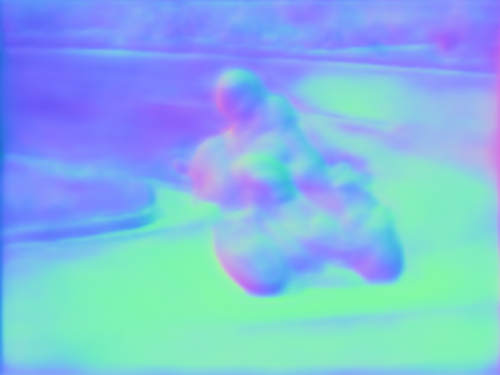} &
    \includegraphics[width=\sz\linewidth]{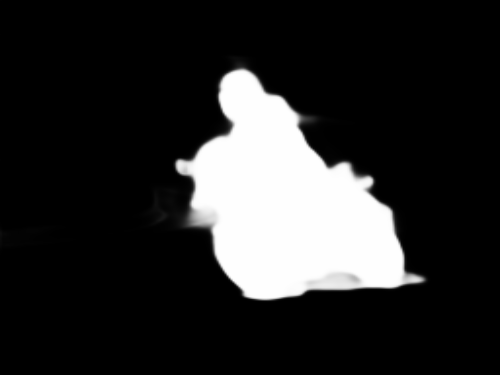} 

  \end{tabular}
 \caption{\textbf{Unsuccessful single task predictions.} }
    \label{fig:fail_st_predictions}
\end{figure*}

\begin{figure*}[hbt!]
  \centering
  \scriptsize
  \setlength{\tabcolsep}{1pt}
  \vspace*{-2pt}
  \newcommand{\sz}{0.189}           
  \newcommand{\cgs}{\hspace{3pt}}   
  \begin{tabular}{ccccc}
    \textbf{Image} & \textbf{Task palette} & \textbf{Prediction} & \textbf{Predicted Palette} & \textbf{Prediction} \\    
    \includegraphics[width=\sz\linewidth]{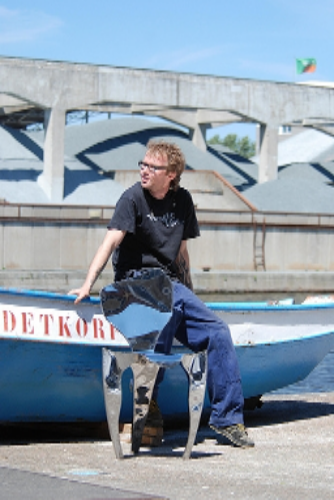} &
    \includegraphics[width=\sz\linewidth]{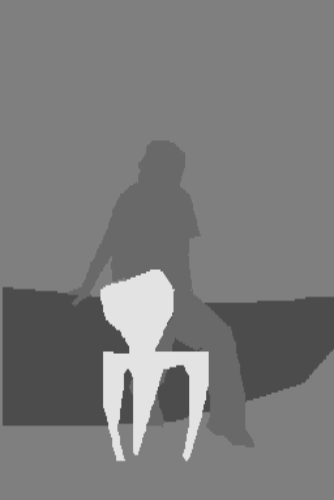} &
    \includegraphics[width=\sz\linewidth]{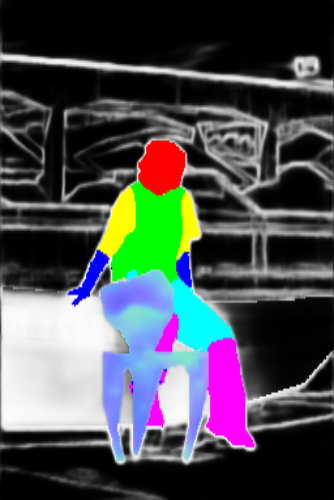} &
    \includegraphics[width=\sz\linewidth]{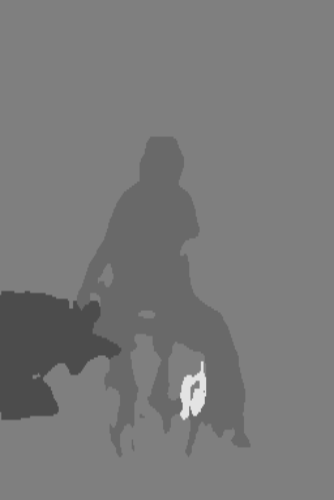} &
    \includegraphics[width=\sz\linewidth]{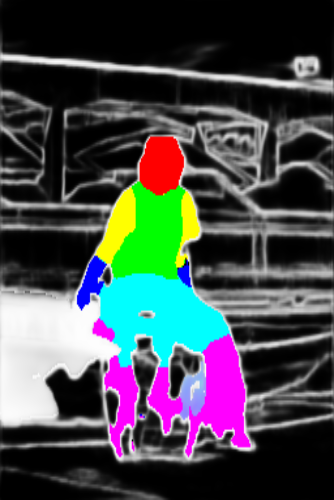}

  \end{tabular}
 \caption{\textbf{Unsuccessfully predicting what to do where.} }
    \label{fig:fail_task_map_prediction}
\end{figure*}

\section{Further Analysis}
\label{sec:more_exp}

In this section, we present analysis that shed more light on our choice of hyper-parameters. The average per-task drop metric from Section~\ref{sec:general_behaviour} is also used here. Instead of computing the metric with respect to a single task baseline, now it is computed with respect to the performance of the model with the chosen hyper-parameters. All the following analysis are conducted by evaluating on the single task rule $\mathcal{S}$.

In the task composition blocks we use one shared convolution that processes the Task Palette embedding $\mathcal{E}(\mathcal{T})$, before it is converted to the affine transformation parameters $\gamma$ and $\beta$. In Table \ref{table:n_conv_spade} we see that one shared convolution was the best choice. If we use more shared convolutions, we achieve a slightly worse performance with an increased computational cost.

\begin{table}[t]
	\centering\small
    \captionsetup{font=small}
	\caption{{\textbf{Number of shared convolutions that process the Task Palette embedding $\mathcal{E}(\mathcal{T})$ in the task composition block}. }}
	\label{table:n_conv_spade}
	\vspace*{-2pt}
	\resizebox{\columnwidth}{!}{
	\setlength\tabcolsep{6pt}
	\renewcommand\arraystretch{1}
	\begin{tabular}{|c|c|c|c|c|c||c|c|c}
		\hline
\rowcolor{mygray}
\# of convolutions  & Edge$\uparrow$& SemSeg$\uparrow$ & Parts$\uparrow$ & Normals$\downarrow$ & Sal$\uparrow$ & $\Delta_m\%$$\downarrow$ \\  \hline \hline
$1$ (\textbf{chosen})  & 68.60 & 62.45 & 52.59 & 16.93 & 67.81 & 0.00\% \\  \hline
$2$  &  68.60 & 61.93 & 51.49 & 17.15 & 67.56 & -0.92\%  \\  \hline 
$3$  &  68.30 & 62.01 & 52.28 & 17.16 & 67.60 & -0.68\%  \\  \hline 
\end{tabular}
}
\end{table}

We used a regular convolution with kernel size $3 \times 3$ in the task composition blocks of the decoder. In Table \ref{table:kernel_size_dec} we see that a kernel size of $1 \times 1$ achieves worse performance then our choice.

\begin{table}[t]
	\centering\small
    \captionsetup{font=small}
	\caption{{\textbf{Kernel size of regular convolutions inside the decoders task composition blocks}. }}
	\label{table:kernel_size_dec}
	\vspace*{-2pt}
	\resizebox{\columnwidth}{!}{
	\setlength\tabcolsep{6pt}
	\renewcommand\arraystretch{1}
	\begin{tabular}{|c|c|c|c|c|c||c|c|c}
		\hline
\rowcolor{mygray}
Kernel size  & Edge$\uparrow$& SemSeg$\uparrow$ & Parts$\uparrow$ & Normals$\downarrow$ & Sal$\uparrow$ & $\Delta_m\%$$\downarrow$ \\  \hline \hline
$3 \times 3$ (\textbf{chosen})  & 68.60 & 62.45 & 52.59 & 16.93 & 67.81 & 0.00\% \\  \hline
$1 \times 1$  & 66.20 & 61.50 & 51.05 & 17.87 & 67.16 & -2.89\%  \\  \hline 
\end{tabular}
}
\end{table}

We choose to do image-to-image translation by having $3$ output channels. In Table \ref{table:n_out_ch} we observe that if we use more output channels, there is no gain in predictive performance.

\begin{table}[t]
	\centering\small
    \captionsetup{font=small}
	\caption{{\textbf{Number of output channels of the model}. }}
	\label{table:n_out_ch}
	\vspace*{-2pt}
	\resizebox{\columnwidth}{!}{
	\setlength\tabcolsep{6pt}
	\renewcommand\arraystretch{1}
	\begin{tabular}{|c|c|c|c|c|c||c|c|c}
		\hline
\rowcolor{mygray}
\# of output channels  & Edge$\uparrow$& SemSeg$\uparrow$ & Parts$\uparrow$ & Normals$\downarrow$ & Sal$\uparrow$ & $\Delta_m\%$$\downarrow$ \\  \hline \hline
$3$ (\textbf{chosen})  & 68.60 & 62.45 & 52.59 & 16.93 & 67.81 & 0.00\% \\  \hline
$6$  & 68.50 & 62.35 & 52.99 & 17.08 & 67.65 & -0.13\%  \\  \hline 
$9$  & 68.50 & 62.42 & 52.50 & 17.39 & 67.89 & -0.59\%  \\  \hline 
\end{tabular}
}
\end{table}

When a Task Palette embedding $\mathcal{E}(\mathcal{T})$ is produced, it is down-scaled in all spatial sizes that the decoder needs. In Table \ref{table:interpolation_type} we compare our choice of bi-linear interpolation to nearest neighbours. We choose the bi-linear interpolation because it achieves slightly better results. This discrepancy in performance may be caused due to the negligence of the sampling theorem, i.e. ignoring that downsampled pixels on the task boundary represent more than just one distinct task~\cite{zhang2019making}. Also, the nearest neighbours interpolation can be used as a trade-off between slightly worse performance and slightly faster computations. When applying nearest neighbours interpolation there is no need to take into account all the values of $\mathcal{E}(\mathcal{T})$, but only the closest elements to the new pixel centers.

\begin{table}[t]
	\centering\small
    \captionsetup{font=small}
	\caption{{\textbf{Type of interpolation used during the Task Palette embedding $\mathcal{E}(\mathcal{T})$ downsampling}. }}
	\label{table:interpolation_type}
	\vspace*{-2pt}
	\resizebox{\columnwidth}{!}{
	\setlength\tabcolsep{6pt}
	\renewcommand\arraystretch{1}
	\begin{tabular}{|c|c|c|c|c|c||c|c|c}
		\hline
\rowcolor{mygray}
Interpolation  & Edge$\uparrow$& SemSeg$\uparrow$ & Parts$\uparrow$ & Normals$\downarrow$ & Sal$\uparrow$ & $\Delta_m\%$$\downarrow$ \\  \hline \hline
Bi-linear (\textbf{chosen})  & 68.60 & 62.45 & 52.59 & 16.93 & 67.81 & 0.00\% \\  \hline
Nearest neighbour  & 67.60 & 60.92 & 51.37 & 17.20 & 67.51 & -1.65\%  \\  \hline 
\end{tabular}
}
\end{table}

In Table \ref{table:n_fc_z_w} we analyze what is the good number of fully connected layers in the task representation block. We can see that if we pick the number to be too small or too big, there appears to be problems during learning. Especially for a very big number of layers like $12$, the network is having a hard time converging. Having $6$ and $9$ layers gave very similar results, and $6$ was chosen simply because it is computationally less demanding.

\begin{table}[t]
	\centering\small
    \captionsetup{font=small}
	\caption{{\textbf{Number of fully connected layers in the task representation block}. }}
	\label{table:n_fc_z_w}
	\vspace*{-2pt}
	\resizebox{\columnwidth}{!}{
	\setlength\tabcolsep{6pt}
	\renewcommand\arraystretch{1}
	\begin{tabular}{|c|c|c|c|c|c||c|c|c}
		\hline
\rowcolor{mygray}
\# of layers  & Edge$\uparrow$& SemSeg$\uparrow$ & Parts$\uparrow$ & Normals$\downarrow$ & Sal$\uparrow$ & $\Delta_m\%$$\downarrow$ \\  \hline \hline
$3$ & 68.00 & 7.61 & 42.62 & 16.37 & 71.53 &  -19.77\%  \\  \hline 
$6$ (\textbf{chosen}) & 68.60 & 62.45 & 52.59 & 16.93 & 67.81 & 0.00\% \\  \hline
$9$ & 68.70 & 62.29 & 52.18 & 16.95 & 67.43 &  -0.31\%  \\  \hline 
$12$ & 52.80 & 36.46 & 13.69 & 67.60 & 44.93 &  -94.33\%  \\  \hline 
\end{tabular}
}
\end{table}

From Table \ref{table:dim_w} we can see that the model is pretty robust to choosing the number of channels of the Task Palette embedding $\mathcal{E}(\mathcal{T})$. We chosen the number of channels which gave the best performance.

\begin{table}[t]
	\centering\small
    \captionsetup{font=small}
	\caption{{\textbf{Number of channels of the Task Palette embedding $\mathcal{E}(\mathcal{T})$}. }}
	\label{table:dim_w}
	\vspace*{-2pt}
	\resizebox{\columnwidth}{!}{
	\setlength\tabcolsep{6pt}
	\renewcommand\arraystretch{1}
	\begin{tabular}{|c|c|c|c|c|c||c|c|c}
		\hline
\rowcolor{mygray}
\# of channels  & Edge$\uparrow$& SemSeg$\uparrow$ & Parts$\uparrow$ & Normals$\downarrow$ & Sal$\uparrow$ & $\Delta_m\%$$\downarrow$ \\  \hline \hline
$32$  & 68.70 & 62.14 & 51.86 & 16.95 & 67.46 & -0.47\%  \\  \hline 
$64$  & 68.60 & 62.06 & 52.67 & 17.06 & 67.86 & -0.23\%  \\  \hline 
$128$ (\textbf{chosen})  & 68.60 & 62.45 & 52.59 & 16.93 & 67.81 & 0.00\% \\  \hline
$256$  & 68.40 & 60.36 & 51.53 & 17.11 & 67.35 & -1.48\%  \\  \hline 
$512$  & 1.40 & 11.22 & 4.53 & 30.87 & 18.64 & -85.26\%  \\  \hline 
\end{tabular}
}
\end{table}

In Table \ref{table:enc_backbone} we analyze what happens if we use encoder backbones of different capacity. We can see a clear trend from the table --- as the backbone has more capacity (more learnable parameters and computations) the predictive performance of the model gets better. This was expected, and it shows one possible way to improve the performance. The choice for ResNet$34$ trades-off between performance and memory/computational demands.

\begin{table}[t]
	\centering\small
    \captionsetup{font=small}
	\caption{{\textbf{Encoder backbone}. }}
	\label{table:enc_backbone}
	\vspace*{-2pt}
	\resizebox{\columnwidth}{!}{
	\setlength\tabcolsep{6pt}
	\renewcommand\arraystretch{1}
	\begin{tabular}{|c|c|c|c|c|c||c|c|c}
		\hline
\rowcolor{mygray}
Backbone  & Edge$\uparrow$& SemSeg$\uparrow$ & Parts$\uparrow$ & Normals$\downarrow$ & Sal$\uparrow$ & $\Delta_m\%$$\downarrow$ \\  \hline \hline
ResNet$18$ & 67.40 & 60.40 & 49.25 & 17.25 & 66.87 &  -2.93\%  \\  \hline 
ResNet$34$ (\textbf{chosen}) & 68.60 & 62.45 & 52.59 & 16.93 & 67.81 & 0.00\% \\  \hline
ResNet$50$  & 69.40 & 62.34 & 53.81 & 16.64 & 68.19 & +1.12\%  \\  \hline 
ResNet$101$  & 69.60 & 63.75 & 55.78 & 16.71 & 69.55 & +2.69\%  \\  \hline 
\end{tabular}
}
\end{table}

In Table \ref{table:normlas_loss} we analyze the choice of the loss function for surface normals. As we can see, using cosine similarity achieves much better results that the popular L$1$ loss. Also, when using L$1$ loss, the values of the loss function over epochs looked much more unstable than compared to using cosine similarity.

\begin{table}[t]
	\centering\small
    \captionsetup{font=small}
	\caption{{\textbf{Loss choice for the task of estimating surface normals}. }}
	\label{table:normlas_loss}
	\vspace*{-2pt}
	\resizebox{\columnwidth}{!}{
	\setlength\tabcolsep{6pt}
	\renewcommand\arraystretch{1}
	\begin{tabular}{|c|c|c|c|c|c||c|c|c}
		\hline
\rowcolor{mygray}
Loss function  & Edge$\uparrow$& SemSeg$\uparrow$ & Parts$\uparrow$ & Normals$\downarrow$ & Sal$\uparrow$ & $\Delta_m\%$$\downarrow$ \\  \hline \hline
cosine similarity (\textbf{chosen})  & 68.60 & 62.45 & 52.59 & 16.93 & 67.81 & 0.00\% \\  \hline
$l_1$ distance  & 67.60 & 58.91 & 49.20 & 19.77 & 67.53 & -6.15\%  \\  \hline 
\end{tabular}
}
\end{table}

In Table \ref{table:seg_loss_gamma} we analyze the choice of $\gamma$ in the focal loss used in segmentation and human parts. We choose $\gamma=2$ because that it achieved the best performance. Using $\gamma=0$ is the same as using regular cross-entropy. Even though regular cross-entropy was close in performance to using the focal loss, when supervising with it the models often failed to converge.

\begin{table}[t]
	\centering\small
    \captionsetup{font=small}
	\caption{{\textbf{Choice of parameter $\gamma$ from focal loss used in segmentation and human parts}. }}
	\label{table:seg_loss_gamma}
	\vspace*{-2pt}
	\resizebox{\columnwidth}{!}{
	\setlength\tabcolsep{6pt}
	\renewcommand\arraystretch{1}
	\begin{tabular}{|c|c|c|c|c|c||c|c|c}
		\hline
\rowcolor{mygray}
$\gamma$  & Edge$\uparrow$& SemSeg$\uparrow$ & Parts$\uparrow$ & Normals$\downarrow$ & Sal$\uparrow$ & $\Delta_m\%$$\downarrow$ \\  \hline \hline
$0$ & 68.40 & 61.39 & 51.93 & 17.29 & 67.40 &  -1.19\%  \\  \hline 
$1$ & 68.10 & 61.99 & 51.61 & 17.46 & 67.05 &  -1.52\%  \\  \hline 
$2$ (\textbf{chosen}) & 68.60 & 62.45 & 52.59 & 16.93 & 67.81 & 0.00\% \\  \hline
$3$ & 68.70 & 61.60 & 51.39 & 17.06 & 67.88 &  -0.83\%  \\  \hline 
\end{tabular}
}
\end{table}
\section{Limitations}
\label{sec:limitations}

\begin{figure}[t]
  \centering
  \captionsetup{font=small}
  \scriptsize
  \setlength{\tabcolsep}{1pt}
  \vspace*{-2pt}
  \newcommand{\sz}{0.48}           
  \newcommand{\cgs}{\hspace{3pt}}   
  \begin{tabular}{cc}
    \textbf{Image} & \textbf{Pred} \\
    \includegraphics[width=\sz\columnwidth]{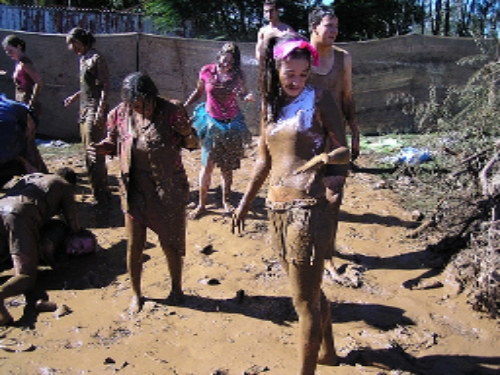} &
    \includegraphics[width=\sz\columnwidth]{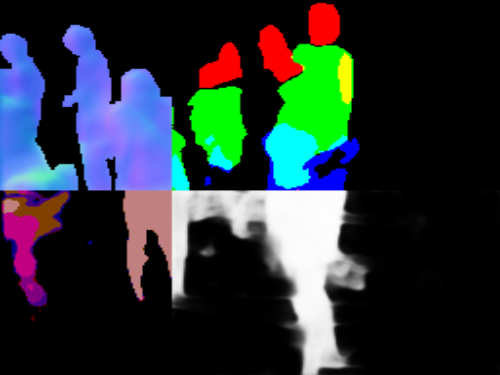} \\
    \includegraphics[width=\sz\columnwidth]{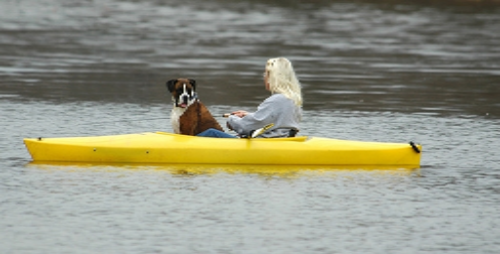} &
    \includegraphics[width=\sz\columnwidth]{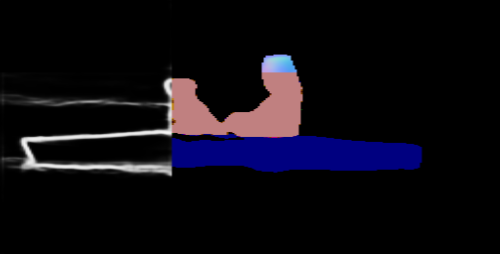} 

  \end{tabular}
 \caption{\textbf{Unsuccessful random mosaic composition predictions.}}
    \label{fig:fail_mosaic_prediction}
\end{figure}

The proposed network has the ability to only predict one task for each spatial location, during one forward pass. In some cases, this could be a limitation. When making predictions on high-level tasks, modern approaches predict many different low-level tasks and fuse them in order to make the final prediction. Although it is unnecessary to predict every task everywhere -- as laid out in our introduction --, sometimes it could certainly be necessary to perform more than one task per each spatial location. As one example, lets think about predicting where to turn the steering wheel in a self-driving application. When a person is in the frame we might want to detect them as a person (semantic segmentation), but we would also like to know the body part locations and the depth at its location so we can reason about the person's pose, intended motion, and how close the person is to the car. 

Even in the case of our one-task-per-location model however, predictions are not always perfect. Now we will look at some examples, where the model does not make good predictions. In the $1$st row of Figure \ref{fig:fail_mosaic_prediction}, we can see that the model fails to predict the body parts, and the semantic segmentation correctly. This happens often when there are many people in the scene, especially in strange circumstances like in this example. In the $2$nd row of Figure \ref{fig:fail_mosaic_prediction}, we can see that the model segmented the dog as a human, probably because it was unlikely that a dog is riding on a boat in the dataset. In Figure~\ref{fig:fail_st_predictions}, we can see that the model is not able to predict the human parts successfully, because the model recognises most of the human body as the motorcycle as can be notices in the semantic segmentation prediction. Finally, on Figure~\ref{fig:fail_task_map_prediction} we can see what happens if the Task Palette is not predicted correctly. Here the model is able to make a very good prediction given the appropriate palette. But when the palette is predicted wrongly, probably because of a lot of objects overlapping, the model makes a bad prediction.

{\small
\bibliographystyle{ieee_fullname}
\bibliography{main}

\begin{thebibliography}{10}\itemsep=-1pt

\bibitem{achille2019task2vec}
Alessandro Achille, Michael Lam, Rahul Tewari, Avinash Ravichandran, Subhransu
  Maji, Charless~C Fowlkes, Stefano Soatto, and Pietro Perona.
\newblock Task2vec: Task embedding for meta-learning.
\newblock In {\em Proceedings of the IEEE International Conference on Computer
  Vision}, pages 6430--6439, 2019.

\bibitem{aditya2019integrating}
Somak Aditya, Yezhou Yang, and Chitta Baral.
\newblock Integrating knowledge and reasoning in image understanding.
\newblock {\em arXiv preprint arXiv:1906.09954}, 2019.

\bibitem{Alonso2017CoralSegmentationTD}
I{\~n}igo Alonso, Ana~B. Cambra, A. Mu{\~n}oz, T. Treibitz, and A.~C. Murillo.
\newblock Coral-segmentation: Training dense labeling models with sparse ground
  truth.
\newblock {\em 2017 IEEE International Conference on Computer Vision Workshops
  (ICCVW)}, pages 2874--2882, 2017.

\bibitem{Ba2016LayerN}
Jimmy Ba, J. Kiros, and Geoffrey~E. Hinton.
\newblock Layer normalization.
\newblock {\em ArXiv}, abs/1607.06450, 2016.

\bibitem{Bilen2017UniversalRT}
Hakan Bilen and A. Vedaldi.
\newblock Universal representations: The missing link between faces, text,
  planktons, and cat breeds.
\newblock {\em ArXiv}, abs/1701.07275, 2017.

\bibitem{Bogo2014FAUST}
Federica Bogo, Javier Romero, Matthew Loper, and Michael~J. Black.
\newblock {FAUST}: Dataset and evaluation for {3D} mesh registration.
\newblock In {\em Proceedings IEEE Conf. on Computer Vision and Pattern
  Recognition (CVPR)}, Piscataway, NJ, USA, June 2014. IEEE.

\bibitem{DBLP:journals/corr/abs-1708-02551}
Bert~De Brabandere, Davy Neven, and Luc~Van Gool.
\newblock Semantic instance segmentation with a discriminative loss function.
\newblock {\em CoRR}, abs/1708.02551, 2017.

\bibitem{Bragman2019StochasticFG}
Felix J.~S. Bragman, Ryutaro Tanno, S{\'e}bastien Ourselin, D. Alexander, and
  M. Cardoso.
\newblock Stochastic filter groups for multi-task cnns: Learning specialist and
  generalist convolution kernels.
\newblock {\em 2019 IEEE/CVF International Conference on Computer Vision
  (ICCV)}, pages 1385--1394, 2019.

\bibitem{Brock2019LargeSG}
A. Brock, J. Donahue, and K. Simonyan.
\newblock Large scale gan training for high fidelity natural image synthesis.
\newblock {\em ArXiv}, abs/1809.11096, 2019.

\bibitem{caruana1997multitask}
Rich Caruana.
\newblock Multitask learning.
\newblock {\em Machine learning}, 28(1):41--75, 1997.

\bibitem{chen2019touchdown}
Howard Chen, Alane Suhr, Dipendra Misra, Noah Snavely, and Yoav Artzi.
\newblock Touchdown: Natural language navigation and spatial reasoning in
  visual street environments.
\newblock In {\em Proceedings of the IEEE Conference on Computer Vision and
  Pattern Recognition}, pages 12538--12547, 2019.

\bibitem{Crawshaw2020MultiTaskLW}
M. Crawshaw.
\newblock Multi-task learning with deep neural networks: A survey.
\newblock {\em ArXiv}, abs/2009.09796, 2020.

\bibitem{8014800}
B. {De Brabandere}, D. {Neven}, and L. {Van Gool}.
\newblock Semantic instance segmentation for autonomous driving.
\newblock In {\em 2017 IEEE Conference on Computer Vision and Pattern
  Recognition Workshops (CVPRW)}, pages 478--480, 2017.

\bibitem{Doersch2017MultitaskSV}
C. Doersch and Andrew Zisserman.
\newblock Multi-task self-supervised visual learning.
\newblock {\em 2017 IEEE International Conference on Computer Vision (ICCV)},
  pages 2070--2079, 2017.

\bibitem{Durand2019LearningAD}
T. Durand, Nazanin Mehrasa, and G. Mori.
\newblock Learning a deep convnet for multi-label classification with partial
  labels.
\newblock {\em 2019 IEEE/CVF Conference on Computer Vision and Pattern
  Recognition (CVPR)}, pages 647--657, 2019.

\bibitem{PASCAL}
Mark Everingham, Luc Gool, Christopher~K. Williams, John Winn, and Andrew
  Zisserman.
\newblock The pascal visual object classes (voc) challenge.
\newblock {\em Int. J. Comput. Vision}, 88(2):303–338, June 2010.

\bibitem{DBLP:journals/corr/FathiWRWSGM17}
Alireza Fathi, Zbigniew Wojna, Vivek Rathod, Peng Wang, Hyun~Oh Song, Sergio
  Guadarrama, and Kevin~P. Murphy.
\newblock Semantic instance segmentation via deep metric learning.
\newblock {\em CoRR}, abs/1703.10277, 2017.

\bibitem{he2016resnet}
K. {He}, X. {Zhang}, S. {Ren}, and J. {Sun}.
\newblock Deep residual learning for image recognition.
\newblock In {\em 2016 IEEE Conference on Computer Vision and Pattern
  Recognition (CVPR)}, pages 770--778, 2016.

\bibitem{Huang2017ArbitraryST}
X. Huang and Serge~J. Belongie.
\newblock Arbitrary style transfer in real-time with adaptive instance
  normalization.
\newblock {\em 2017 IEEE International Conference on Computer Vision (ICCV)},
  pages 1510--1519, 2017.

\bibitem{Huynh2020InteractiveMC}
D. Huynh and E. Elhamifar.
\newblock Interactive multi-label cnn learning with partial labels.
\newblock {\em 2020 IEEE/CVF Conference on Computer Vision and Pattern
  Recognition (CVPR)}, pages 9420--9429, 2020.

\bibitem{ioffe2015BN}
Sergey Ioffe and Christian Szegedy.
\newblock Batch normalization: Accelerating deep network training by reducing
  internal covariate shift.
\newblock pages 448--456, 2015.

\bibitem{Kanakis2020ReparameterizingCF}
M. Kanakis, David Bruggemann, Suman Saha, S. Georgoulis, Anton Obukhov, and L.
  Gool.
\newblock Reparameterizing convolutions for incremental multi-task learning
  without task interference.
\newblock {\em ArXiv}, abs/2007.12540, 2020.

\bibitem{Karras2019ASG}
Tero Karras, S. Laine, and Timo Aila.
\newblock A style-based generator architecture for generative adversarial
  networks.
\newblock {\em 2019 IEEE/CVF Conference on Computer Vision and Pattern
  Recognition (CVPR)}, pages 4396--4405, 2019.

\bibitem{kim2018visual}
Seung~Wook Kim, Makarand Tapaswi, and Sanja Fidler.
\newblock Visual reasoning by progressive module networks.
\newblock In {\em International Conference on Learning Representations}, 2018.

\bibitem{kingma2014adam}
Diederik Kingma and Jimmy Ba.
\newblock Adam: A method for stochastic optimization, 2014.
\newblock cite arxiv:1412.6980Comment: Published as a conference paper at the
  3rd International Conference for Learning Representations, San Diego, 2015.

\bibitem{Krizhevsky2017ImageNetCW}
A. Krizhevsky, Ilya Sutskever, and Geoffrey~E. Hinton.
\newblock Imagenet classification with deep convolutional neural networks.
\newblock {\em Commun. ACM}, 60:84--90, 2017.

\bibitem{DBLP:journals/pami/LiangLWSYY18}
Xiaodan Liang, Liang Lin, Yunchao Wei, Xiaohui Shen, Jianchao Yang, and
  Shuicheng Yan.
\newblock Proposal-free network for instance-level object segmentation.
\newblock {\em {IEEE} Trans. Pattern Anal. Mach. Intell.}, 40(12):2978--2991,
  2018.

\bibitem{lpt2013ikea}
Joseph~J. Lim, Hamed Pirsiavash, and Antonio Torralba.
\newblock {Parsing IKEA Objects: Fine Pose Estimation}.
\newblock {\em ICCV}, 2013.

\bibitem{lin2017focalloss}
T. {Lin}, P. {Goyal}, R. {Girshick}, K. {He}, and P. {Dollár}.
\newblock Focal loss for dense object detection.
\newblock In {\em 2017 IEEE International Conference on Computer Vision
  (ICCV)}, pages 2999--3007, 2017.

\bibitem{Lin2014MicrosoftCC}
Tsung-Yi Lin, M. Maire, Serge~J. Belongie, James Hays, P. Perona, D. Ramanan,
  Piotr Doll{\'a}r, and C.~L. Zitnick.
\newblock Microsoft coco: Common objects in context.
\newblock {\em ArXiv}, abs/1405.0312, 2014.

\bibitem{Liu2019EndToEndML}
Shikun Liu, Edward Johns, and A. Davison.
\newblock End-to-end multi-task learning with attention.
\newblock {\em 2019 IEEE/CVF Conference on Computer Vision and Pattern
  Recognition (CVPR)}, pages 1871--1880, 2019.

\bibitem{Lu2017FullyAdaptiveFS}
Y. Lu, Abhishek Kumar, Shuangfei Zhai, Yu Cheng, T. Javidi, and R. Feris.
\newblock Fully-adaptive feature sharing in multi-task networks with
  applications in person attribute classification.
\newblock {\em 2017 IEEE Conference on Computer Vision and Pattern Recognition
  (CVPR)}, pages 1131--1140, 2017.

\bibitem{Manen2017PathTrackFT}
S. Manen, Michael Gygli, Dengxin Dai, and L. Gool.
\newblock Pathtrack: Fast trajectory annotation with path supervision.
\newblock {\em 2017 IEEE International Conference on Computer Vision (ICCV)},
  pages 290--299, 2017.

\bibitem{kevis2019attentive}
Kevis-Kokitsi Maninis, Ilija Radosavovic, and Iasonas Kokkinos.
\newblock Attentive single-tasking of multiple tasks.
\newblock In {\em CVPR}, 2019.

\bibitem{odsF}
D.~R. {Martin}, C.~C. {Fowlkes}, and J. {Malik}.
\newblock Learning to detect natural image boundaries using local brightness,
  color, and texture cues.
\newblock {\em IEEE Transactions on Pattern Analysis and Machine Intelligence},
  26(5):530--549, 2004.

\bibitem{Misra2016CrossStitchNF}
I. Misra, Abhinav Shrivastava, A. Gupta, and M. Hebert.
\newblock Cross-stitch networks for multi-task learning.
\newblock {\em 2016 IEEE Conference on Computer Vision and Pattern Recognition
  (CVPR)}, pages 3994--4003, 2016.

\bibitem{mittal2020emoticon}
Trisha Mittal, Pooja Guhan, Uttaran Bhattacharya, Rohan Chandra, Aniket Bera,
  and Dinesh Manocha.
\newblock Emoticon: Context-aware multimodal emotion recognition using frege's
  principle, 2020.

\bibitem{pascal-context}
Roozbeh Mottaghi, Xianjie Chen, Xiaobai Liu, Nam-Gyu Cho, Seong-Whan Lee, Sanja
  Fidler, Raquel Urtasun, and Alan Yuille.
\newblock The role of context for object detection and semantic segmentation in
  the wild.
\newblock In {\em CVPR}, 2014.

\bibitem{DBLP:journals/corr/abs-1708-02550}
Davy Neven, Bert~De Brabandere, Stamatios Georgoulis, Marc Proesmans, and
  Luc~Van Gool.
\newblock Fast scene understanding for autonomous driving.
\newblock {\em CoRR}, abs/1708.02550, 2017.

\bibitem{DBLP:conf/cvpr/NevenBPG19}
Davy Neven, Bert~De Brabandere, Marc Proesmans, and Luc~Van Gool.
\newblock Instance segmentation by jointly optimizing spatial embeddings and
  clustering bandwidth.
\newblock In {\em {IEEE} Conference on Computer Vision and Pattern Recognition,
  {CVPR} 2019, Long Beach, CA, USA, June 16-20, 2019}, pages 8837--8845.
  Computer Vision Foundation / {IEEE}, 2019.

\bibitem{DBLP:conf/nips/NewellHD17}
Alejandro Newell, Zhiao Huang, and Jia Deng.
\newblock Associative embedding: End-to-end learning for joint detection and
  grouping.
\newblock In Isabelle Guyon, Ulrike von Luxburg, Samy Bengio, Hanna~M. Wallach,
  Rob Fergus, S.~V.~N. Vishwanathan, and Roman Garnett, editors, {\em Advances
  in Neural Information Processing Systems 30: Annual Conference on Neural
  Information Processing Systems 2017, 4-9 December 2017, Long Beach, CA,
  {USA}}, pages 2277--2287, 2017.

\bibitem{ntavelis2020sesame}
Evangelos Ntavelis, Andrés Romero, Iason Kastanis, Luc~Van Gool, and Radu
  Timofte.
\newblock Sesame: Semantic editing of scenes by adding, manipulating or erasing
  objects, 2020.

\bibitem{park2019SPADE}
Taesung Park, Ming-Yu Liu, Ting-Chun Wang, and Jun-Yan Zhu.
\newblock Semantic image synthesis with spatially-adaptive normalization.
\newblock In {\em Proceedings of the IEEE Conference on Computer Vision and
  Pattern Recognition}, 2019.

\bibitem{seism}
Jordi Pont-Tuset and Ferran Marques.
\newblock Measures and meta-measures for the supervised evaluation of image
  segmentation.
\newblock In {\em Computer Vision and Pattern Recognition}, 2013.

\bibitem{Qiu2019DeepLiDARDS}
Jiaxiong Qiu, Zhaopeng Cui, Y. Zhang, Xingdi Zhang, Shuaicheng Liu, B. Zeng,
  and M. Pollefeys.
\newblock Deeplidar: Deep surface normal guided depth prediction for outdoor
  scene from sparse lidar data and single color image.
\newblock {\em 2019 IEEE/CVF Conference on Computer Vision and Pattern
  Recognition (CVPR)}, pages 3308--3317, 2019.

\bibitem{RA_series}
Sylvestre-Alvise Rebuffi, Hakan Bilen, and Andrea Vedaldi.
\newblock Learning multiple visual domains with residual adapters.
\newblock In {\em Advances in Neural Information Processing Systems}, pages
  506--516, 2017.

\bibitem{Rebuffi2018EfficientPO}
Sylvestre-Alvise Rebuffi, Hakan Bilen, and A. Vedaldi.
\newblock Efficient parametrization of multi-domain deep neural networks.
\newblock {\em 2018 IEEE/CVF Conference on Computer Vision and Pattern
  Recognition}, pages 8119--8127, 2018.

\bibitem{ronneberger2015u}
Olaf Ronneberger, Philipp Fischer, and Thomas Brox.
\newblock U-net: Convolutional networks for biomedical image segmentation.
\newblock In {\em MICAI}, 2015.

\bibitem{Russakovsky2015ImageNetLS}
Olga Russakovsky, J. Deng, H. Su, J. Krause, S. Satheesh, S. Ma, Zhiheng Huang,
  A. Karpathy, A. Khosla, Michael~S. Bernstein, A. Berg, and Li Fei-Fei.
\newblock Imagenet large scale visual recognition challenge.
\newblock {\em International Journal of Computer Vision}, 115:211--252, 2015.

\bibitem{sistu2019neurall}
Ganesh Sistu, Isabelle Leang, Sumanth Chennupati, Senthil Yogamani, Ciar{\'a}n
  Hughes, Stefan Milz, and Samir Rawashdeh.
\newblock Neurall: Towards a unified visual perception model for automated
  driving.
\newblock In {\em 2019 IEEE Intelligent Transportation Systems Conference
  (ITSC)}, pages 796--803. IEEE, 2019.

\bibitem{Strezoski2019ManyTL}
Gjorgji Strezoski, Nanne van Noord, and M. Worring.
\newblock Many task learning with task routing.
\newblock {\em 2019 IEEE/CVF International Conference on Computer Vision
  (ICCV)}, pages 1375--1384, 2019.

\bibitem{Ulyanov2016InstanceNT}
D. Ulyanov, A. Vedaldi, and V. Lempitsky.
\newblock Instance normalization: The missing ingredient for fast stylization.
\newblock {\em ArXiv}, abs/1607.08022, 2016.

\bibitem{van2019disentangled}
Sjoerd van Steenkiste, Francesco Locatello, J{\"u}rgen Schmidhuber, and Olivier
  Bachem.
\newblock Are disentangled representations helpful for abstract visual
  reasoning?
\newblock In {\em Advances in Neural Information Processing Systems}, pages
  14245--14258, 2019.

\bibitem{Vandenhende2019BranchedMN}
Simon Vandenhende, Bert~De Brabandere, and L. Gool.
\newblock Branched multi-task networks: Deciding what layers to share.
\newblock {\em ArXiv}, abs/1904.02920, 2019.

\bibitem{Vandenhende2020MTINetMT}
Simon Vandenhende, S. Georgoulis, and L. Gool.
\newblock Mti-net: Multi-scale task interaction networks for multi-task
  learning.
\newblock In {\em ECCV}, 2020.

\bibitem{Wang2018RecoveringRT}
Xintao Wang, K. Yu, C. Dong, and Chen~Change Loy.
\newblock Recovering realistic texture in image super-resolution by deep
  spatial feature transform.
\newblock {\em 2018 IEEE/CVF Conference on Computer Vision and Pattern
  Recognition}, pages 606--615, 2018.

\bibitem{wu2017visual}
Qi Wu, Damien Teney, Peng Wang, Chunhua Shen, Anthony Dick, and Anton van~den
  Hengel.
\newblock Visual question answering: A survey of methods and datasets.
\newblock {\em Computer Vision and Image Understanding}, 163:21--40, 2017.

\bibitem{Wu2018GroupN}
Yuxin Wu and Kaiming He.
\newblock Group normalization.
\newblock In {\em ECCV}, 2018.

\bibitem{Xu2018PADNetMG}
D. Xu, Wanli Ouyang, X. Wang, and N. Sebe.
\newblock Pad-net: Multi-tasks guided prediction-and-distillation network for
  simultaneous depth estimation and scene parsing.
\newblock {\em 2018 IEEE/CVF Conference on Computer Vision and Pattern
  Recognition}, pages 675--684, 2018.

\bibitem{xu2015learning}
Zhenqi Xu, Shan Li, and Weihong Deng.
\newblock Learning temporal features using lstm-cnn architecture for face
  anti-spoofing.
\newblock In {\em 2015 3rd IAPR Asian Conference on Pattern Recognition
  (ACPR)}, pages 141--145. IEEE, 2015.

\bibitem{ye2020seeing}
Xin Ye and Yezhou Yang.
\newblock From seeing to moving: A survey on learning for visual indoor
  navigation (vin).
\newblock {\em arXiv preprint arXiv:2002.11310}, 2020.

\bibitem{zador2019critique}
Anthony~M Zador.
\newblock A critique of pure learning and what artificial neural networks can
  learn from animal brains.
\newblock {\em Nature communications}, 10(1):1--7, 2019.

\bibitem{Zhang2018TheUE}
Richard Zhang, Phillip Isola, Alexei~A. Efros, E. Shechtman, and O. Wang.
\newblock The unreasonable effectiveness of deep features as a perceptual
  metric.
\newblock {\em 2018 IEEE/CVF Conference on Computer Vision and Pattern
  Recognition}, pages 586--595, 2018.

\bibitem{Zhang2018JointTL}
Z. Zhang, Zhen Cui, Chunyan Xu, Zequn Jie, Xiang Li, and Jian Yang.
\newblock Joint task-recursive learning for semantic segmentation and depth
  estimation.
\newblock In {\em ECCV}, 2018.

\bibitem{Zhang2019PatternAffinitivePA}
Z. Zhang, Zhen Cui, Chunyan Xu, Yan Yan, N. Sebe, and J. Yang.
\newblock Pattern-affinitive propagation across depth, surface normal and
  semantic segmentation.
\newblock {\em 2019 IEEE/CVF Conference on Computer Vision and Pattern
  Recognition (CVPR)}, pages 4101--4110, 2019.

\bibitem{zhao2018modulation}
Xiangyun Zhao, Haoxiang Li, Xiaohui Shen, Xiaodan Liang, and Ying Wu.
\newblock A modulation module for multi-task learning with applications in
  image retrieval.
\newblock In {\em ECCV}, 2018.

\end{thebibliography}
}

\end{document}